\documentclass{article} %
\usepackage[final]{colm2026_conference}

\usepackage{microtype}
\usepackage{hyperref}
\usepackage{url}
\usepackage{amsthm}
\usepackage{booktabs}
\usepackage{subcaption}
\usepackage{enumitem}
\usepackage{tikz}
\newcommand{\circnum}[1]{%
  \tikz[baseline=(char.base)]{
    \node[shape=circle, draw, inner sep=1.5pt] (char) {\small #1};
  }%
}

\usepackage{tcolorbox}
\usepackage{etoc}

\theoremstyle{definition}
\newtheorem{definition}{Definition}

\usepackage{amsmath,amsfonts,bm}

\def\eqref#1{equation~\ref{#1}}

\def\1{\bm{1}}

\DeclareMathAlphabet{\mathsfit}{\encodingdefault}{\sfdefault}{m}{sl}
\SetMathAlphabet{\mathsfit}{bold}{\encodingdefault}{\sfdefault}{bx}{n}

\newcommand{\KL}{D_{\mathrm{KL}}}

\usepackage{cleveref}
\crefname{appendix}{Appendix}{Appendices}
\Crefname{appendix}{Appendix}{Appendices}

\usepackage{todonotes}
\usepackage[normalem]{ulem}

\definecolor{darkgreen}{rgb}{0.0, 0.5, 0.0}

\makeatletter
\newcommand*\iftodonotes{\if@todonotes@disabled\expandafter\@secondoftwo\else\expandafter\@firstoftwo\fi}  %
\makeatother

\usepackage{xspace}
\newcommand{\logitlens}{Logit Lens\xspace}

\definecolor{probcolor}{RGB}{0, 112, 192}    %
\definecolor{embcolor}{RGB}{192, 80, 77}     %
\definecolor{matcolor}{RGB}{80, 150, 80}     %

\newcommand{\prob}{{\color{probcolor}p}}

\newcommand{\emb}{{\color{embcolor}e}}
\newcommand{\ctemb}{{\color{embcolor}\tilde{e}}}
\newcommand{\hidden}{{\color{embcolor}h}}

\definecolor{unembcolor}{RGB}{128, 0, 128}   %
\newcommand{\Eemb}{{\color{matcolor}E}}
\newcommand{\Eunemb}{{\color{unembcolor}U}}

\newcommand{\seq}[1]{\mathbf{#1}}

\usepackage{lineno}

\definecolor{darkblue}{rgb}{0, 0, 0.5}
\hypersetup{colorlinks=true, citecolor=darkblue, linkcolor=darkblue, urlcolor=darkblue}

\title{The Illusion of Superposition? A Principled Analysis of Latent Thinking in Language Models}

\def\mystrut{\rule{0pt}{1.1\normalbaselineskip}}
\author{
\begin{tabular}{@{}l}
Michael Rizvi-Martel$^{1,2}$\thanks{Corresponding author. Contact: \href{mailto:michael.rizvi-martel@mila.quebec}{\texttt{\footnotesize michael.rizvi-martel@mila.quebec}}}\quad Guillaume Rabusseau$^{1,2}$\quad Marius Mosbach$^{1,3}$ \mystrut \\
\end{tabular}\\ [1.4em]
$^1$Mila -- Quebec AI Institute \quad $^2$Universit{\'e} de Montr{\'e}al \quad $^3$McGill University\\
}

\begin{document}
\etocdepthtag.toc{default}

\ifcolmsubmission
\linenumbers
\fi

\maketitle

\begin{abstract}
Latent reasoning via continuous chain-of-thoughts (Latent CoT) has emerged as a promising alternative to discrete CoT reasoning.
Operating in continuous space increases expressivity and has been hypothesized to enable \textit{superposition}: the ability to maintain multiple candidate solutions simultaneously within a single representation.
Despite theoretical arguments, it remains unclear whether language models actually leverage superposition when reasoning using latent CoTs.
We investigate this question across three regimes: a \textit{training-free} regime that constructs latent thoughts as convex combinations of token embeddings, a \textit{fine-tuned} regime where a base model is adapted to produce latent thoughts, and a \textit{from-scratch} regime where a model is trained entirely with latent thoughts to solve a given task.
Using \logitlens and entity-level probing to analyze internal representations, we find that only models trained \textit{from scratch} exhibit signs of superposition. In training-free and fine-tuned regimes, superposition either collapses or is not used at all, with models discovering shortcut solutions instead.
We propose mechanisms explaining both the failure of training-free approaches and the limitations of training latent reasoners from scratch, finding that: i) in the training-free case, pretraining biases models to commit to a single token, collapsing the superposition; ii) in the from-scratch regime, success is governed by embedding width rather than depth.
Together, our results shed light on when and why superposition arises in continuous chain-of-thought reasoning, and propose hypotheses as to conditions which drive its collapse.
\looseness-1
\end{abstract}

\section{Introduction}
\label{sec:introduction}

Chain-of-thought (CoT) reasoning has become the standard approach for tackling complex problems with large language models (LLMs), enabling them to break down problems by reasoning ``step-by-step''~\citep{wei2022chain}.
Various works have tried to further improve LLM reasoning through methods like self-consistency~\citep{wang2022selfconsistency}, tree-of-thoughts~\citep{yao2023tree}, and stream-of-search~\citep{gandhi2024stream}. Post-training LLMs on CoT data is now a crucial part of the LLM training pipeline~\citep{o3_system_card_2025, guo2025deepseek}.
Exploring an alternative approach for reasoning, recent work proposed to have LLMs reason \textit{directly in latent space}~\citep{hao2024training, butt2025soft}, which has been shown theoretically to be more expressive than discrete CoT~\citep{zhu2025reasoning}.
A compelling hypothesis for latent CoT's advantage is \textit{superposition}: models could maintain multiple candidate solutions simultaneously, exploring several reasoning paths before committing to an answer~\citep{zhu2025reasoning}.
This would represent a fundamental advantage over discrete CoT, which must commit to a single token at each step.
However, there is little empirical evidence that LLMs leverage this capability, motivating the following research question:

\begin{center}
\textit{Does superposition actually occur in latent CoT models?}
\end{center}

We investigate this question across three complementary settings.
First, we analyze Soft Thinking~\citep{zhang2025soft}, a training-free latent CoT method that creates superposition by computing linear combinations of input embeddings.
Second, we examine Coconut~\citep{hao2024training}, a method that fine-tunes models to reason with continuous latent thoughts.
Lastly, we analyze a from-scratch variant of Coconut, where a small GPT-2-style model is entirely trained with latent thoughts to solve a given task.

Empirically, we make the following contributions:
First, using \logitlens~\citep{nostalgebraist2020logitlens} to probe internal representations, we find that \textbf{off-the-shelf LLMs collapse superposed inputs to a single interpretation} within the first few layers: 
when comparing with a discrete CoT baseline,
entropy profiles are nearly identical. 
Moreover, replacing a soft token with a standard one yields nearly indistinguishable KL divergence and cosine similarity.
Second, through entity-level probing, we show that \textbf{the Coconut model learns to extract answers directly from the question representations}, achieving comparable accuracy \textit{without any latent tokens}. 
Our belief evolution analysis explains this: we find that models do not leverage step-by-step reasoning during latent computation.
Third, when applying the same Coconut analysis to a model trained from scratch, we find that it \textbf{indeed shows signs of leveraging superposition}: the model encodes uncertainty between the correct next step and other possible next steps within its latent thoughts.
Together, these results suggest that \textit{training from scratch} is the most conducive mechanism to develop superposition in models. 

Based on these results, we conduct an additional set of experiments aimed at better understanding this discrepancy. This analysis yields two distinct findings:
\begin{enumerate}[noitemsep,topsep=0pt,leftmargin=1.75em,label=\protect\circnum{\arabic*}]
    \item \textbf{Models trained on next-token prediction learn to commit to a token in the last layers.} We find that across all considered training-free and fine-tuned models, entropy of the logit distributions drops heavily at the last layers. This drop is much less significant in from-scratch models. 
    \item \textbf{Embedding size matters.} We find that from-scratch models suffer from a \textit{width bottleneck}: larger embedding widths have a much more pronounced effect than depth on accuracy on tasks which require parallel exploration.
\end{enumerate}
We believe our work highlights important caveats of current latent thinking methodologies and offers principled guidelines to design the next generation of latent reasoning models.
\section{Related Work}
\label{sec:related-work}

\paragraph{Latent Reasoning.}

Many works investigate the use of continuous tokens and latent representations in LLMs. \citet{hao2024training} show that fine-tuning LLMs to output a reasoning trace of continuous tokens provides considerable gains on logical reasoning tasks that require search during planning.
\citet{zhu2025reasoning} popularized the notion of ``superposition'' by showing theoretically that superposition in the latent state allows transformer models to solve graph reachability tasks more efficiently.
Follow-up work by \citet{butt2025soft} proposes a novel method to train continuous CoTs via reinforcement learning which achieves comparable performance to discrete CoT on known math reasoning benchmarks.
We base our analysis on the ``Soft Thinking'' approach by \citet{zhang2025soft}; a training-free method to generate latent CoTs based on convex combinations of embedding vectors.
They report that their method offers a slight improvement on math benchmarks compared to discrete CoT baselines.
Also close to our work, \citet{deng2025latent} claim to have devised a training scheme which enables superposition in LLMs.

Alternative continuous thinking schemes have also been explored.
Many works investigate the use of ``filler'' or ``thinking'' tokens: blank tokens which can be used to store intermediary computations \citep{pfau2024let, goyal2023think, herel2024thinking}.
Moreover, there is a growing interest in ``looped layers'', a method which trains transformers with recurrent attention layers~\citep{yang2023looped, mcleish2025teaching}. These methods also show promise in increasing reasoning abilities through additional latent computation.

\paragraph{Interpretability of Reasoning Models.}

Understanding the internal representations of models has been a central research topic in NLP, even prior to LLMs \citep[\textit{inter alia}]{adi2016fine,linzen2016assessing,gulordava-etal-2018-colorless,belinkov-2022-probing}.
Recently, several works investigated how CoT reasoning changes internal computations~\citep{yang2024large, dutta2024think, cywinski2025towards}.
To the best of our knowledge, no other works have attempted to understand the inner workings of latent CoT models from an interpretability perspective.\looseness-1

\section{Background}
\label{sec:method}

\begin{figure}[t]
    \centering
    \includegraphics[width=0.95\linewidth]{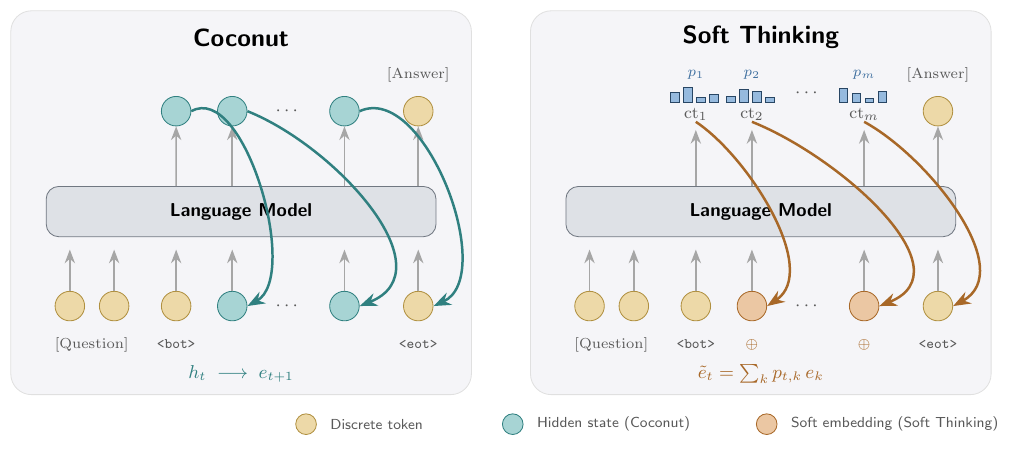}
    \caption{Two latent CoT approaches. \textbf{Left:} Coconut feeds the last hidden state directly back as the next input embedding, forming a recurrent loop in continuous space. \textbf{Right:} Soft Thinking computes a probability distribution over the vocabulary and forms the next input as a weighted sum of token embeddings. Both methods replace discrete reasoning tokens with continuous representations, but differ in how these representations are constructed.}
    \label{fig:latent-cot}
\end{figure}

Let $\mathcal{M}$ be a transformer language model with $L$ layers and vocabulary $\mathcal{V}$. We denote the \emph{embedding matrix} by $\Eemb \in \mathbb{R}^{|\mathcal{V}| \times d}$, mapping discrete tokens to $d$-dimensional vectors, and the \emph{unembedding matrix} by $\Eunemb \in \mathbb{R}^{|\mathcal{V}| \times d}$, projecting hidden states back to the vocabulary space.\looseness-1
For a token $k \in \mathcal{V}$, we write $\emb_k = \Eemb[k]$ for its embedding. Given an input sequence $\seq{x} = (x_1, \ldots, x_n)$, at each position $i$, $\mathcal{M}$ computes a hidden representation $\hidden_i^{(\ell)} \in \mathbb{R}^d$ at each layer $\ell \in \{1, \ldots, L\}$. We start by defining the notion of superposition which is crucial to our analysis:

\begin{definition}[Superposition]
    A model reasons in \emph{superposition} at position $i$ if its hidden state $\hidden_i^{(\ell)}$ encodes in token space a distribution over multiple candidate continuations.
\end{definition}

We will consider two ways of obtaining superposition throughout this paper: \textit{forced} superposition and \textit{learned} superposition. We define both below:
\begin{definition}[Forced superposition]
    Forced superposition is superposition introduced at the input by explicitly constructing embeddings as convex combinations of token embeddings: $\ctemb = \sum_k \alpha_k \, \emb_k$ with $\alpha \in \Delta^{|\mathcal{V}|-1}$.
\end{definition}

\begin{definition}[Learned superposition]
    Learned superposition is superposition that emerges from training a model to use latent thoughts on a task which rewards parallel exploration of reasoning paths.
\end{definition}

\paragraph{CoT Generation.}

We consider a setting where, given an input query, a model generates a CoT followed by a final answer.
A sequence consists of \emph{input tokens} $\seq{x} = (x_1, \ldots, x_n)$, \emph{reasoning tokens} $\seq{r} = (r_1, \ldots, r_T)$, and \emph{answer tokens} $\seq{y} = (y_1, \ldots, y_m)$. Generation proceeds autoregressively: at step $t$, the model computes $\prob(r_t \mid \seq{x}, r_1, \ldots, r_{t-1})$ and selects a token according to some decoding strategy (we focus on greedy decoding). In \emph{discrete} CoT, each $r_t \in \mathcal{V}$ is a vocabulary token with embedding $\emb_{r_t}$ fed as input to the next step. The key difference with \emph{latent} CoT is which point in embedding space is used: discrete CoT is constrained to the vocabulary manifold $\{\emb_k : k \in \mathcal{V}\}$, while latent CoT can use any point in embedding space.\looseness-1

\paragraph{Soft Thinking.}
Soft Thinking~\citep{zhang2025soft} is a form of \textit{forced} superposition which uses information from the logit distribution to craft the superposition.
At each step $t$, instead of selecting a discrete token, the method computes a distribution over the vocabulary simplex $\prob_t \in \Delta^{|\mathcal{V}|-1}$ and forms the embedding
\begin{equation}
\ctemb_t = \sum_{k \in \mathcal{V}} \prob_{t,k} \, \emb_k,
\end{equation}
which lies in the convex hull of vocabulary embeddings.
According to \citet{zhang2025soft}, this method ``naturally preserves a `superposition' which retains the entire information in each step''.

\paragraph{Coconut.}

Coconut~\citep{hao2024training} takes a different approach: instead of constructing soft embeddings from vocabulary distributions, it feeds the model's own last hidden representation back as the next input embedding, enabling recurrent ``reasoning in continuous latent space.''
The model is trained via a staged curriculum, progressively replacing discrete CoT tokens with continuous latent thoughts.
On ProsQA, a synthetic graph-traversal QA task, the authors report that latent tokens encode a breadth-first search (BFS) over the graph, citing \logitlens probing that reveals intermediate entities at latent positions.
In~\Cref{sec:coconut}, we revisit these claims and conduct an interpretability analysis on trained Coconut models.

\paragraph{\logitlens.}

To understand how Soft Thinking tokens are processed, we employ \logitlens~\citep{nostalgebraist2020logitlens}, a technique for interpreting intermediate LLM computations.
Normally, only the final layer representation $\hidden_i^{(L)}$ is projected to vocabulary space via $\prob^{(L)}_i = \mathrm{softmax}(\Eunemb \, \hidden_i^{(L)})$.
\logitlens applies this projection to intermediate representations, yielding $\prob^{(\ell)}_i = \mathrm{softmax}(\Eunemb \, \hidden_i^{(\ell)})$ at any layer $\ell$, revealing how predictions evolve across layers.
For Soft Thinking, since soft tokens are linear combinations of embeddings, using \logitlens is well motivated. For Coconut, we use GPT-2, whose tied embedding/unembedding matrices justify applying logit lens as a cosine-similarity measure between latent thoughts and vocabulary tokens.

\section{Do Off-the-Shelf Models Reason in Superposition?}
\label{sec:superposition-experiments}

If LLMs are indeed capable of leveraging forced superposition, their internal representations when processing Soft Thinking tokens should differ meaningfully from those for discrete tokens, maintaining uncertainty and showing higher entropy at intermediate layers.
We test this hypothesis with two experiments: (1) a \textit{side-by-side comparison} examining entropy profiles across layers when using latent CoT vs.\ discrete CoT, and (2) an \textit{embedding-level intervention} measuring how changing a single token from Soft Thinking to discrete affects representations.

\paragraph{Experimental setup.}
We use QwQ-32B~\citep{qwq2025} (reasoning model) and Qwen2-1.5B~\citep{yang2024qwen2} (base; results in \Cref{app:st-details}). We perform our analysis on MATH500~\citep{lightman2023let}, AIME2024~\citep{amc2025aime} and a 500-example subset of the test set from 
GSM8K~\citep{cobbe2021trainingverifierssolvemath}.
We apply \logitlens at five evenly spaced layers (see \Cref{app:logit-lens-setup}) and focus on presenting QwQ-32B results on MATH500 in the main text.
Results on Qwen2-1.5B and additional results concerning AIME2024 and GSM8K are reported in \Cref{app:st-results}.

\subsection{Comparing entropy profiles in latent vs. discrete CoT}
\label{subsec:entropy_profile}

\begin{figure}[t]
    \centering
    \begin{subfigure}[b]{0.48\textwidth}
        \includegraphics[width=\textwidth]{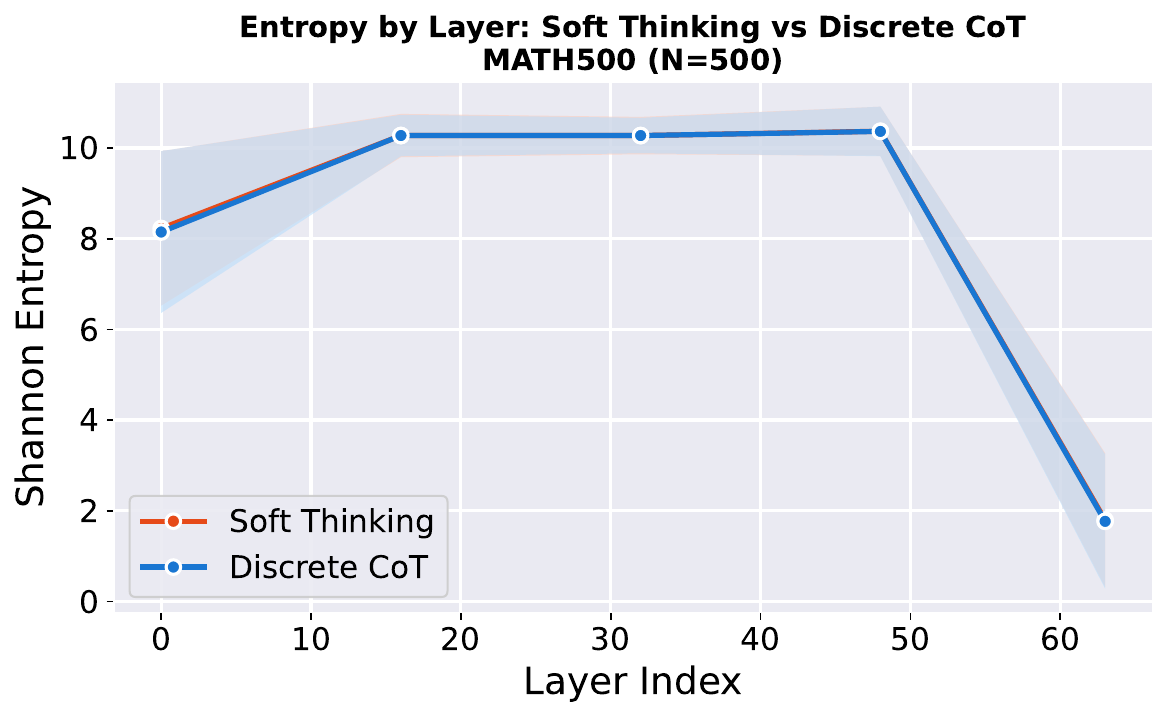}
        \caption{Entropy comparison (MATH500, N=500)}
        \label{fig:entropy_comparison}
    \end{subfigure}
    \hfill
    \begin{subfigure}[b]{0.48\textwidth}
        \includegraphics[width=\textwidth]{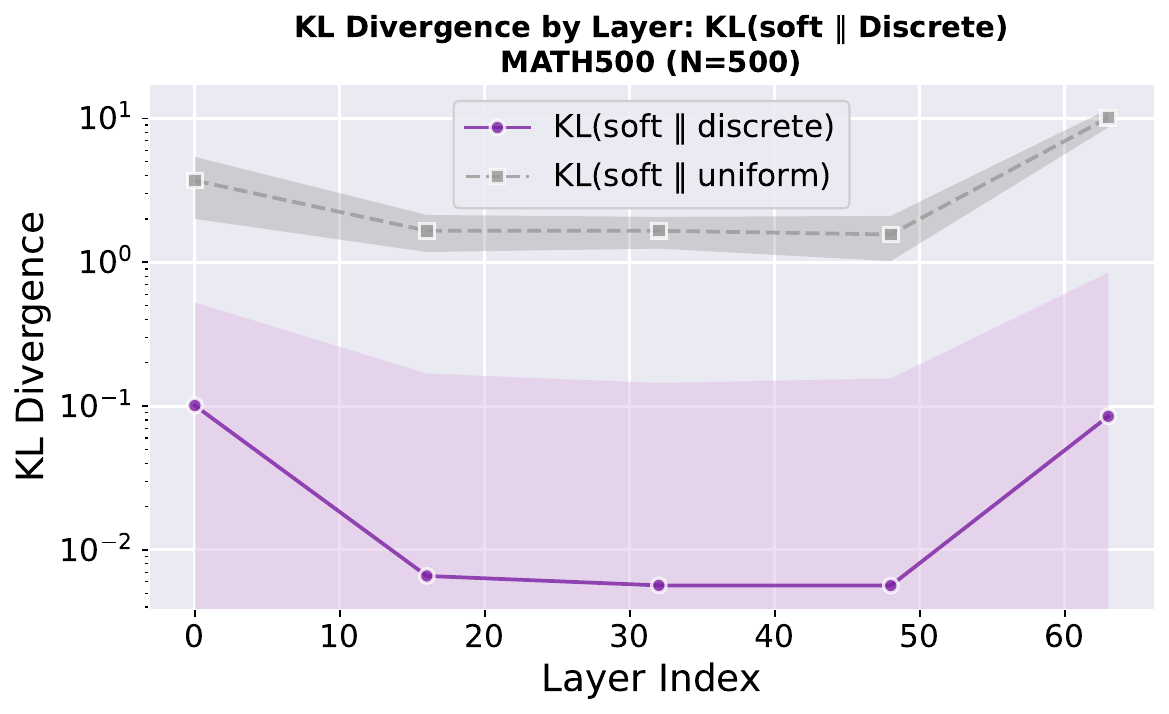}
        \caption{KL divergence (MATH500, N=500)}
        \label{fig:kl_layer}
    \end{subfigure}
    \caption{
    \textbf{Superposition collapses early in the forward pass of QwQ-32B.}
    \textbf{(a)} Shannon entropy shows identical patterns for Soft Thinking (orange) and discrete CoT (blue), both converging to near-zero entropy at the same rate.
    \textbf{(b)} KL divergence drops to $\sim 10^{-4}$ in middle layers, showing that Soft Thinking tokens become functionally identical to discrete tokens within the first few layers.
    The uncertainty in Soft Thinking token embeddings does not propagate through the network. The gray line represents the KL with a uniform distribution over the vocabulary as baseline.
    }
    \label{fig:logit_lens_analysis}
\end{figure}
We start by comparing the internal computations of a model using Soft Thinking to a discrete CoT baseline via \logitlens. We prompt both models with each problem, have them generate their respective CoTs, and at
every 50 decoding steps apply \logitlens to compute the Shannon entropy of the distribution
over $\mathcal{V}$ at selected layers.
If entropy profiles differ significantly, this would be evidence that the soft tokens meaningfully alter the internal computations.

\Cref{fig:entropy_comparison} shows entropy averaged over all CoT steps and problems.
The entropy across layers is nearly \emph{identical} for both approaches, with the same
pattern: high entropy in early-to-middle layers collapsing to near-zero at the final layer.
This is inconsistent with superposition.
If the model truly maintained multiple solutions in parallel, Soft Thinking should exhibit
higher entropy throughout; the indistinguishable profiles instead suggest that Soft Thinking
tokens are processed like discrete CoT tokens, collapsing to a single interpretation early
in the forward pass.

However, this compares independently generated CoTs that may differ in content.
To control for this, we next investigate changing only a \textit{single} token from soft to discrete.

\subsection{Intervening with discrete tokens during Soft Thinking}
\label{subsec:token_intervention}
We perform an intervention experiment to test more directly whether internal representations
differ when processing soft vs.\ discrete tokens.
At every 50 steps of a Soft Thinking generation, we run two independent forward passes:
one using the usual Soft Thinking token $\ctemb_t$, and one replacing it with the discrete
argmax embedding $\emb_{\mathrm{argmax}\,\prob_t}$.
For both, we apply \logitlens at selected layers and compute KL divergence and cosine
similarity between the resulting hidden representations.
Note that only the intervened token changes; previous tokens in the KV cache remain Soft Thinking
tokens.

\begin{table}[h]
\centering
\caption{Summary of Soft Thinking token intervention metrics. Cosine similarity is averaged across all layers, steps, and problems. Mixing entropy is the Shannon entropy of the convex combination weights $\prob_t$ used to form each soft token.
}
\label{tab:soft-thinking-summary}
\begin{tabular}{lcc}
\toprule
\textbf{Metric} & \textbf{Qwen2-1.5B} & \textbf{QwQ-32B} \\
\midrule
Cosine similarity           & $0.998 \pm 0.013$ & $0.996 \pm 0.025$ \\
Mixing weight entropy (nats) & $0.10 \pm 0.26$   & $0.18 \pm 0.34$   \\
\bottomrule
\end{tabular}
\end{table}

As can be seen in \Cref{fig:kl_layer}, the KL divergence remains small (relative to the baseline) across thinking steps and layers, achieving at most values of $10^{-1}$. The KL is largest at the first and last layers. We hypothesize that this is due to embedding differences in the first layers and to minor logit differences in the final layer.
\Cref{tab:soft-thinking-summary} corroborates this: the average cosine similarity between argmax and soft tokens is consistently high.
Finally, \Cref{fig:top_tokens_visualization} (in Appendix) shows that top predicted tokens are typically very similar. 
Moreover, token predictions with high entropy do not seem to encode ``hesitation'' between key entities, but rather show the model hesitating between prepositions or punctuation.
Taken together, these results suggest that even at the token level, Soft Thinking methods produce computations that do not significantly differ from the discrete baseline.

\section{Do Trained Models Reason in Superposition?}
\label{sec:coconut}

The previous section showed that off-the-shelf LLMs do not leverage superposition when given superposed inputs.
A natural follow-up question is whether models \emph{trained} for latent reasoning behave differently. In order to answer this question, we perform experiments using Coconut, widely regarded as one of the canonical frameworks for latent CoT training. We test both fine-tuned and from-scratch Coconut variants \citep{hao2024training}.

\subsection{Fine-Tuned Models}
\label{sec:coconut-finetuned}

\begin{figure}[t]
    \centering
    \includegraphics[width=0.9\linewidth]{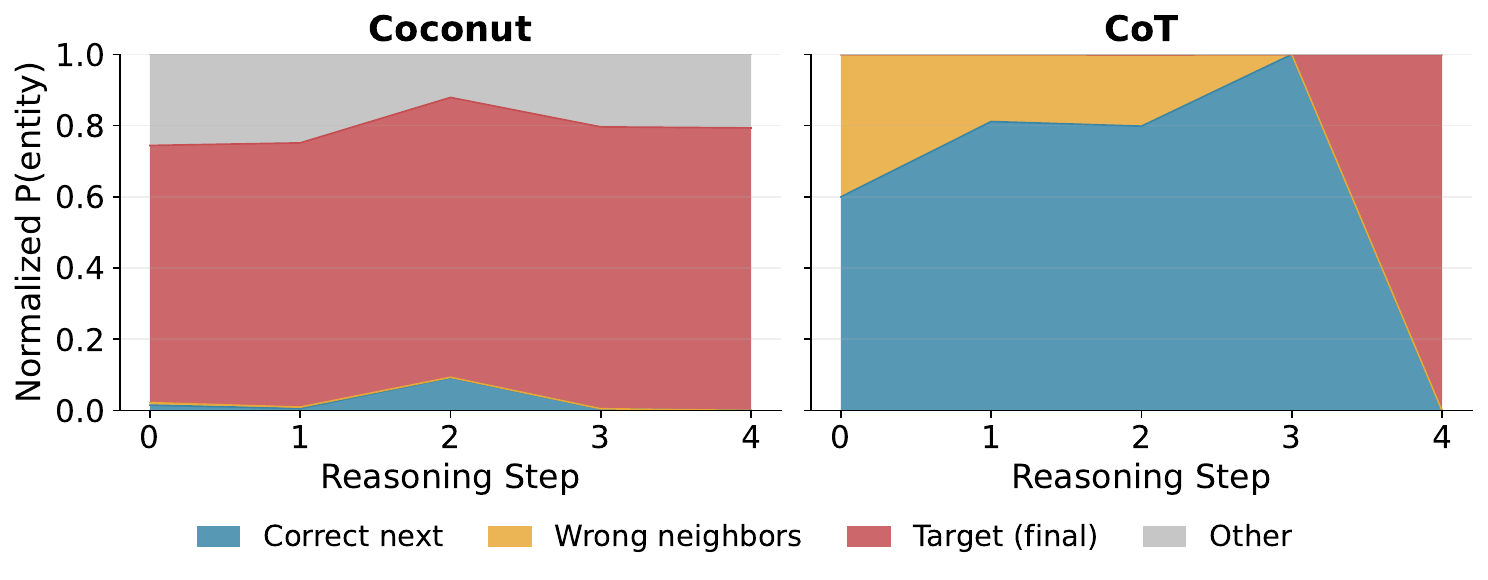}
    \caption{Step-aware entity belief for fine-tuned GPT-2 on 5-step ProsQA examples (normalized among all entities).
    \textbf{Coconut (left)}: target entity dominates from step~0 onward with no progression.
    \textbf{CoT (right)}: correct-next entity dominates early steps, target takes over at the final step, the expected pattern for genuine multi-hop reasoning.}
    \label{fig:belief-evolution-finetuned}
\end{figure}

We start by studying how fine-tuning pretrained language models to use latent thinking impacts their ability to reason in superposition. Here, we investigate if the model is able to \textit{learn} to encode such a superposition into its latent thoughts.

\paragraph{Experimental setup.}
Following the methodology of~\citet{hao2024training}, we evaluate GPT-2 (124M)~\citep{radford2019language} as well as three models from the SmolLM2 suite~\citep{allal2025smollm2} (135M, 360M and 1.7B) on ProsQA, a synthetic graph-traversal QA task requiring multi-hop logical inference over defined relationships (e.g., \textit{``Every dax is a wug. Every dax is a zug. Every wug is a blicket. Rex is a dax. Is Rex a blicket or a gorple? Blicket.''}).
The CoT baseline is trained with standard CoT supervised fine-tuning.
The Coconut model is trained with the staged curriculum of \citet{hao2024training}, which
progressively replaces CoT steps with continuous latent tokens.

\paragraph{Latent tokens are unnecessary for performance.}
\label{sec:coconut-nolatent}
We perform a causal intervention to measure the effect of the latent tokens on the model's predictions.
Since ProsQA poses a binary query (of the form \textit{``Is Rex a blicket or a gorple?''}), removing the latent thoughts amounts to a \emph{do} operation on the input; we then counterfactually evaluate the log probabilities the model assigns to both outcomes, thereby measuring how it would have scored each completion without any latent computation.
As can be seen in~\Cref{tab:prosqa-accuracy} (left), the maximum observed drop is \textbf{1.0\%} across all considered models; this suggests that the latent thinking makes a negligible contribution to the performance of the Coconut model.
We hypothesize that the small increase in performance given by latent thinking is due to a positive feedback loop phenomenon.
For cases where $P($target$)$ is lower at step 0, it is possible that the ``latent thinking'' procedure increases the probability. Next, we use probing to investigate the structure of the latent thoughts to better understand what motivates this behavior.\looseness-1

\paragraph{Probing reveals no step-by-step reasoning.}
\label{sec:coconut-evolution}

\begin{table}[t]
\centering
\caption{ProsQA accuracy for fine-tuned and from-scratch models. (a) reports the no-latent evaluation across models. (b) reports it across model depth.}
\begin{minipage}{0.48\linewidth}
    \centering
    \subcaption{Fine-tuned, counterfactual no-latent evaluation.}
    {
    \footnotesize
    \setlength{\tabcolsep}{2pt}
    \begin{tabular}{lcccc}
        \toprule
        \textbf{Model} & \textbf{CoT} & \textbf{Coconut} & \textbf{No lat.} & \textbf{Drop} \\
        \midrule
        GPT-2 (124M) & 85.3 & 99.0 & 99.0 & $-0.0$ \\
        SmolLM2-135M & 72.7 & 93.3 & 92.3 & $-1.0$ \\
        SmolLM2-360M & 85.0 & 98.7 & 98.0 & $-0.7$ \\
        SmolLM2-1.7B & 98.3 & 100.0 & 100.0 & $-0.0$ \\
        \bottomrule
    \end{tabular}
    }
\end{minipage}
\hfill
\begin{minipage}{0.48\linewidth}
    \centering
    \subcaption{Trained from scratch, 8 heads, 768-dim.}
    \begin{tabular}{lc}
        \toprule
        \textbf{Layers} & \textbf{w/ \ \ / \ \ w/o latent} \\
        \midrule
        2  & 94.5 / 13.8 \\
        4  & 96.2 / 16.0 \\
        8  & {95.2 / 53.0} \\
        12 & {91.2 / 34.1} \\
        \bottomrule
    \end{tabular}
\end{minipage}
\label{tab:prosqa-accuracy}
\end{table}

We probe by projecting hidden states at each reasoning position through the LM head. To better understand how belief evolves, we track how the normalized entity distribution evolves across reasoning steps.
At each step, graph entities are categorized into four groups: \emph{correct next} (the right answer for the given step), \emph{wrong neighbors} (nodes that are adjacent but do not lead to the correct path), \emph{target} (the final answer), and \emph{other} (entities that are part of the graph but neither the correct answer nor reachable from the current node). For instance in the example above, the \textit{correct next} would be ``wug'', the \textit{wrong neighbor} would be ``zug'', and the target entity would be ``blicket''.
For Coconut,~\Cref{fig:belief-evolution-finetuned} (left) shows the target entity dominating the distribution throughout the entire reasoning process.
For CoT, (right) the correct-next entity dominates early steps and the target takes over only at
the final step: the expected signature of step-by-step multi-hop reasoning.
This behavior suggests that the model learned a \textit{shortcut solution}: it first obtains the correct answer in a single forward pass, then copies it over through the latent tokens.

Notably, this synthetic task was explicitly designed by~\citet{hao2024training} to require parallel exploration of multiple reasoning paths, making the failure to learn superposition particularly striking. Despite this favorable setting, the model converges to a shortcut solution, suggesting that superposition is not a naturally preferred strategy under standard fine-tuning. This casts doubt on whether scaling model size or applying similar training procedures on more complex benchmarks would, by themselves, induce superposition-based reasoning. We observe similar behavior on ProntoQA~\citep{saparov2022language}; see \Cref{app:coconut-entity} for details.

\subsection{From-Scratch Models}
\label{sec:fromscratch}

\begin{figure}[t]
    \centering
    \includegraphics[width=0.9\linewidth]{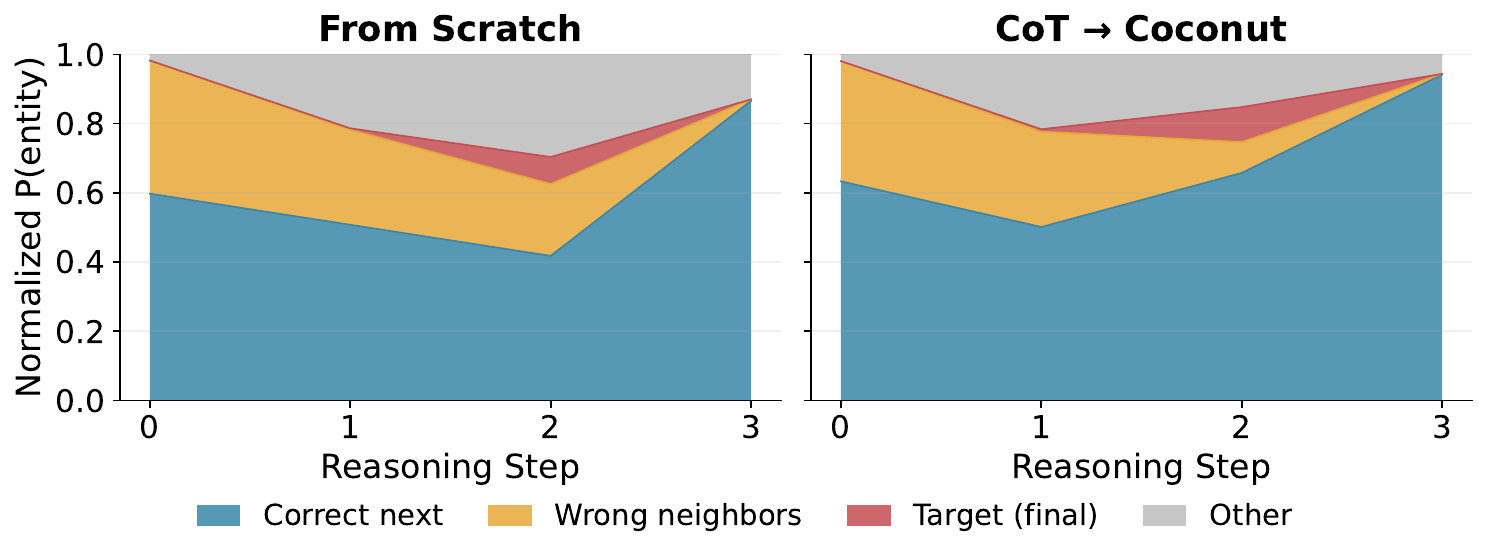}
    \caption{Step-aware entity belief for from-scratch 2-layer models on 4-step ProsQA examples.
    \textbf{From Scratch (left)}: correct-next and wrong-neighbor entities dominate intermediate steps, with belief evolving across the reasoning trace.
    \textbf{CoT $\to$ Coconut (right)}: similar pattern with slightly more concentrated belief on correct-next entities. We show 4-step examples here and 5-step examples for the fine-tuned model in \Cref{fig:belief-evolution-finetuned}; the qualitative pattern is consistent across hop counts (see \Cref{app:coconut-entity}).}
    \label{fig:belief-evolution-fromscratch}
\end{figure}

Next, we investigate the use of superposition in models trained from scratch.
We train small GPT-2 style transformers on a simplified variant of ProsQA with a symbolic tokenizer (40 tokens)~\citep{zhu2025reasoning}.
For these models, we ablate over layer count (2, 4, 8 and 12 layers) but leave the number of heads fixed at 8 and the embedding dimension fixed to 768.
We compare two training regimes: i) from scratch with latent thinking (Coconut); ii) from scratch using cross entropy on the gold CoT \textit{then} training with Coconut (CoT + Coconut).
Note that the variant of Coconut used here is different from that of~\citet{hao2024training}: here, at every step in the problem models are trained to use $i$ latent thoughts to predict the value of the $i$th node.
Crucially, this methodology never trains the model on the entire gold CoT. As in \Cref{sec:coconut-finetuned}, we employ probing to understand how belief evolves through the model's latent CoT.

\paragraph{Superposition occurs in from-scratch models.}

\Cref{fig:belief-evolution-fromscratch} shows belief evolution across thinking steps for from-scratch Coconut variants.
For both the Coconut and CoT+Coconut training methods, the models show evidence of leveraging superposition. The state is dominated by the correct next entity probability but still leaves significant probability to the other potential neighbors. This remains true even when the model is first trained to perform the task with a discrete CoT, suggesting that next-token pretraining is not the sole factor leading to superposition collapse.

\paragraph{Latent tokens are necessary for performance.} As can be seen in \Cref{tab:prosqa-accuracy} (right),
models trained from scratch do indeed leverage their latent thoughts contrary to the fine-tuned case. Removing access to latent steps produces significant performance drops across all depths (for instance, from 94.5\% to 13.8\% for the 2-layer model), thus corroborating the entity belief results. 
This remains true even when training larger and larger models, all the way up to the size and depth of the pretrained GPT-2 model (12 layers, 768 embedding dim and 8 heads), which is surprising given the theoretical construction proposed by~\citet{zhu2025reasoning} needed only 2 layers. This suggests that shortcut learning in the fine-tuned case is not solely explained by the capacity of the model relative to the task; there may be biases from the pretraining or fine-tuning procedures which incentivize shortcut learning.

\section{Exploring the Limitations of Latent Thinking}
\label{sec:why}

The previous two sections provide evidence that superposition only occurs in a very limited set of scenarios: only from-scratch models manage to leverage superposition in their reasoning process. 
In this section, we further investigate this discrepancy through two distinct lenses. First, we propose an explanation as to why superposition collapses in the training-free regime: next-token pretraining biases models to commit to a single token in its final layers, projecting superposed inputs onto a near-discrete representation. Second, we characterize \textit{when} learned superposition emerges in from-scratch models. We show empirically that width has a far greater effect than depth on performance and models' propensity to leverage latents.

\paragraph{Models trained on next-token prediction commit to a token in the last layers.}

\begin{figure}[t]
    \centering
    \includegraphics[width=0.49\linewidth]{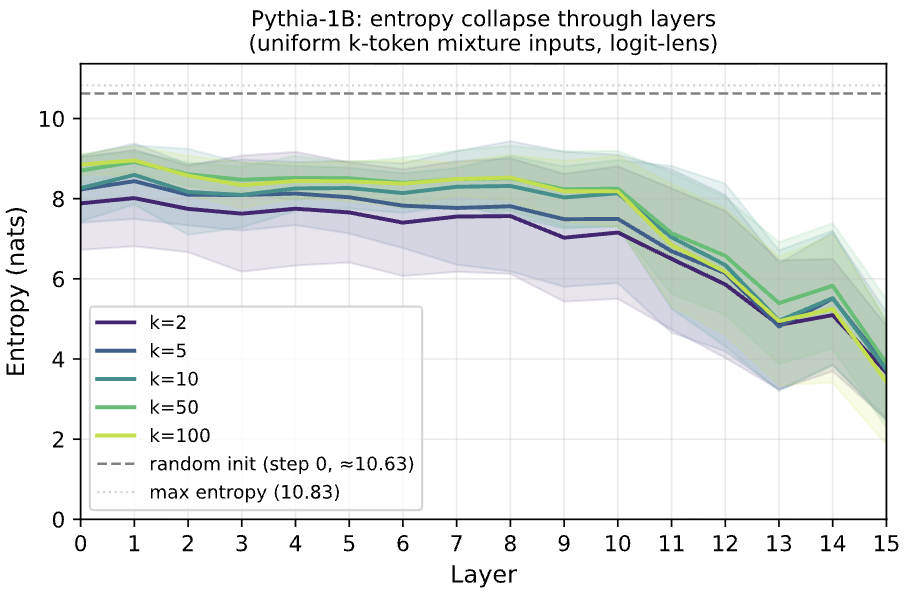}
    \hfill
    \includegraphics[width=0.49\linewidth]{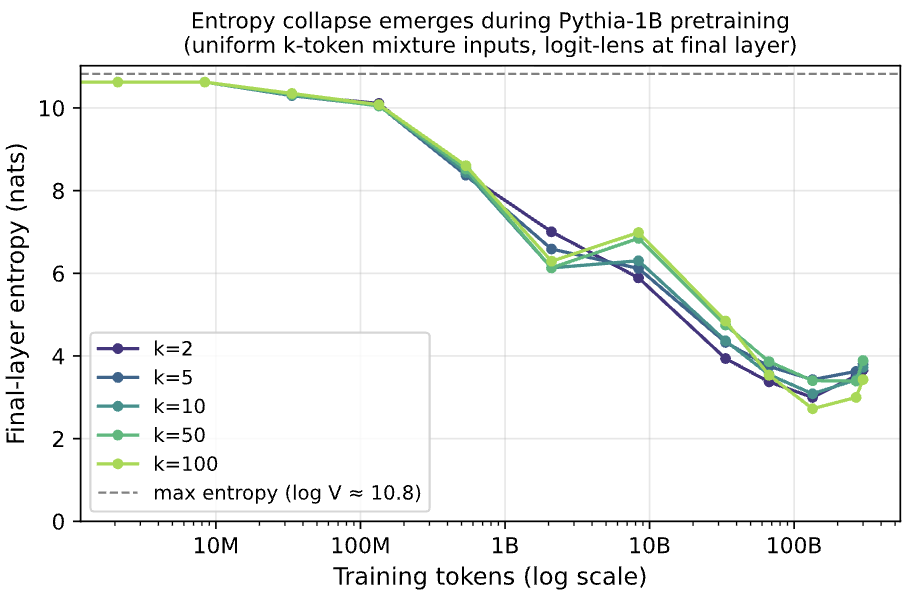}
    \caption{Entropy of synthetic uniform superpositions for Pythia-1B, where $k$ tokens are sampled uniformly at random and combined with equal weights ($1/k$).
    \textbf{Left}: entropy across layers of the fully pretrained model collapses at the final layers regardless of $k$; the dashed line shows a randomly initialized model, whose entropy stays near the maximum.
    \textbf{Right}: final-layer entropy across pretraining checkpoints, showing that the collapse emerges gradually as the number of training tokens increases.
    Shaded regions show $\pm 1$ std.}
    \label{fig:attractor}
\end{figure}

\Cref{fig:attractor} (left) shows, for Pythia-1B, the entropy across layers of a forced superposition where all tokens in the combination have uniform weight. Uniform weights are chosen to avoid the confound of soft tokens computed from peaky logit distributions.
Moreover, we compare the pretrained model to a model with weights reinitialized at random to isolate the confound of the learning process and the architecture.
Across different numbers of tokens in the mixture, the trend is clear: entropy drops rapidly when reaching the final layers. This contrasts sharply with the random weights baseline: here, the entropy remains high throughout. That this phenomenon only occurs in the pretrained model suggests that the pretraining is the driving factor to the final-layer entropy collapse.
We also note that fine-tuning does not seem to be enough to fix this issue; the last-layer entropy collapse persists when fine-tuning GPT-2 using the Coconut methodology (see \Cref{app:coconut-logitlens}).

\paragraph{Entropy collapse appears gradually during training.}

To determine if the final-layer entropy collapse is actually a consequence of the pretraining process, we evaluate the final-layer entropy over uniform superpositions at successive Pythia-1B pretraining checkpoints~\citep{biderman2023pythia}.
As shown in~\Cref{fig:attractor} (right), for all combinations of soft tokens the entropy starts near its maximum and decreases gradually as the number of training tokens increases.
This gradual emergence suggests that the collapse is tied to the pretraining process, and not merely to the output geometry of a fully trained model.

\paragraph{Embedding size matters when training models from scratch.}

Our previous results indicate that latent reasoners trained from scratch learn to leverage their latent CoT to solve the task, even as depth increases.
In this section, we turn our attention to the effect of \textit{width} on latent reasoning.
We train GPT-2 style models using the same setup as described in \Cref{sec:fromscratch}. We start by performing a \textit{parameter-matched} experiment: keeping the overall size of the model constant, we vary the depth to width ratio. The 2-layer model configuration by~\citet{zhu2025reasoning} serves as a baseline for parameter count. \Cref{tab:param-matched} shows that at fixed parameter count, shallower, wider models outperform deeper, narrower ones: the 2- and 4-layer variants exceed 90\% accuracy, while the deeper 8- and 12-layer models lag behind by 10 to 20 points.

To confirm this behavior, we perform an additional ablation. First, we find the critical width at which performance degrades in 2-layer and 4-layer models (see \Cref{tab:embd_ablation}). Then, for a set of embedding sizes below the critical width, we increase gradually the depth of the model, going all the way up to 12 layers. \Cref{tab:depth_x_embd} shows that increasing capacity along the depth axis yields poor performance; the best model obtains 68.3\% accuracy whereas wider models easily saturate the task, obtaining accuracy above 90\%.

\begin{table}[t]
\centering
\caption{Parameter-matched from-scratch Coconut on ProsQA (${\sim}15$M parameters). Starting from the 2-layer model, depth is increased while the embedding width is shrunk to keep the parameter count fixed. Accuracies (\%) are reported with and without latent tokens as mean $\pm$ std over 3 seeds.}
\label{tab:param-matched}
{
\begin{tabular}{lccc}
\toprule
\textbf{Model} & \textbf{Params} & \textbf{w/ latents} & \textbf{w/o latents} \\
\midrule
2L-768d  & 14.97M & $96.0 \pm 0.9$  & $30.9 \pm 8.4$  \\
4L-544d  & 14.78M & $91.8 \pm 4.9$  & $39.7 \pm 20.8$ \\
8L-384d  & 14.61M & $81.6 \pm 4.3$  & $73.2 \pm 1.4$  \\
12L-320d & 15.14M & $72.4 \pm 7.1$  & $70.0 \pm 6.3$  \\
\bottomrule
\end{tabular}
}
\end{table}
\begin{table}[t]
\centering
\caption{Test accuracy (\%) with and without latent tokens across depths, for sub-threshold embedding sizes ($d_\text{embd} \leq 64$).}
\label{tab:depth_x_embd}
{
\setlength{\tabcolsep}{4pt}
\begin{tabular}{r cc cc cc cc}
\toprule
& \multicolumn{2}{c}{\textbf{2L}} & \multicolumn{2}{c}{\textbf{4L}} & \multicolumn{2}{c}{\textbf{8L}} & \multicolumn{2}{c}{\textbf{12L}} \\
\cmidrule(lr){2-3} \cmidrule(lr){4-5} \cmidrule(lr){6-7} \cmidrule(lr){8-9}
$d_\text{embd}$ & w/ & w/o & w/ & w/o & w/ & w/o & w/ & w/o \\
\midrule
8  & 40.8 & 40.6 & 35.3 & 35.1 & 24.6 & 24.8 & 34.8 & 33.9 \\
16 & 57.5 & 54.9 & 55.6 & 55.4 & 55.4 & 55.4 & 58.9 & 58.7 \\
32 & 58.9 & 57.8 & 53.9 & 53.2 & 56.8 & 58.2 & 61.6 & 60.1 \\
64 & 64.9 & 62.3 & 63.2 & 62.5 & 67.3 & 64.4 & 68.3 & 65.4 \\
\bottomrule
\end{tabular}
}
\end{table}

\section{Discussion and Conclusion}
\label{sec:conclusion}
In conclusion, our experiments provide evidence as to when reasoning in superposition occurs and when it does not across three complementary settings.
For Soft Thinking, off-the-shelf LLMs process superposed inputs nearly identically to discrete tokens: entropy profiles match, KL divergences approach zero, and cosine similarities exceed 0.99.
For fine-tuned Coconut, counterfactually removing the latent tokens from a fine-tuned model changes its predictions by at most 1.0\% across all considered model families, and entity-level probing reveals no step-by-step reasoning during latent computation.
Finally, superposition only appears in from-scratch Coconut models: removing latents sharply hurts performance, and increasing depth does not induce shortcut learning, even at model sizes matching the pretrained GPT-2 used for fine-tuning.

\paragraph{Why does superposition collapse in pretrained models?}

As we argue above, this is not merely an inductive bias but a consequence of the
pretraining objective: autoregressive training optimizes for discrete next-token prediction,
creating representations that separate token identities.
When presented with a superposed input, the model projects it onto the nearest discrete
interpretation, precisely what the training objective rewards.
Moreover, the Soft Thinking distribution $\prob_t$ is itself very peaky across steps (see
\Cref{fig:top_tokens_visualization} and the entropy heatmaps in \Cref{app:st-details}),
meaning the input is already near-discrete and there is little superposition to exploit in
the first place.

\paragraph{Is token-level superposition desirable?}

Beyond finding that models do not reason in superposition, we question whether \emph{token-level}
superposition is a desirable property in the first place.
Many Soft Thinking tokens combine semantically unrelated items (see
\Cref{fig:top_tokens_visualization}): formatting characters, punctuation, or tokens with
similar logits but unrelated meanings.
A superposition of ``('' and ``\{'' represents syntactic uncertainty, not exploration of
alternative reasoning paths.
Meaningful parallel exploration likely requires superposition at a higher level of
abstraction (over entire reasoning strategies, not individual tokens), which we consider an interesting avenue for future work.

\paragraph{Latent reasoning as flexibility.}

Despite our negative findings for current methods, latent reasoning remains a promising
direction.
The advantage of continuous embeddings may not be superposition but \textit{flexibility}:
the ability to express intermediate computations that do not correspond to natural language
tokens, avoiding the discretization bottleneck.
Future work should investigate whether latent reasoning provides benefits through other
mechanisms, such as smoother optimization landscapes or more expressive intermediate
representations.

\paragraph{Limitations.}

It would be valuable to run similar experiments on other latent reasoning approaches, such
as the RL-trained continuous CoTs of \citet{butt2025soft} or the recurrent layer frameworks proposed by~\citet{giannou2023looped,mcleish2025teaching}. 
Moreover, our results only pertain to external latent CoTs. It is possible that models may have learned to internalize the latent thinking steps~\citep{deng2024explicit}; our results do not investigate this. Analyzing model internals through probing or circuit discovery to understand their learned reasoning schemes could be an exciting direction for future work.
Finally, we hope in future work to apply the findings from this paper to design novel latent thinking methods.

\subsubsection*{Acknowledgments}
M. Rizvi-Martel’s research is supported by NSERC (CGS-D Scholarship), and G. Rabusseau’s by NSERC and the CIFAR AI Chair program. We also acknowledge NVIDIA for providing computational resources.

\bibliography{colm2026_conference}
\bibliographystyle{colm2026_conference}
\newpage
\etocdepthtag.toc{appendix}
\appendix
\crefalias{section}{appendix}
\crefalias{subsection}{appendix}
\crefalias{subsubsection}{appendix}

\pagebreak
\section*{Appendix Contents}

\etocsettagdepth{default}{none}
\etocsettagdepth{appendix}{subsection}
\etocsettocstyle{}{}
\tableofcontents

\vspace{1em}
\hrule
\vspace{2em}

\section{Disclosure of LLM usage}
\label{app:llm-disclosure}

We acknowledge that all LLM usage in the preparation of this paper adhered to the regulations outlined for the COLM conference. We used \texttt{Claude Opus 4.6} only to assist in the implementation, data visualization, and for shortening of text originally written by the authors.

\section{Soft Thinking: Experimental Details}
\label{app:st-details}

\subsection{Models}
\label{app:models}

We use three models spanning two model families:

\begin{itemize}
    \item \textbf{QwQ-32B}: A 32.5B-parameter reasoning model based on the Qwen2.5 architecture with 64 transformer layers and a hidden dimension of 5120. We use this as our primary model since it was trained for chain-of-thought reasoning.
    \item \textbf{Qwen2-1.5B}: A 1.5B-parameter base language model with 28 transformer layers and a hidden dimension of 1536. We use this smaller model to test whether our findings generalize across model scales.
    \item \textbf{DeepSeek-R1-Distill-Llama-70B}: A 70B-parameter reasoning model distilled from DeepSeek-R1~\citep{guo2025deepseek} into the Llama architecture, with 80 transformer layers and a hidden dimension of 8192. We include this model to verify that our findings extend beyond the Qwen family to a different architecture and scale.
\end{itemize}

The Qwen models use a shared tokenizer with a vocabulary of 151,643 tokens. DeepSeek-R1-Distill-Llama-70B uses the Llama tokenizer with a vocabulary of 128,256 tokens. All experiments use the models in \texttt{bfloat16} precision.

\subsection{Soft Thinking Configuration}
\label{app:soft-thinking-config}

We use the following decoding hyperparameters for all \logitlens experiments, consistent across both models:

\begin{table}[htbp]
    \centering
    \caption{Soft Thinking decoding hyperparameters.}
    \label{tab:soft-thinking-hyperparams}
    \begin{tabular}{lcc}
    \toprule
    \textbf{Parameter} & \textbf{Symbol} & \textbf{Value} \\
    \midrule
    Temperature & $\tau$ & 0.6 \\
    Top-$k$ (sampling) & $k$ & 30 \\
    Soft top-$k$ (embedding mix) & $k_\mathrm{soft}$ & 15 \\
    Weighting scheme & & Softmax \\
    Max new tokens & $T_\mathrm{max}$ & 2048 \\
    \bottomrule
    \end{tabular}
\end{table}

At each reasoning step $t$, the model computes logits over the vocabulary and selects the top-$k_\mathrm{soft}$ tokens. Their logits are normalized via softmax to obtain the mixing weights $\prob_t \in \Delta^{k_\mathrm{soft}-1}$, and the Soft Thinking embedding is formed as $\ctemb_t = \sum_{i=1}^{k_\mathrm{soft}} \prob_{t,i} \, \emb_{v_i}$ where $v_1, \ldots, v_{k_\mathrm{soft}}$ are the top-$k_\mathrm{soft}$ tokens by logit value.

For the benchmark evaluation runs (\Cref{tab:math500,tab:aime2024}), we additionally vary the cold stop threshold $c \in \{0.0, 0.01, 0.1, 0.2\}$, which terminates Soft Thinking when the top token probability exceeds $1 - c$.

\subsection{\logitlens Setup}
\label{app:logit-lens-setup}

We apply \logitlens at 5 evenly spaced probe layers: $\{0, \lfloor L/4 \rfloor, \lfloor L/2 \rfloor, \lfloor 3L/4 \rfloor, L-1\}$, where $L$ is the number of transformer layers. This corresponds to layers $\{0, 7, 14, 21, 27\}$ for Qwen2-1.5B, $\{0, 16, 32, 48, 63\}$ for QwQ-32B, and $\{0, 20, 40, 60, 79\}$ for DeepSeek-R1-Distill-Llama-70B.

\paragraph{Entropy profile comparison (\Cref{subsec:entropy_profile}).}
For each problem, we run two independent generations: one using Soft Thinking and one using standard discrete decoding (greedy argmax). Every 50 decoding steps, we apply \logitlens at each probe layer and record the Shannon entropy of the resulting distribution over $V$. We also store the top-10 predicted tokens at each checkpoint for qualitative comparison.

\paragraph{Token-level intervention~(\Cref{subsec:token_intervention}).}
During Soft Thinking generation, we intervene every 50 steps by performing two fresh forward passes over the full sequence: one using the Soft Thinking embedding and one replacing it with the argmax embedding. Crucially, only the current token's embedding differs; the KV cache from previous (soft) tokens is shared. At each probe layer, we compute:
\begin{itemize}
    \item KL divergence: $\KL(\prob_\mathrm{soft}^{(\ell)} \| \prob_\mathrm{argmax}^{(\ell)})$ between the \logitlens distributions.
    \item Cosine similarity: $\cos(\hidden_\mathrm{soft}^{(\ell)}, \hidden_\mathrm{argmax}^{(\ell)})$ between hidden representations.
    \item Entropy difference: $H(\prob_\mathrm{soft}^{(\ell)}) - H(\prob_\mathrm{argmax}^{(\ell)})$.
    \item Top-$k$ token overlap: $|S_\mathrm{soft}^{(\ell)} \cap S_\mathrm{argmax}^{(\ell)}| / k$ for $k = 10$.
\end{itemize}

\subsection{Evaluation Problems}
\label{app:eval-problems}

We select 5 problems of varying difficulty from three standard math reasoning benchmarks. The same problems are used across all experiments and both models to enable direct comparison.

\begin{tcolorbox}[colback=blue!3, colframe=blue!40, title={\textbf{GSM8K, Problem 42 (Easy)}}, fonttitle=\small]
\small
Grandma Jones baked 5 apple pies for the fireman's luncheon. She cut each pie into 8 pieces and set the five pies out on the buffet table for the guests to serve themselves. At the end of the evening, after the guests had taken and eaten their pieces of pie, there were 14 pieces of pie remaining. How many pieces were taken by the guests?

\medskip
\textit{Answer:} 26
\end{tcolorbox}

\begin{tcolorbox}[colback=blue!3, colframe=blue!40, title={\textbf{MATH500, Problem 10 (Medium)}}, fonttitle=\small]
\small
What is the least positive integer multiple of 30 that can be written with only the digits 0 and 2?

\medskip
\textit{Answer:} 2220
\end{tcolorbox}

\begin{tcolorbox}[colback=blue!3, colframe=blue!40, title={\textbf{MATH500, Problem 55 (Medium)}}, fonttitle=\small]
\small
Suppose that $f$ is a polynomial such that $(x-1)\cdot f(x)=3x^4+x^3 - 25x^2 +38x -17.$ What is the degree of $f$?

\medskip
\textit{Answer:} 3
\end{tcolorbox}

\begin{tcolorbox}[colback=orange!3, colframe=orange!50, title={\textbf{AIME 2024, Problem 3 (Hard)}}, fonttitle=\small]
\small
Let $x,y$ and $z$ be positive real numbers that satisfy the following system of equations:
\[
\log_2\!\left(\frac{x}{yz}\right) = \frac{1}{2}, \quad
\log_2\!\left(\frac{y}{xz}\right) = \frac{1}{3}, \quad
\log_2\!\left(\frac{z}{xy}\right) = \frac{1}{4}.
\]
Then the value of $\left|\log_2(x^4y^3z^2)\right|$ is $\tfrac{m}{n}$ where $m$ and $n$ are relatively prime positive integers. Find $m+n$.

\medskip
\textit{Answer:} 33
\end{tcolorbox}

\begin{tcolorbox}[colback=orange!3, colframe=orange!50, title={\textbf{AIME 2024, Problem 5 (Hard)}}, fonttitle=\small]
\small
Alice chooses a set $A$ of positive integers. Then Bob lists all finite nonempty sets $B$ of positive integers with the property that the maximum element of $B$ belongs to $A$. Bob's list has 2024 sets. Find the sum of the elements of~$A$.

\medskip
\textit{Answer:} 55
\end{tcolorbox}

\subsection{Compute}
\label{app:compute}

All experiments were run on a SLURM cluster using NVIDIA L40S GPUs (48\,GB each). \Cref{tab:compute} summarizes the resource allocation for each experiment.

\begin{table}[htbp]
    \centering
    \caption{Compute resources per experiment.}
    \label{tab:compute}
    \begin{tabular}{llccc}
    \toprule
    \textbf{Experiment} & \textbf{Model} & \textbf{GPUs} & \textbf{RAM} & \textbf{Wall Time} \\
    \midrule
    Comparative (\logitlens) & Qwen2-1.5B & $1 \times$ L40S & 32\,GB & $< 1$\,h \\
    Comparative (\logitlens) & QwQ-32B & $4 \times$ L40S & 128\,GB & $< 2$\,h \\
    Convergence & Qwen2-1.5B & $1 \times$ L40S & 32\,GB & $< 1$\,h \\
    Convergence & QwQ-32B & $4 \times$ L40S & 128\,GB & $< 3$\,h \\
    Benchmark (MATH500) & QwQ-32B & $4 \times$ L40S & 128\,GB & varies \\
    Benchmark (AIME2024) & QwQ-32B & $4 \times$ L40S & 128\,GB & varies \\
    Coconut (ProsQA) & GPT-2 & $4 \times$ L40S & 128\,GB & $\sim 3$\,h \\
    \bottomrule
    \end{tabular}
\end{table}

For QwQ-32B, we distribute the model across 4 GPUs using HuggingFace Accelerate's \texttt{device\_map="balanced"} strategy with a per-GPU memory cap of 75\% to avoid out-of-memory errors from uneven shard allocation.

\section{Soft Thinking: Additional Results}
\label{app:st-results}
In this section, we present additional results for the Soft Thinking experiments. We present token-level comparisons, full dataset \logitlens visualizations (for models and metrics which are not shown in the main text) and additional benchmark results showing accuracy of models for different hyperparameter choices.

\begin{figure}[htbp]
    \centering
    \includegraphics[width=\linewidth]{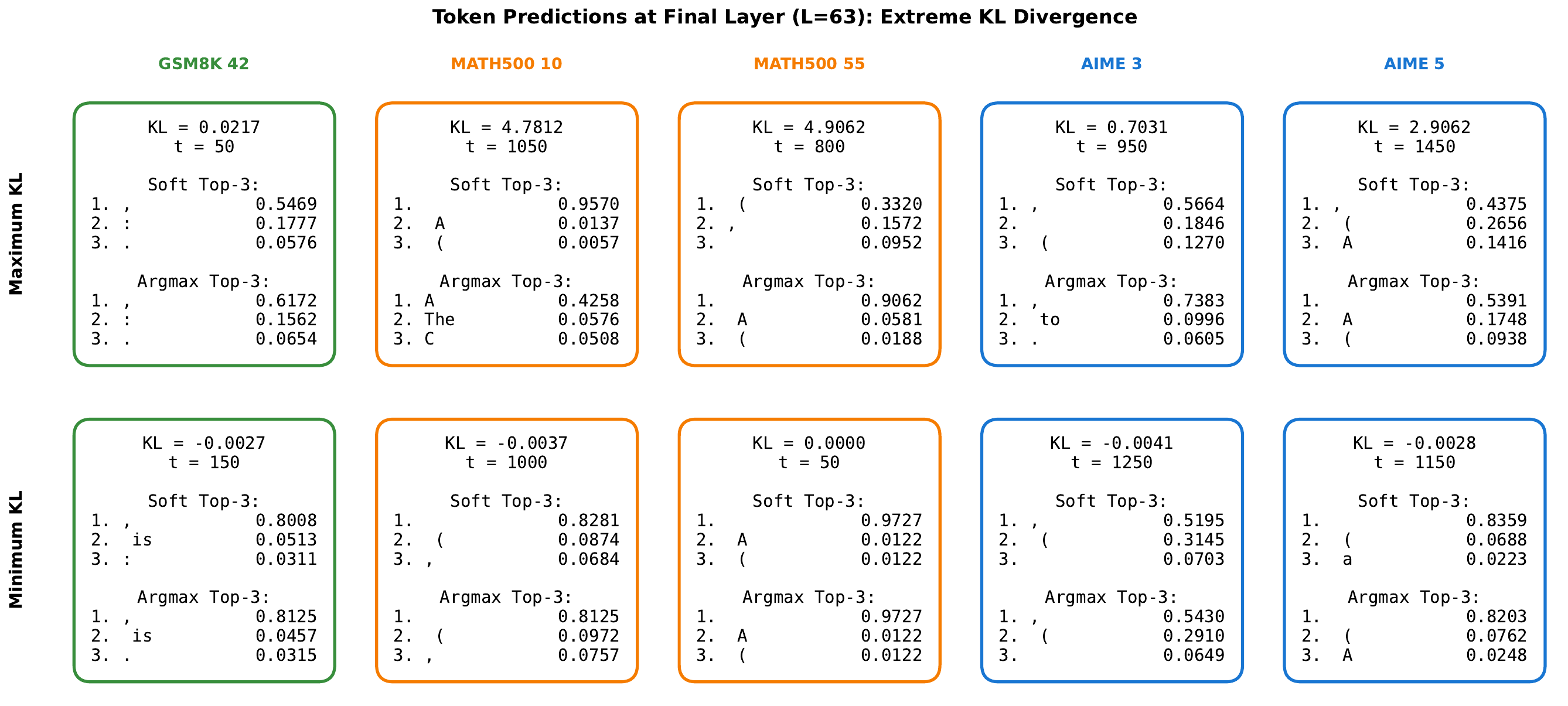}
    \caption{Top 3 tokens at the output layer for time steps with largest (top) and smallest (bottom) KL divergence between soft and argmax representations in \textbf{QwQ-32B}. Each column represents a problem instance.}
    \label{fig:top_tokens_visualization}
\end{figure}

\paragraph{Curated set vs.\ full datasets.}
The entropy profiles and KL divergence curves in \Cref{sec:superposition-experiments} (\Cref{fig:logit_lens_analysis,fig:top_tokens_visualization}) are computed on the \emph{curated set of 5 problems} described in \Cref{app:eval-problems}.
All figures in \Cref{app:fulldataset-logitlens} use the corresponding \emph{full datasets} (MATH500: $N=500$; AIME 2024: $N=30$; GSM8K: $N=500$), with sample sizes labeled in each caption.
The Soft Thinking analysis involves no stochastic training, so reproducibility across runs is exact given fixed model weights and the deterministic decoding procedure.

\subsection{Token-level comparisons}
\label{app:logit-lens-1.5b}
In this section, we include visualizations of the top-3 tokens for both the Soft Thinking and argmax decoding methods. We visualize the top-3 across 5 chosen problem instances (see \Cref{app:eval-problems} for more details) for both the step with highest and lowest KL divergence. Results show that superposition is often being performed on tokens which do not have a meaningful relationship to the problem at hand.

\begin{figure}[htbp]
    \centering
    \includegraphics[width=\linewidth]{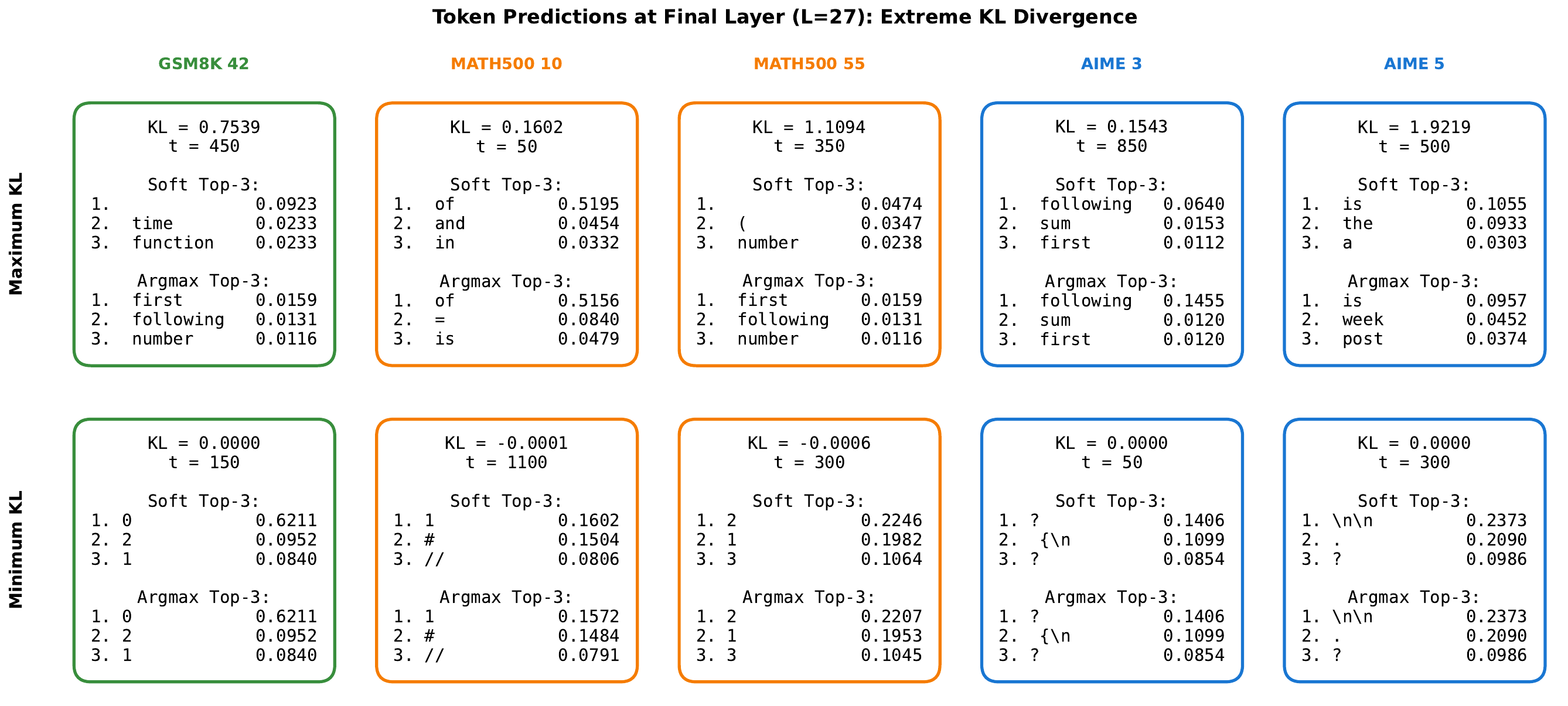}
    \caption{Top 3 predicted tokens at the output layer for time steps with largest (top) and smallest (bottom) KL divergence between soft and argmax representations in \textbf{Qwen2-1.5B}. Each column represents a problem instance.}
    \label{fig:top_tokens_1.5b}
\end{figure}

\subsection{Full-Dataset \logitlens Results}
\label{app:fulldataset-logitlens}
In this section we present visualizations of cosine similarity, KL divergence and entropy difference not shown in the main paper. This section also includes figures for Qwen2-1.5B and DeepSeek-R1-Distill-Llama-70B which are not included in the main text.
\begin{figure}[htbp]
    \centering
    \includegraphics[width=\linewidth]{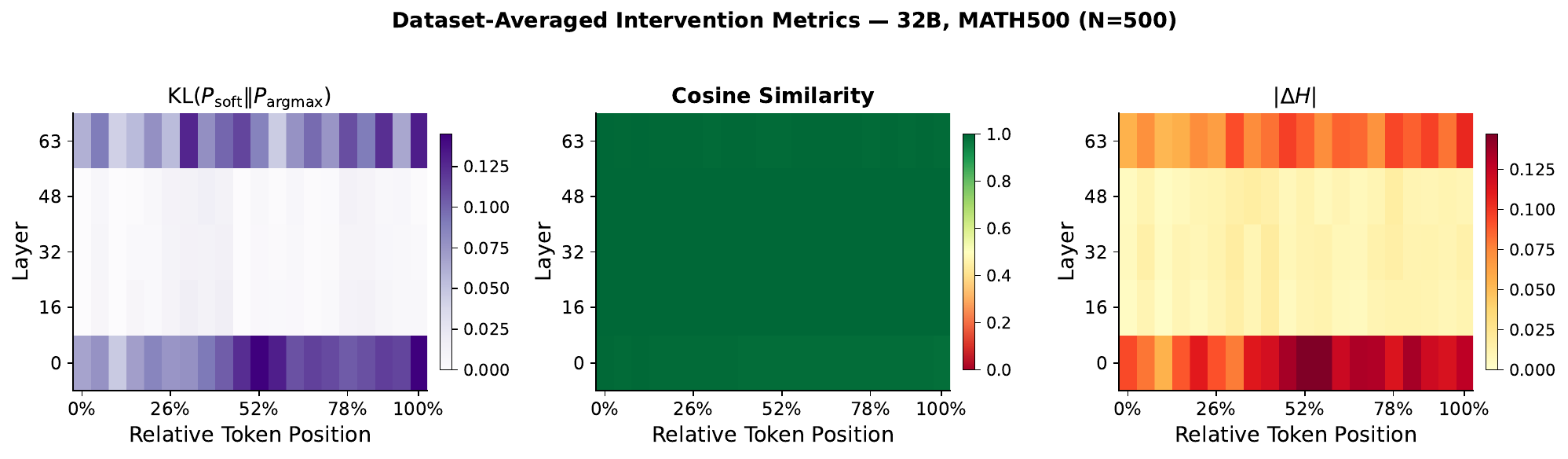}
    \caption{Dataset-averaged intervention metrics for \textbf{QwQ-32B on MATH500} (N=500). Left to right: KL divergence, cosine similarity, absolute entropy difference. Each cell is the mean over all problems at that (layer, relative position) bin.}
    \label{fig:avg-heatmap-32b-math500}
\end{figure}

\begin{figure}[htbp]
    \centering
    \includegraphics[width=\linewidth]{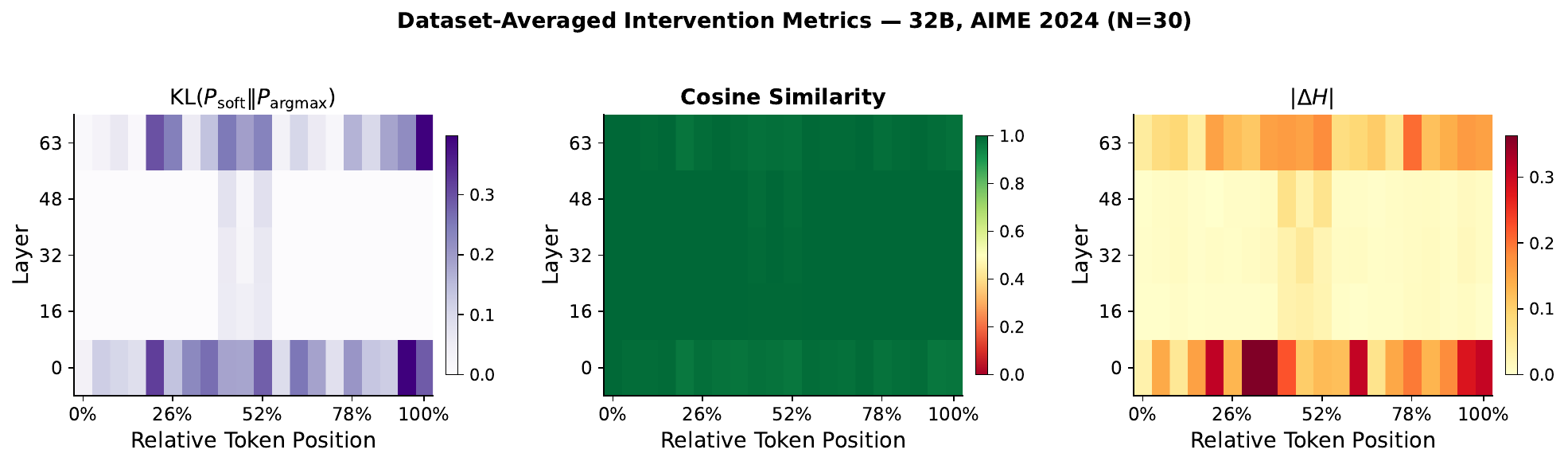}
    \caption{Dataset-averaged intervention metrics for \textbf{QwQ-32B on AIME 2024} (N=30).}
    \label{fig:avg-heatmap-32b-aime}
\end{figure}

\begin{figure}[htbp]
    \centering
    \includegraphics[width=\linewidth]{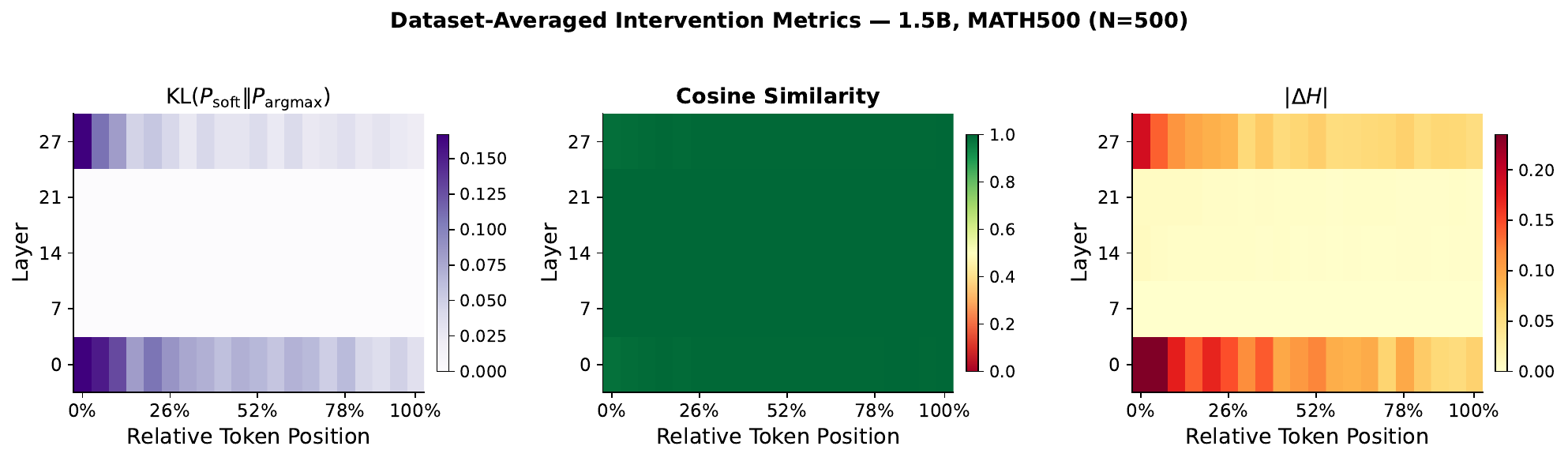}
    \caption{Dataset-averaged intervention metrics for \textbf{Qwen2-1.5B on MATH500} (N=500).}
    \label{fig:avg-heatmap-1.5b-math500}
\end{figure}

\begin{figure}[htbp]
    \centering
    \includegraphics[width=\linewidth]{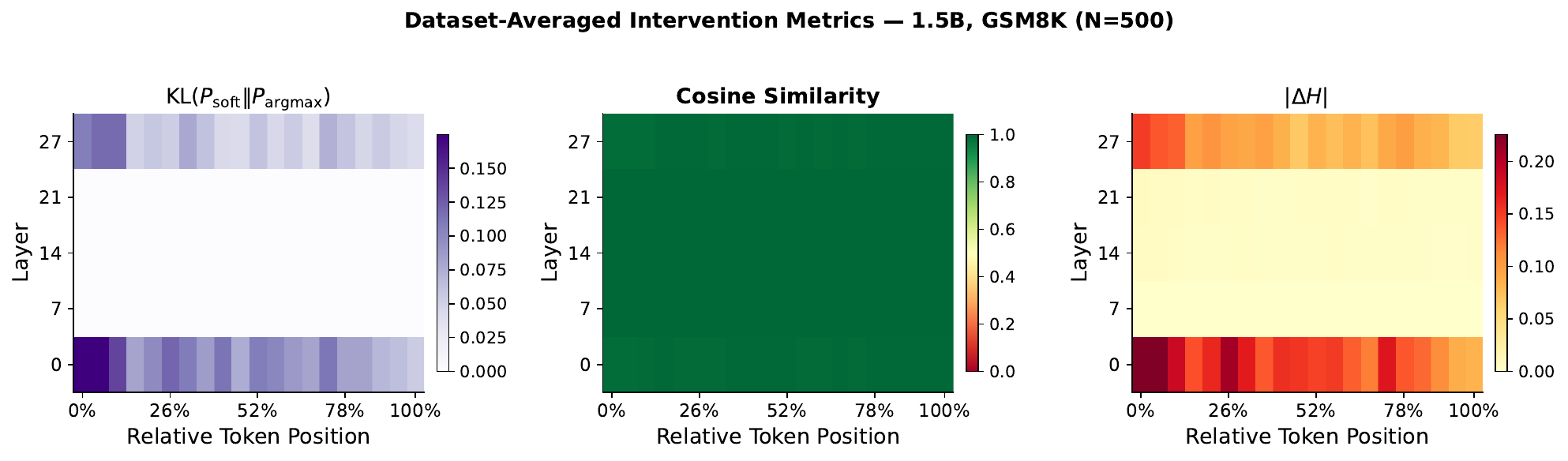}
    \caption{Dataset-averaged intervention metrics for \textbf{Qwen2-1.5B on GSM8K} (N=500).}
    \label{fig:avg-heatmap-1.5b-gsm8k}
\end{figure}

\begin{figure}[htbp]
    \centering
    \includegraphics[width=\linewidth]{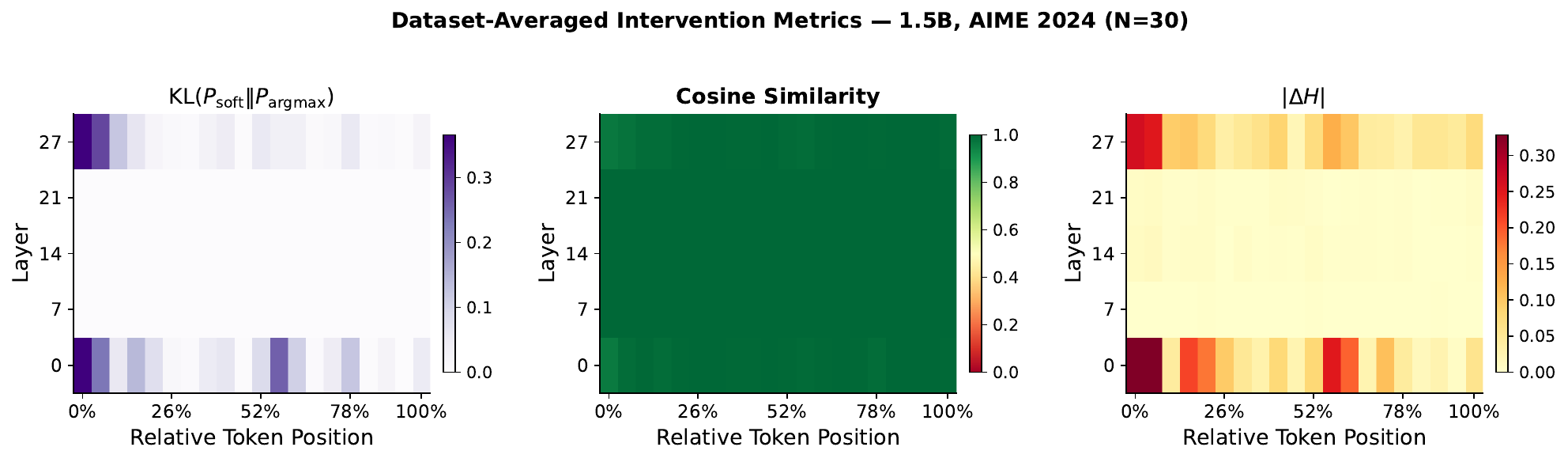}
    \caption{Dataset-averaged intervention metrics for \textbf{Qwen2-1.5B on AIME 2024} (N=30).}
    \label{fig:avg-heatmap-1.5b-aime}
\end{figure}

\begin{figure}[htbp]
    \centering
    \includegraphics[width=\linewidth]{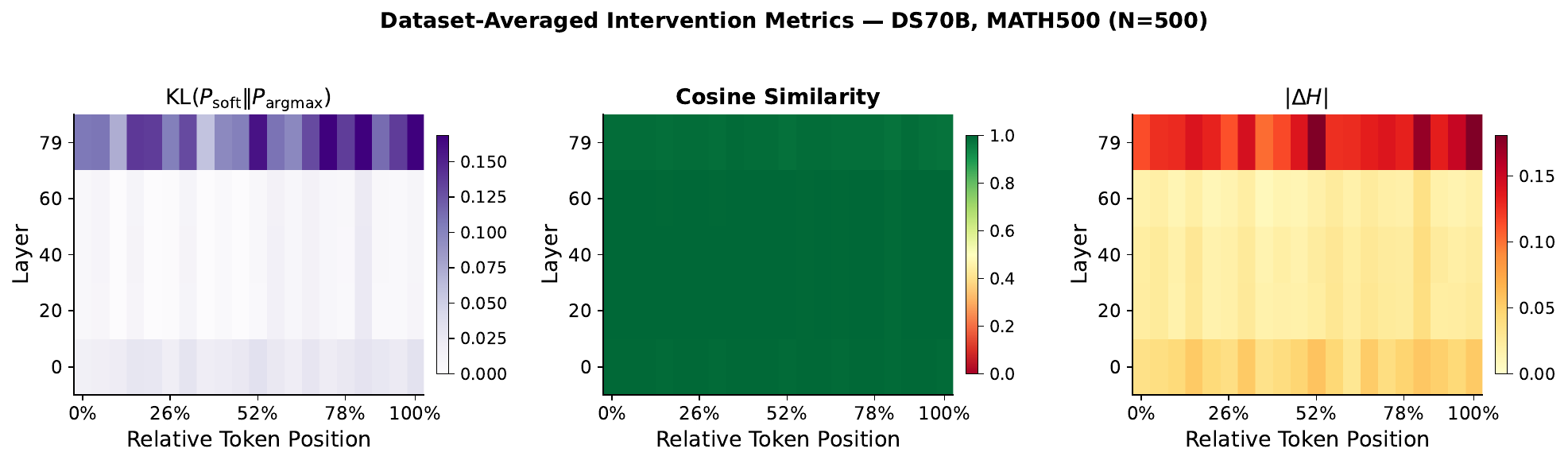}
    \caption{Dataset-averaged intervention metrics for \textbf{DeepSeek-R1-Distill-Llama-70B on MATH500} (N=500).}
    \label{fig:avg-heatmap-ds70b-math500}
\end{figure}

\begin{figure}[htbp]
    \centering
    \includegraphics[width=\linewidth]{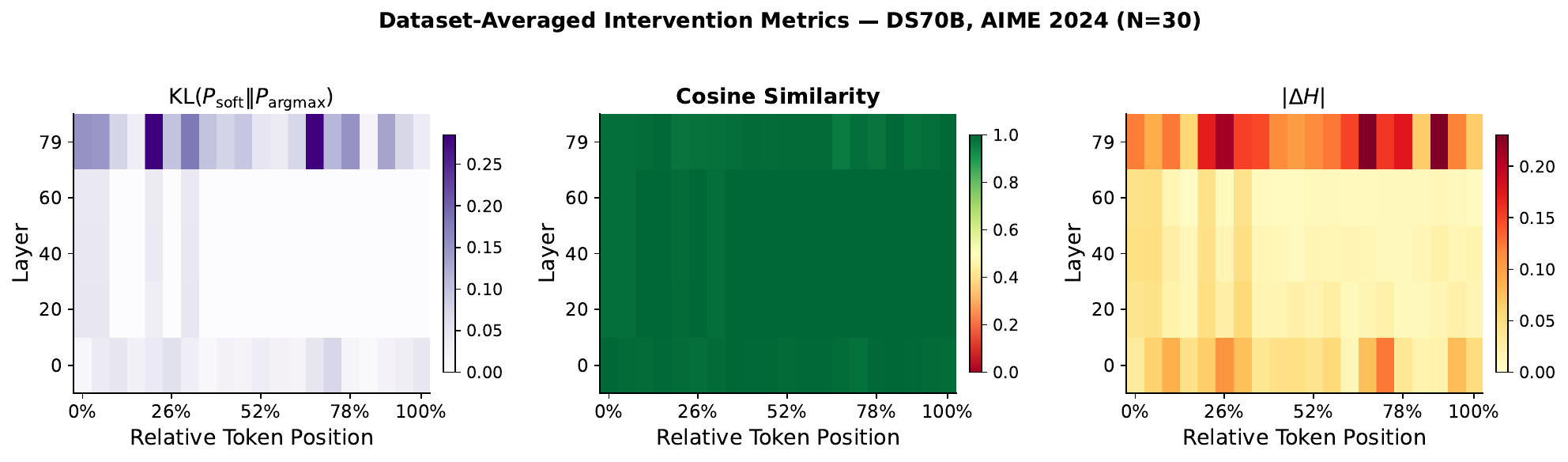}
    \caption{Dataset-averaged intervention metrics for \textbf{DeepSeek-R1-Distill-Llama-70B on AIME 2024} (N=30).}
    \label{fig:avg-heatmap-ds70b-aime}
\end{figure}

\begin{figure}[htbp]
    \centering
    \begin{subfigure}[b]{0.48\textwidth}
        \includegraphics[width=\textwidth]{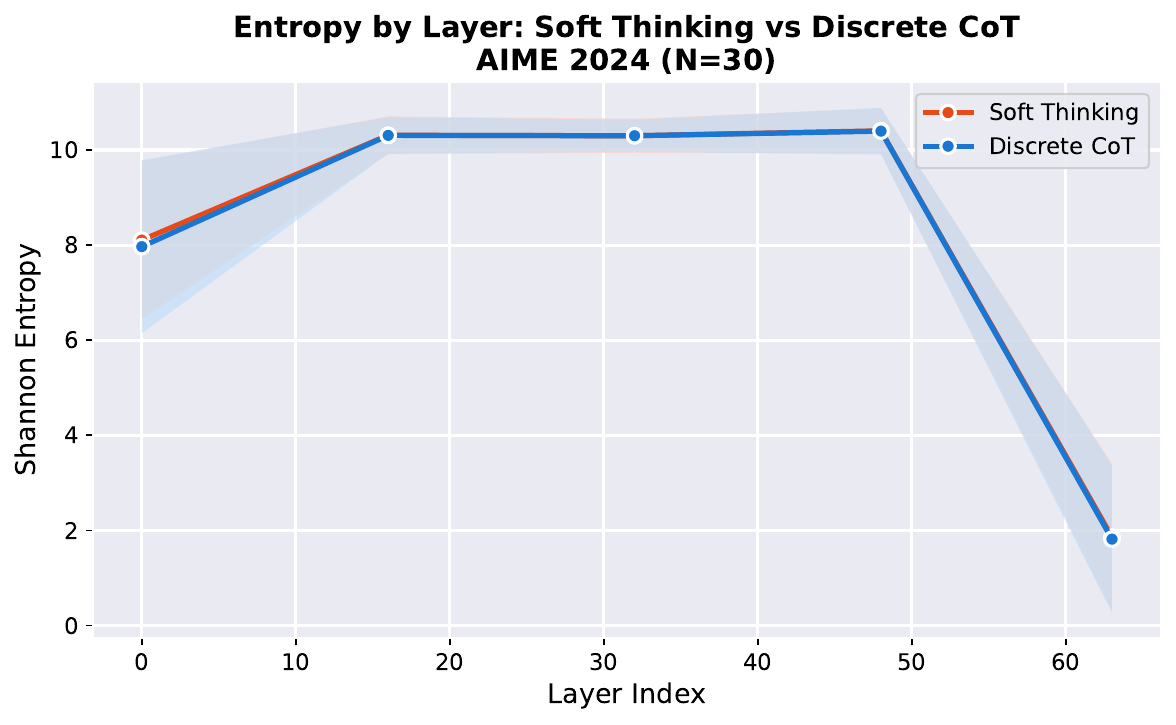}
        \caption{Entropy comparison}
    \end{subfigure}
    \hfill
    \begin{subfigure}[b]{0.48\textwidth}
        \includegraphics[width=\textwidth]{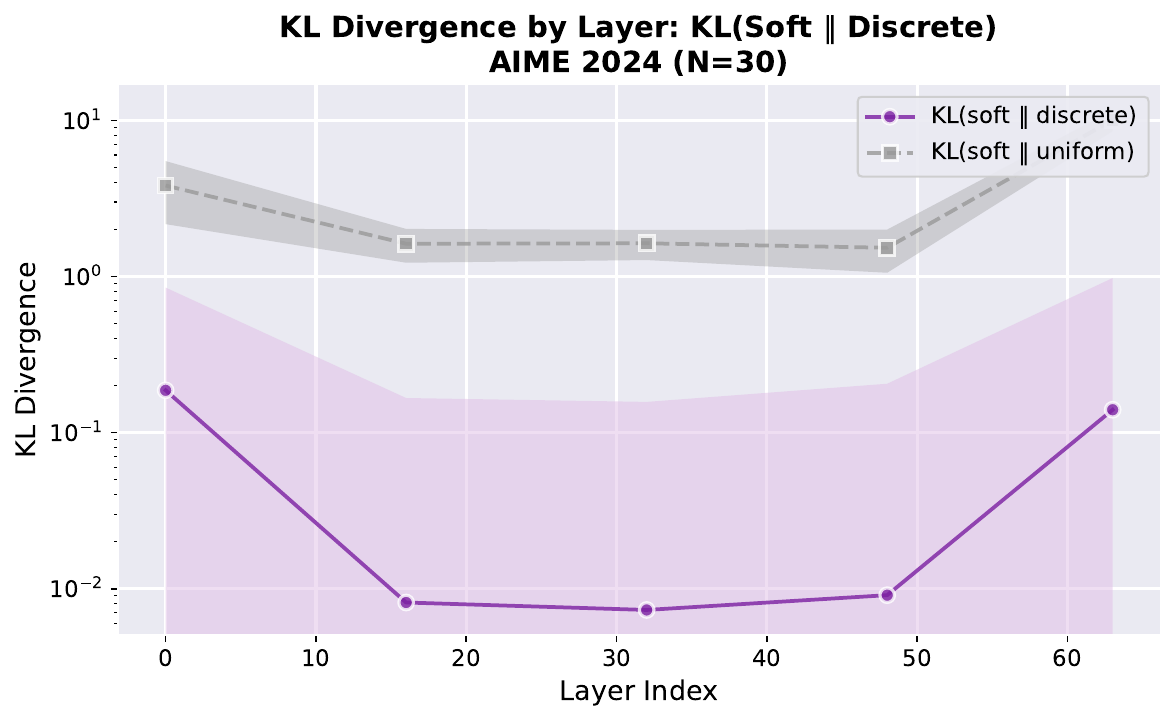}
        \caption{KL divergence by layer}
    \end{subfigure}
    \caption{\textbf{QwQ-32B on AIME2024 (N=30).} Entropy and KL divergence by layer.}
    \label{fig:fulldataset-32b-aime}
\end{figure}

\begin{figure}[htbp]
    \centering
    \begin{subfigure}[b]{0.48\textwidth}
        \includegraphics[width=\textwidth]{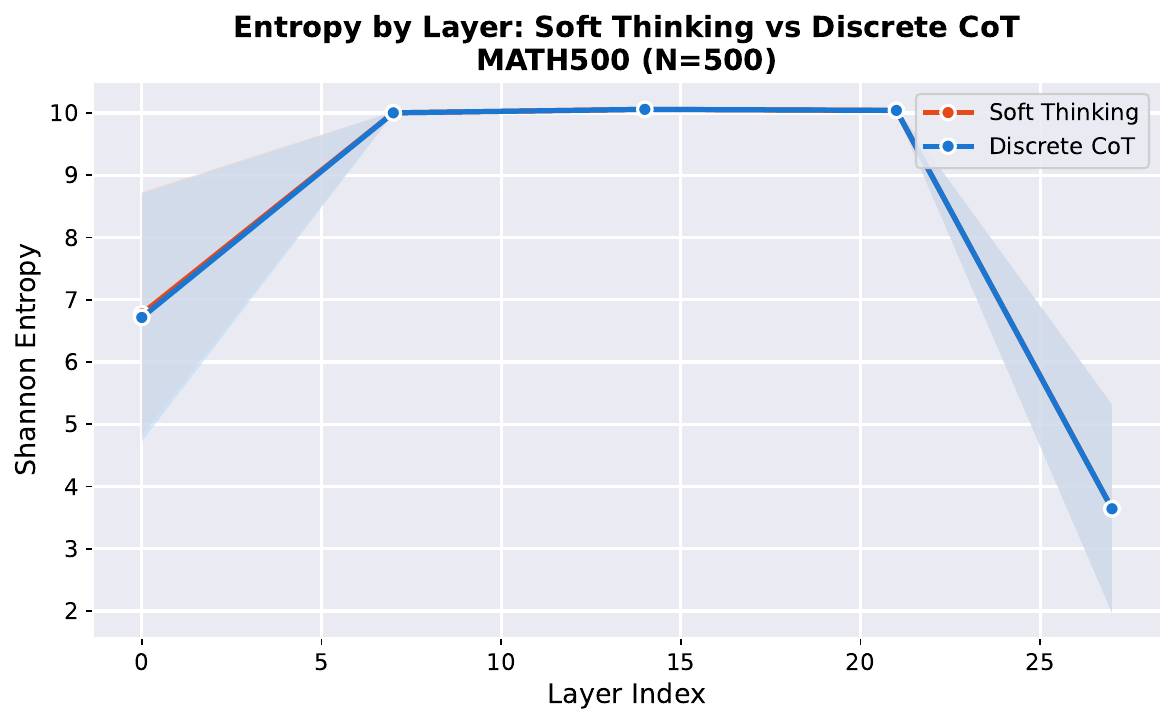}
        \caption{Entropy comparison}
    \end{subfigure}
    \hfill
    \begin{subfigure}[b]{0.48\textwidth}
        \includegraphics[width=\textwidth]{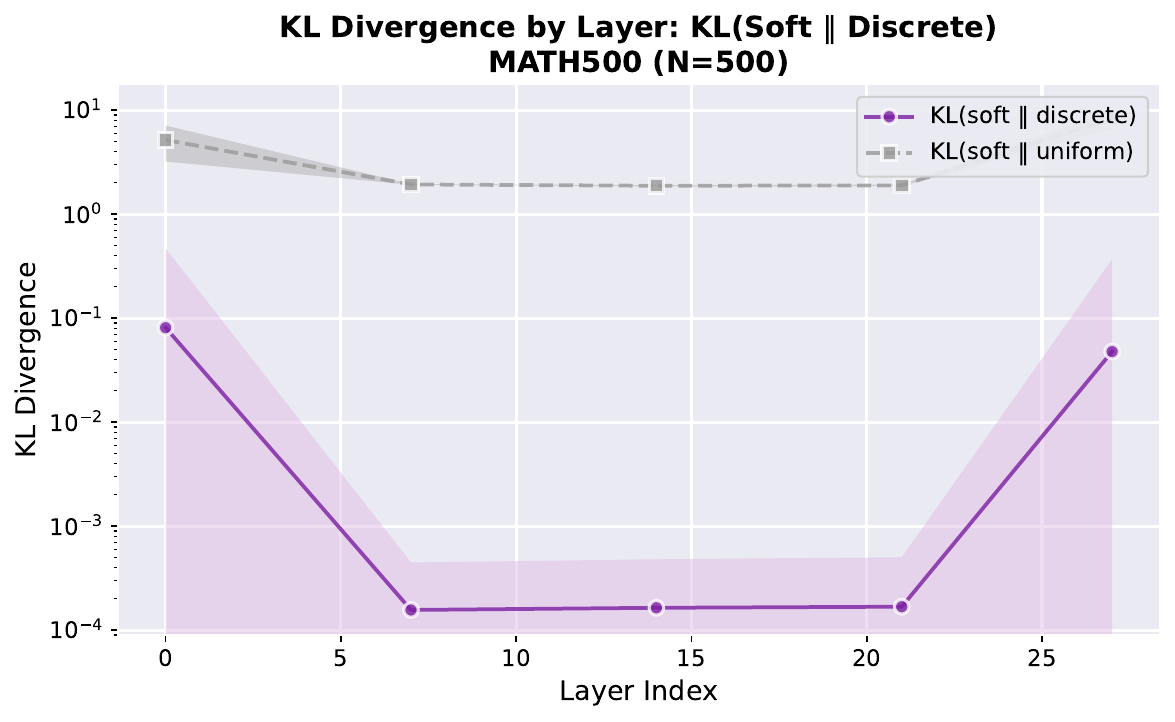}
        \caption{KL divergence by layer}
    \end{subfigure}
    \caption{\textbf{Qwen2-1.5B on MATH500 (N=500).} Entropy and KL divergence by layer.}
    \label{fig:fulldataset-1.5b-math500}
\end{figure}

\begin{figure}[htbp]
    \centering
    \begin{subfigure}[b]{0.48\textwidth}
        \includegraphics[width=\textwidth]{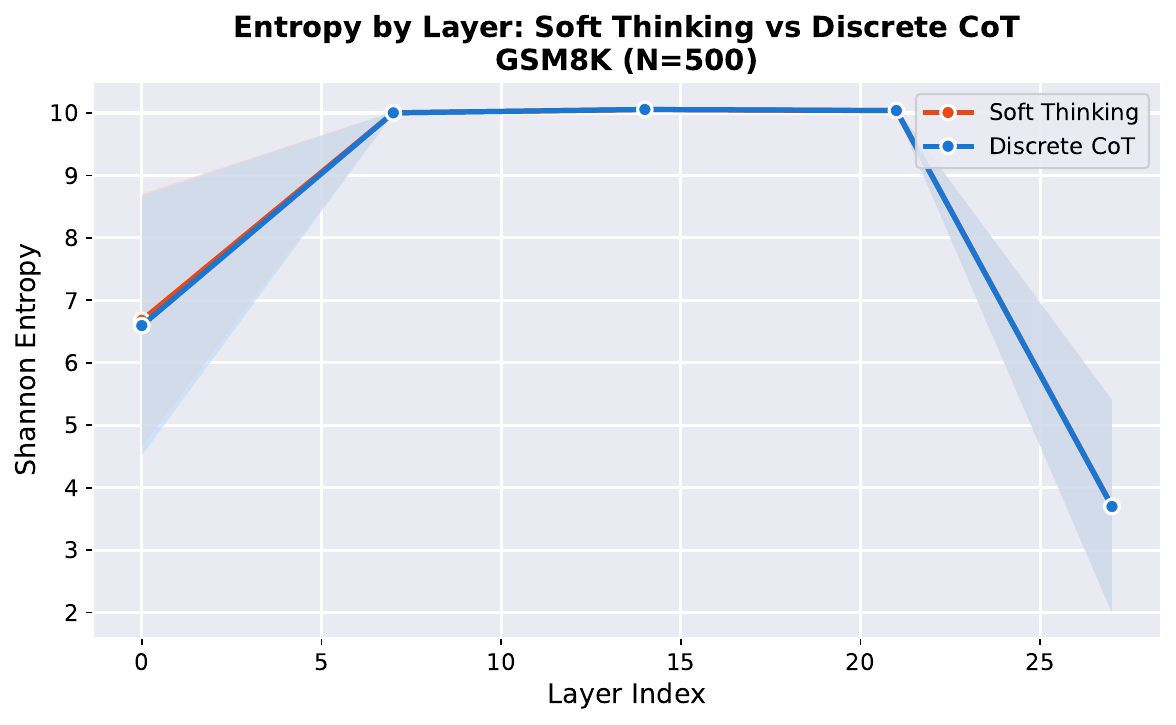}
        \caption{Entropy comparison}
    \end{subfigure}
    \hfill
    \begin{subfigure}[b]{0.48\textwidth}
        \includegraphics[width=\textwidth]{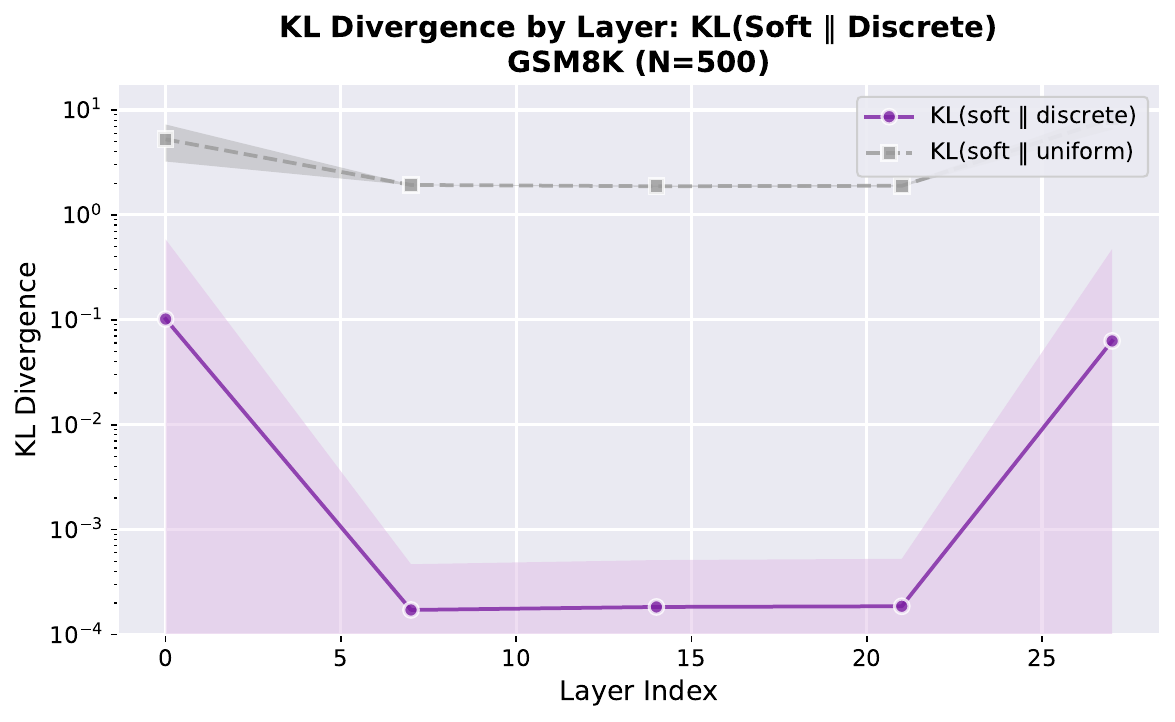}
        \caption{KL divergence by layer}
    \end{subfigure}
    \caption{\textbf{Qwen2-1.5B on GSM8K (N=500).} Entropy and KL divergence by layer.}
    \label{fig:fulldataset-1.5b-gsm8k}
\end{figure}

\begin{figure}[htbp]
    \centering
    \begin{subfigure}[b]{0.48\textwidth}
        \includegraphics[width=\textwidth]{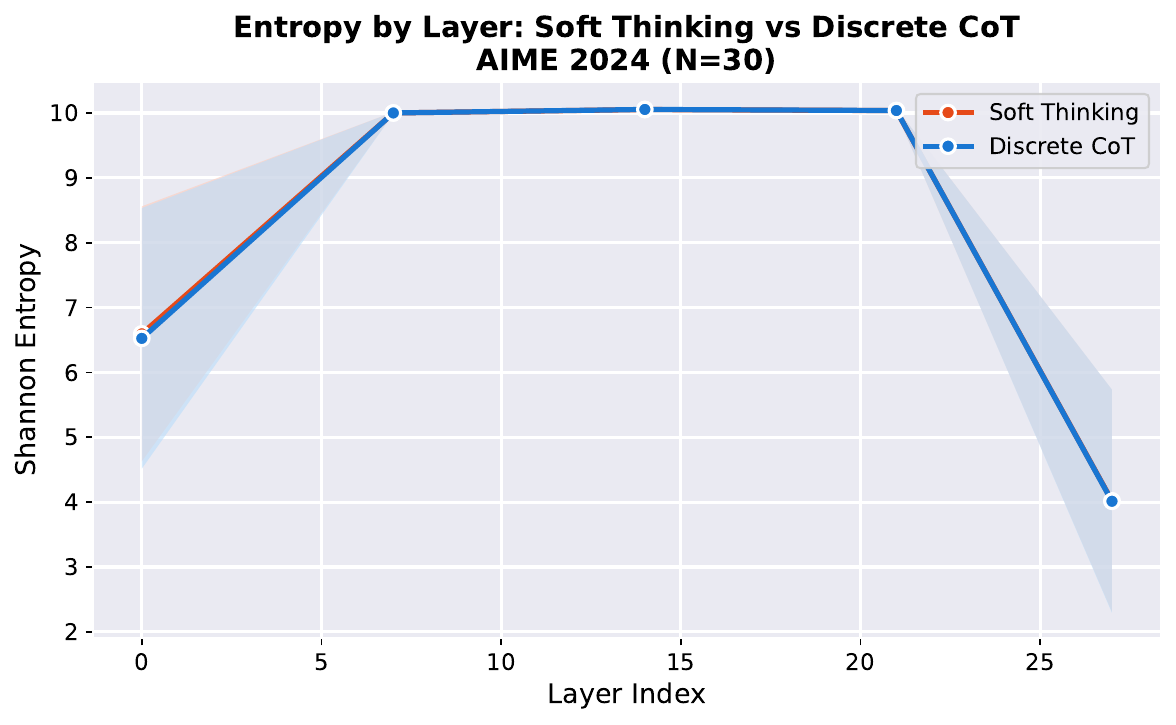}
        \caption{Entropy comparison}
    \end{subfigure}
    \hfill
    \begin{subfigure}[b]{0.48\textwidth}
        \includegraphics[width=\textwidth]{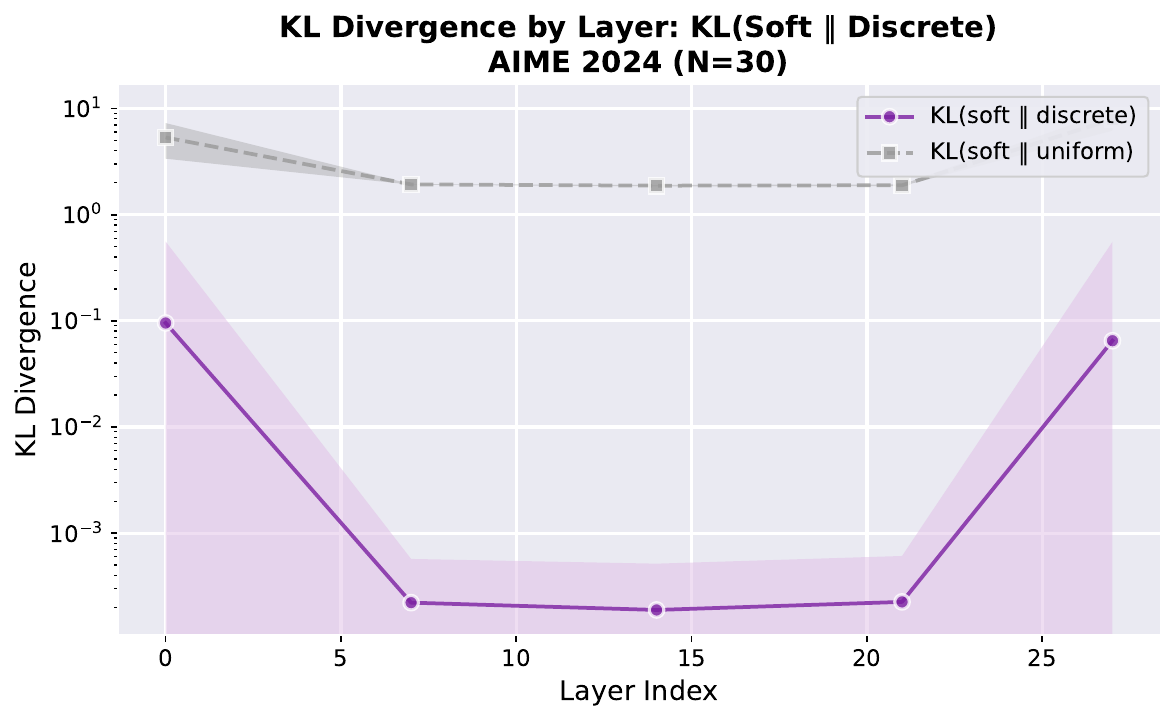}
        \caption{KL divergence by layer}
    \end{subfigure}
    \caption{\textbf{Qwen2-1.5B on AIME2024 (N=30).} Entropy and KL divergence by layer.}
    \label{fig:fulldataset-1.5b-aime}
\end{figure}

\begin{figure}[htbp]
    \centering
    \begin{subfigure}[b]{0.48\textwidth}
        \includegraphics[width=\textwidth]{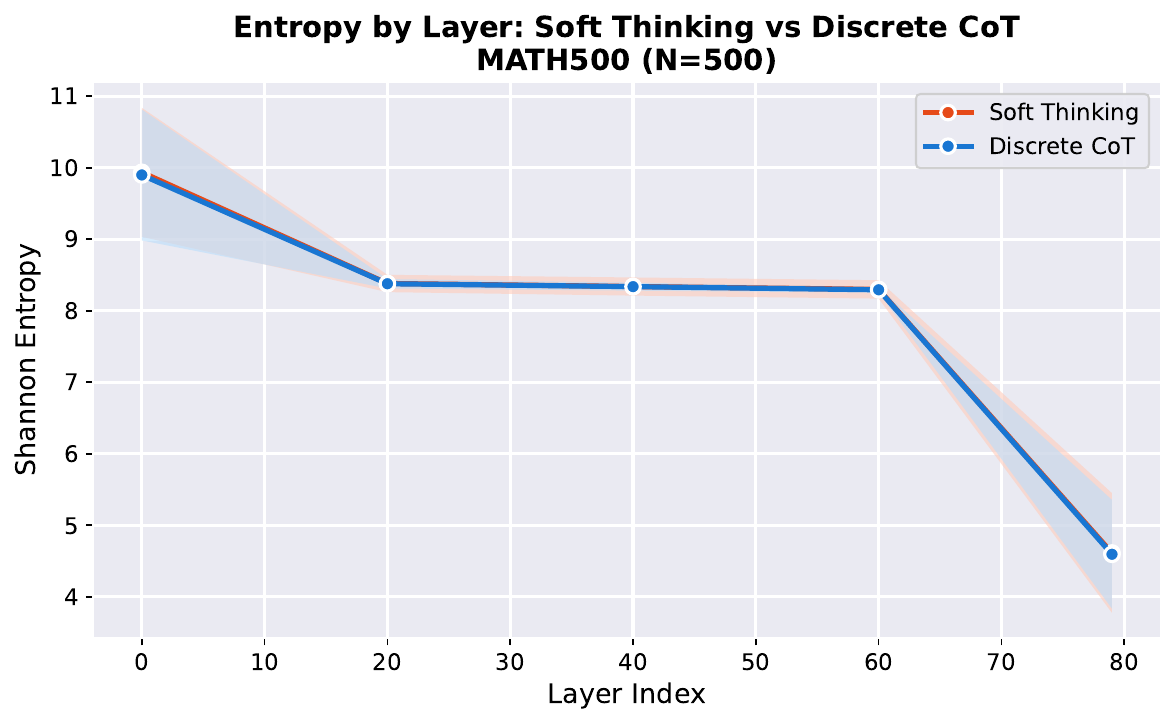}
        \caption{Entropy comparison}
    \end{subfigure}
    \hfill
    \begin{subfigure}[b]{0.48\textwidth}
        \includegraphics[width=\textwidth]{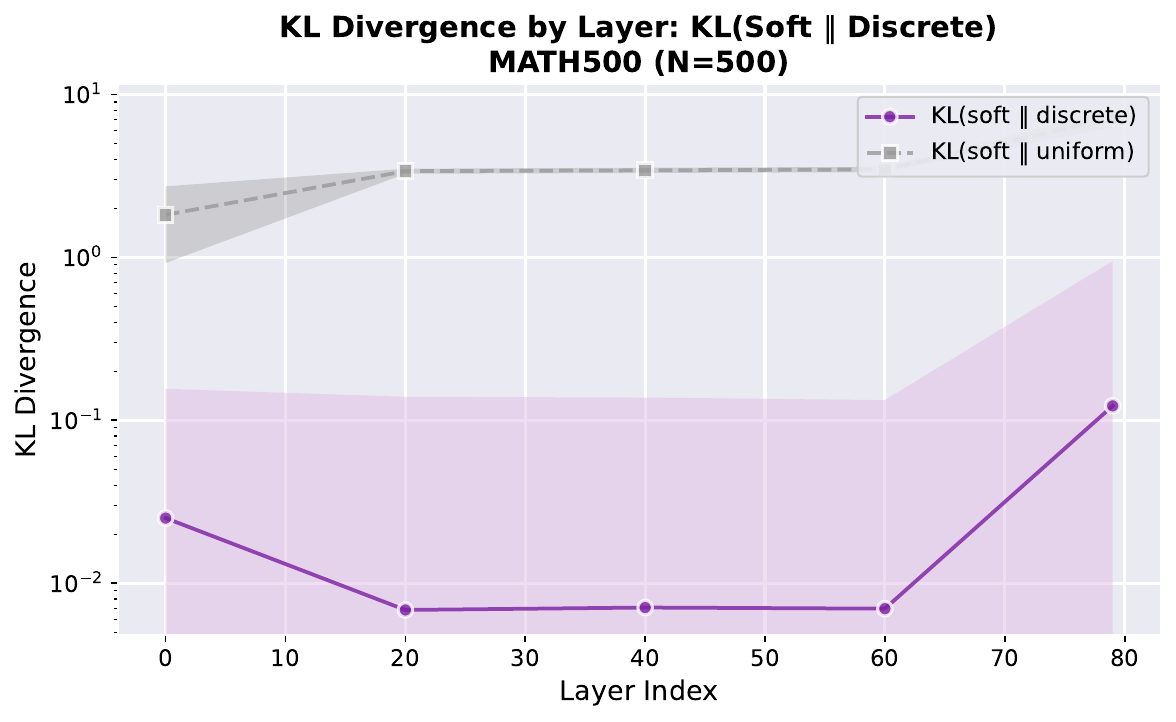}
        \caption{KL divergence by layer}
    \end{subfigure}
    \caption{\textbf{DeepSeek-R1-Distill-Llama-70B on MATH500 (N=500).} Entropy and KL divergence by layer. \logitlens is applied at 5 evenly spaced layers $\{0, 20, 40, 60, 79\}$ of the 80-layer model.}
    \label{fig:fulldataset-ds70b-math500}
\end{figure}

\begin{figure}[htbp]
    \centering
    \begin{subfigure}[b]{0.48\textwidth}
        \includegraphics[width=\textwidth]{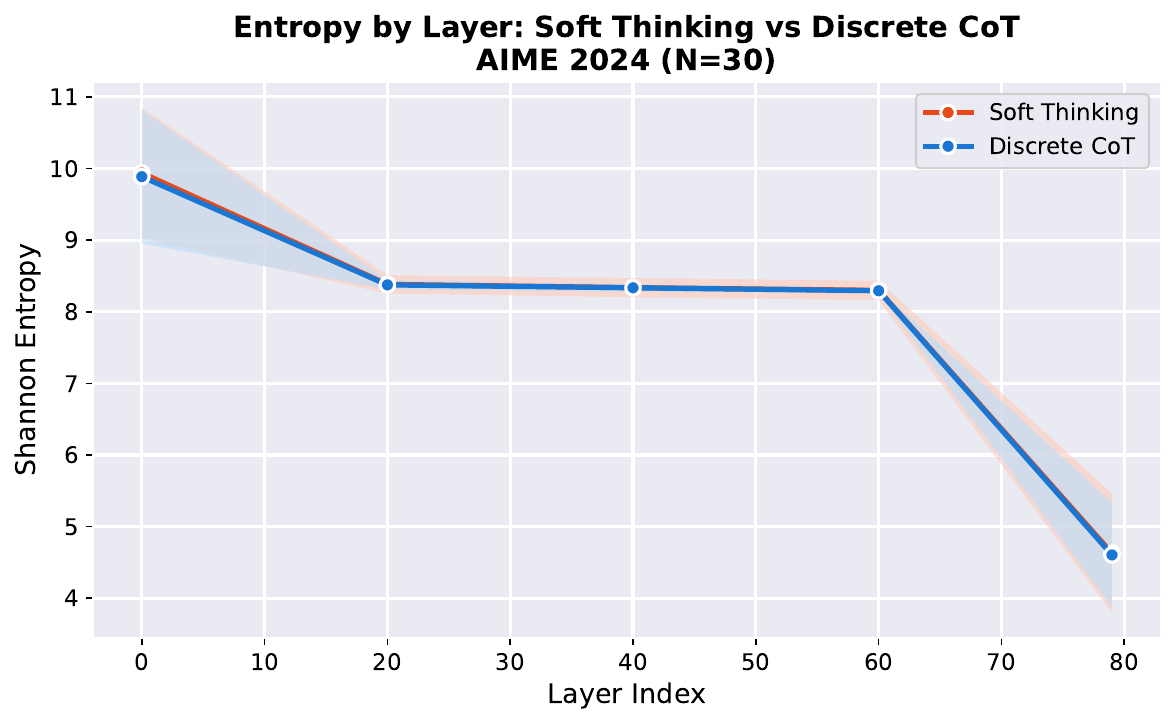}
        \caption{Entropy comparison}
    \end{subfigure}
    \hfill
    \begin{subfigure}[b]{0.48\textwidth}
        \includegraphics[width=\textwidth]{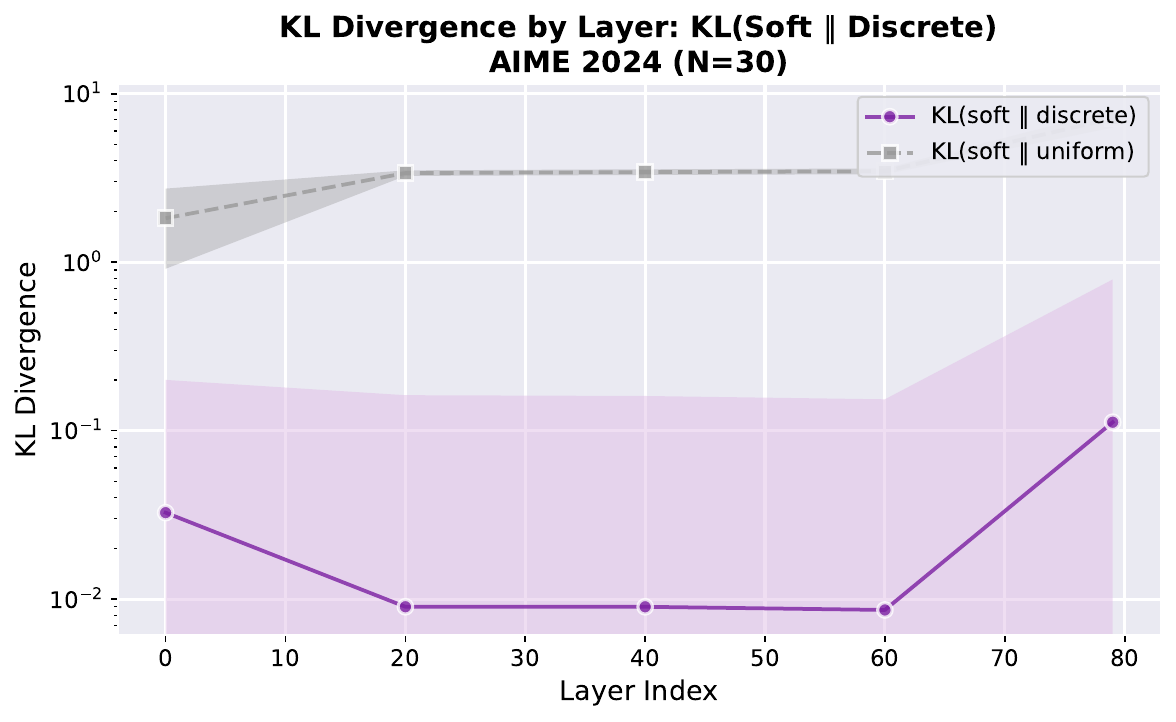}
        \caption{KL divergence by layer}
    \end{subfigure}
    \caption{\textbf{DeepSeek-R1-Distill-Llama-70B on AIME2024 (N=30).} Entropy and KL divergence by layer. Same 5-layer probe as \Cref{fig:fulldataset-ds70b-math500}; KL remains near zero from layer 20 onward.}
    \label{fig:fulldataset-ds70b-aime}
\end{figure}

\begin{figure}[htbp]
    \centering
    \includegraphics[width=\linewidth]{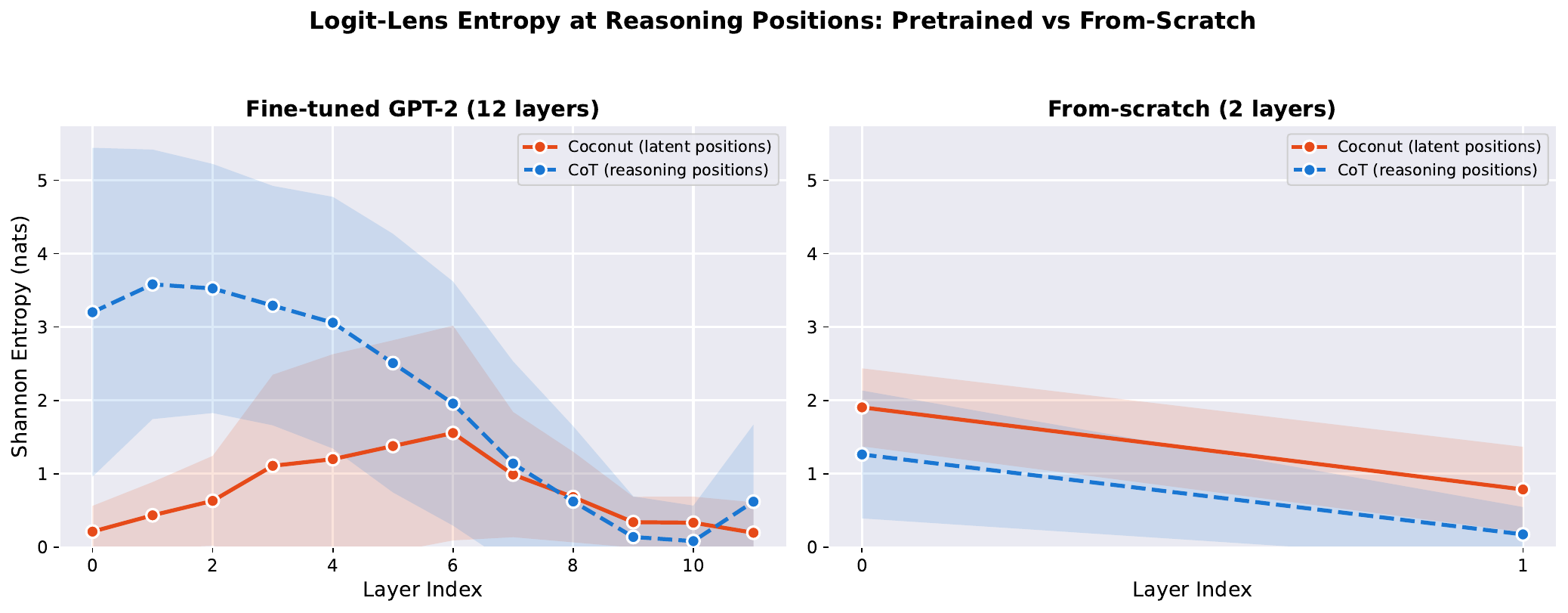}
    \caption{\logitlens entropy at reasoning positions for fine-tuned GPT-2 (left, 12 layers) and a from-scratch 2-layer model (right). In the fine-tuned model, Coconut latent positions maintain uniformly low entropy across all layers, while CoT positions show high early-layer entropy that resolves by layers 8--9. In the from-scratch model, Coconut latent positions retain higher entropy than their CoT counterparts, consistent with the from-scratch model maintaining richer latent representations. Shaded regions denote $\pm 1$ standard deviation across examples.}
    \label{fig:latent-collapse-comparison}
\end{figure}

\subsection{Uniform Superposition: Entropy and KL Divergence}
\label{app:uniform-superposition}

To test whether the collapse observed with softmax-weighted Soft Thinking in \Cref{subsec:entropy_profile} is an artifact of the probability-weighted embedding concentrating mass near the argmax token, we repeat the entropy profile and KL divergence experiments with \emph{uniform} weighting: each of the top-$k$ tokens receives equal weight $1/k$, giving the maximally diverse input embedding.

\begin{figure}[h]
    \centering
    \begin{subfigure}[b]{0.48\textwidth}
        \includegraphics[width=\textwidth]{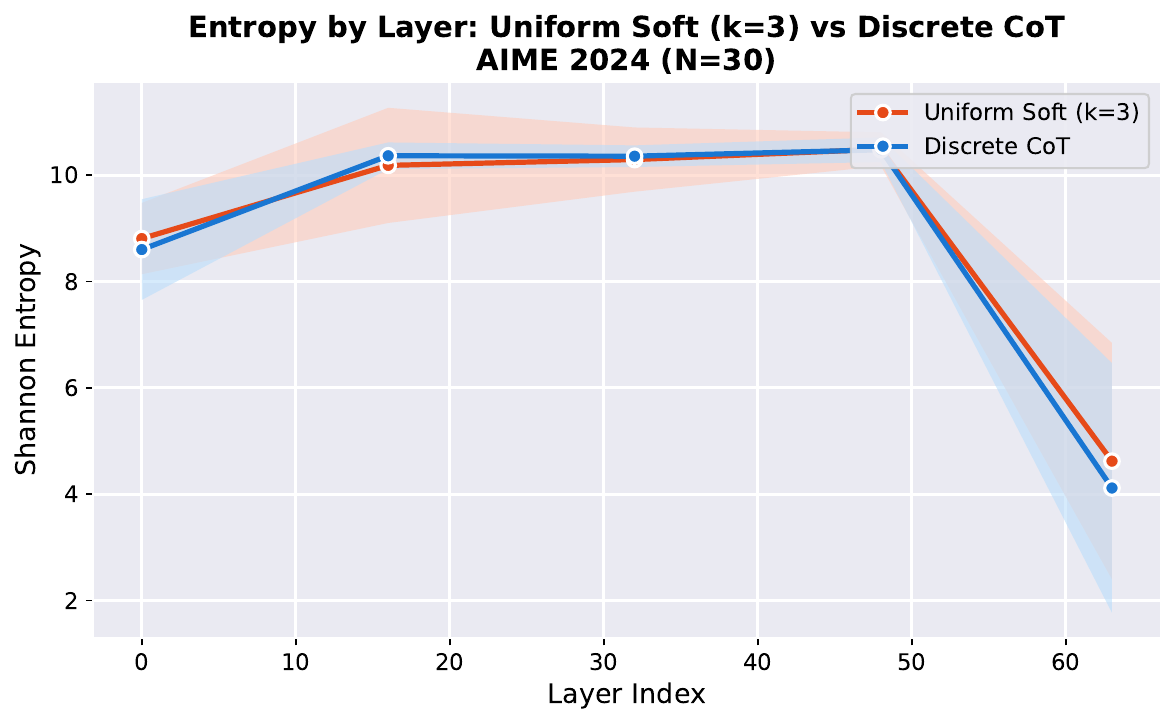}
        \caption{Entropy — AIME 2024}
    \end{subfigure}
    \hfill
    \begin{subfigure}[b]{0.48\textwidth}
        \includegraphics[width=\textwidth]{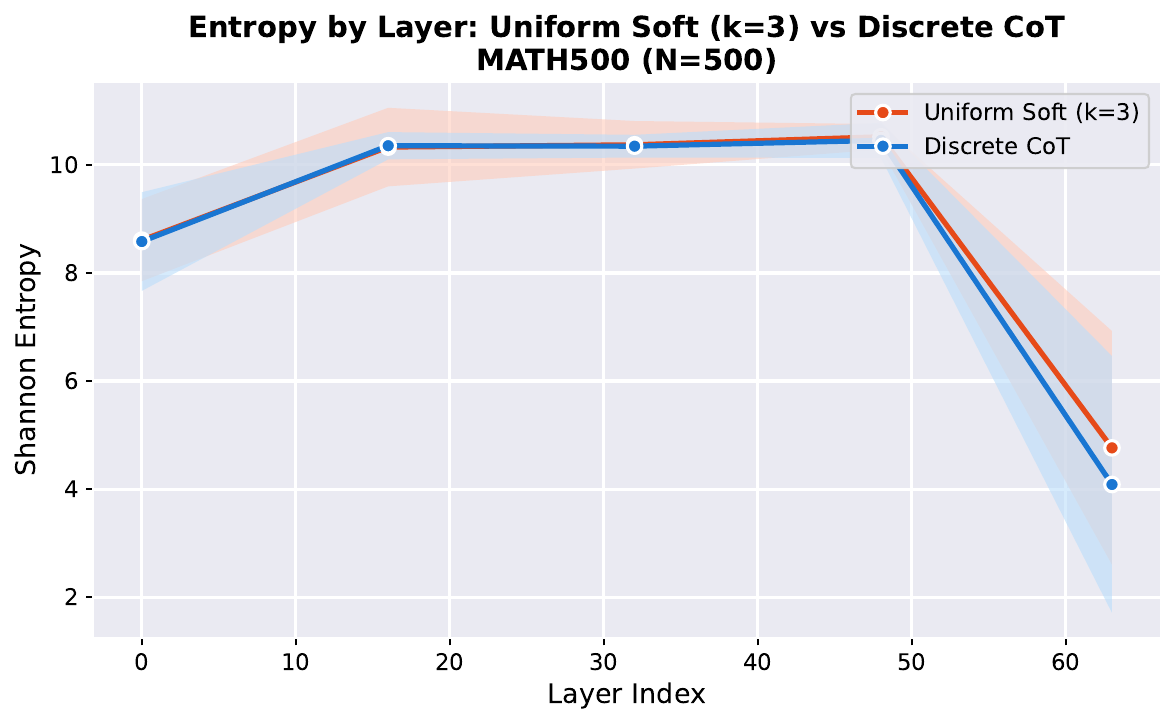}
        \caption{Entropy — MATH500}
    \end{subfigure}

    \vspace{0.5em}

    \begin{subfigure}[b]{0.48\textwidth}
        \includegraphics[width=\textwidth]{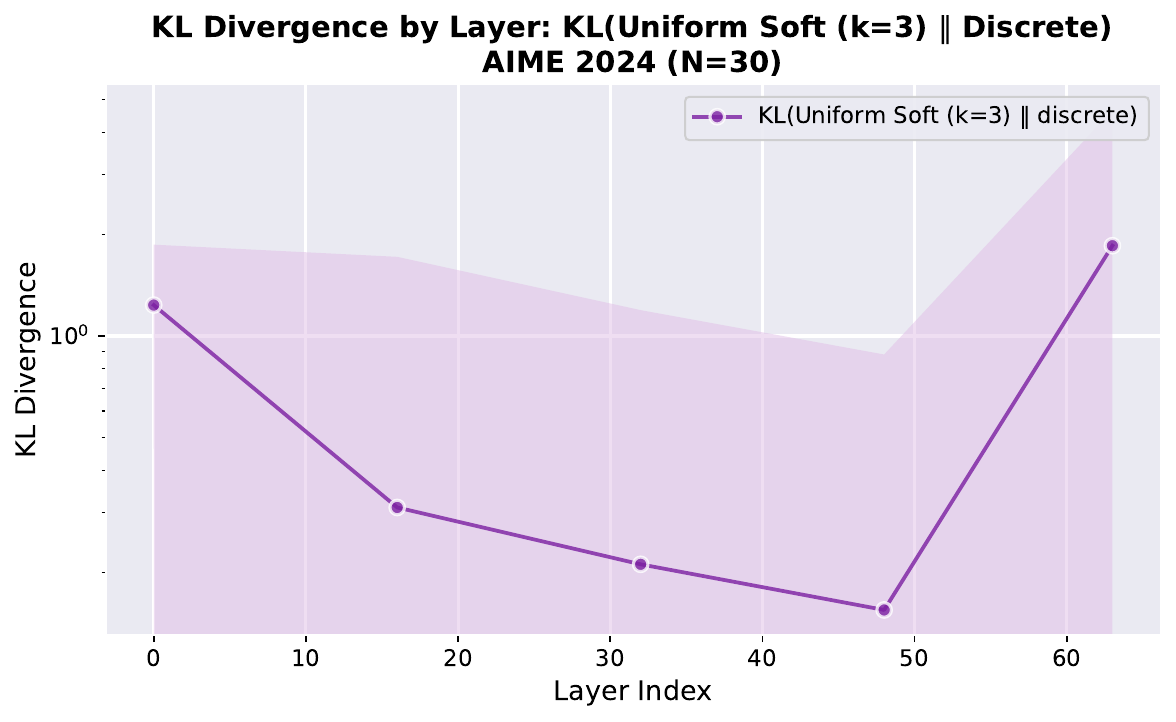}
        \caption{KL divergence — AIME 2024}
    \end{subfigure}
    \hfill
    \begin{subfigure}[b]{0.48\textwidth}
        \includegraphics[width=\textwidth]{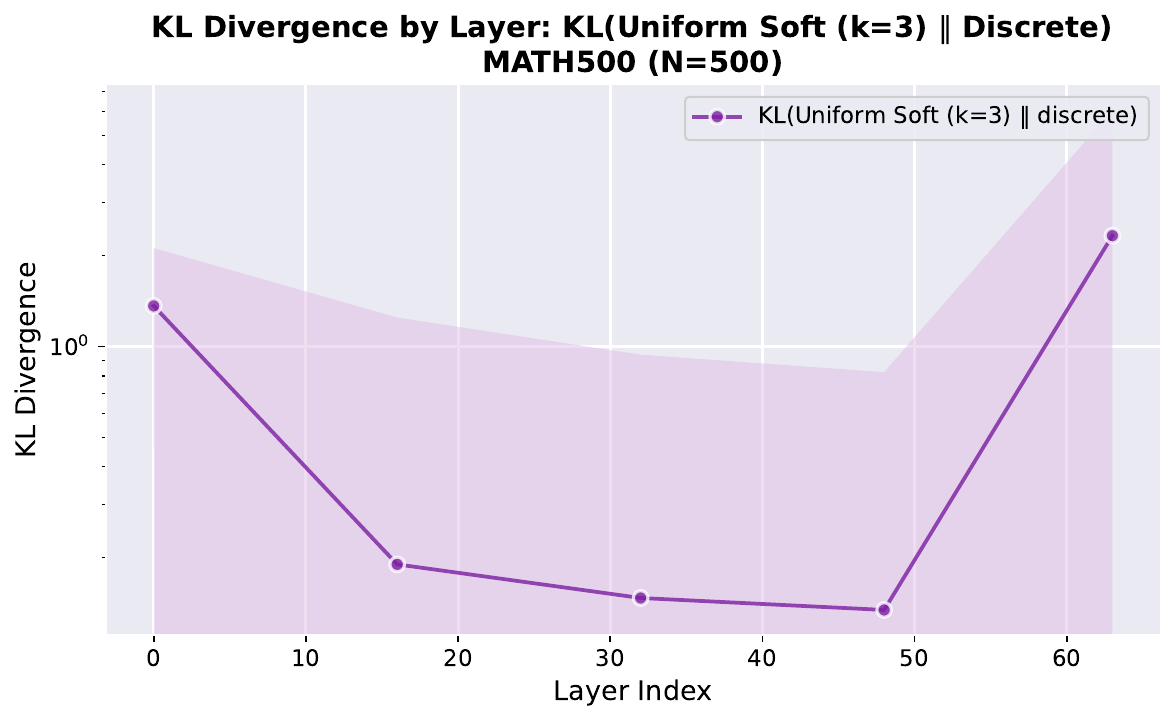}
        \caption{KL divergence — MATH500}
    \end{subfigure}
    \caption{\textbf{QwQ-32B, uniform weighting $k=3$.} Entropy profiles (top) and KL divergence (bottom) for AIME 2024 (left) and MATH500 (right). Despite uniform weights giving equal mass to 3 tokens, representations collapse to near-argmax in middle layers on both datasets.}
    \label{fig:uniform-32b-k3}
\end{figure}

\begin{figure}[h]
    \centering
    \begin{subfigure}[b]{0.48\textwidth}
        \includegraphics[width=\textwidth]{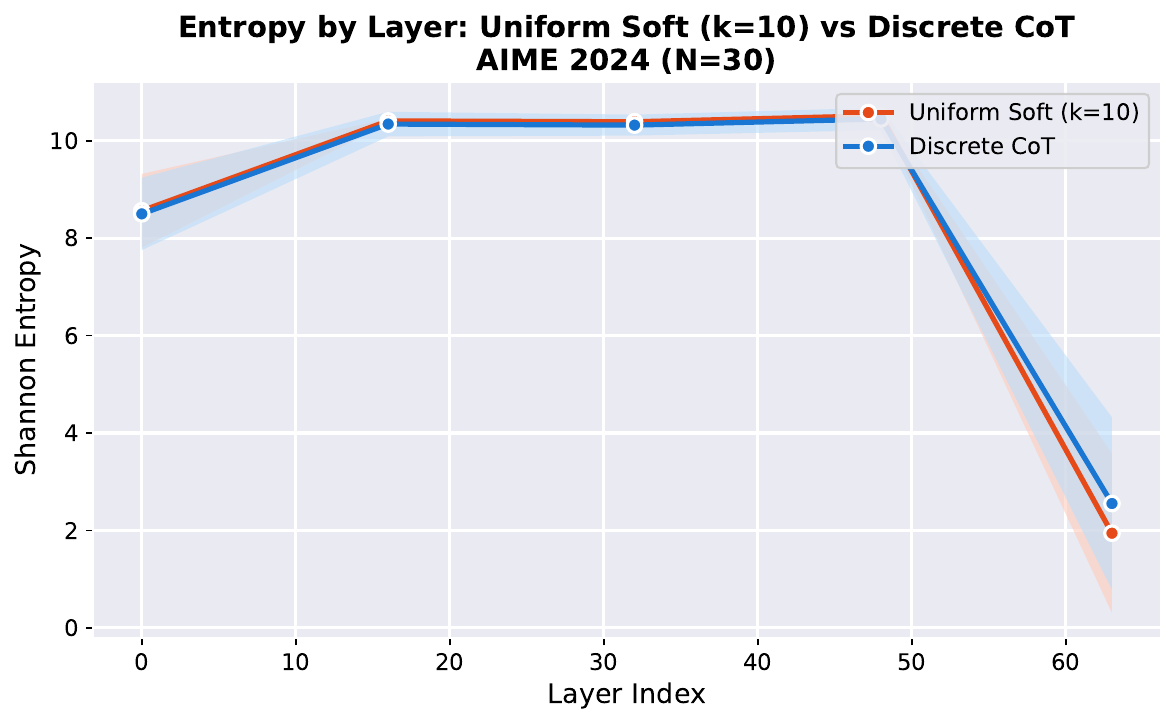}
        \caption{Entropy — AIME 2024}
    \end{subfigure}
    \hfill
    \begin{subfigure}[b]{0.48\textwidth}
        \includegraphics[width=\textwidth]{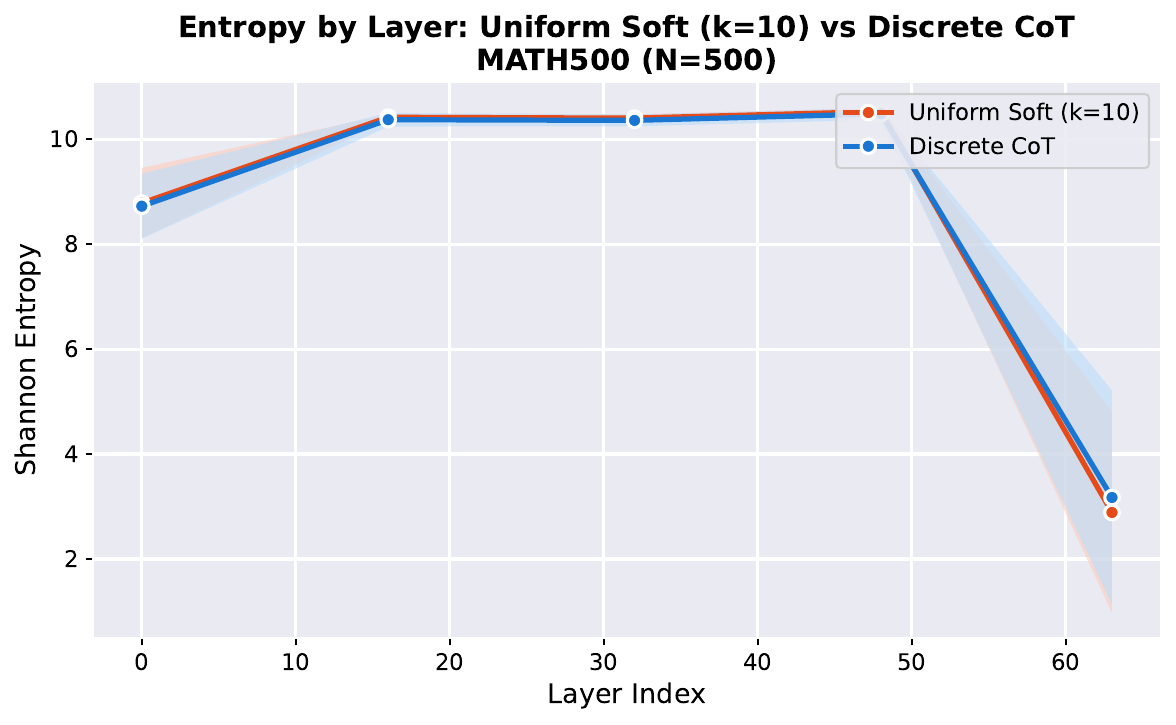}
        \caption{Entropy — MATH500}
    \end{subfigure}

    \vspace{0.5em}

    \begin{subfigure}[b]{0.48\textwidth}
        \includegraphics[width=\textwidth]{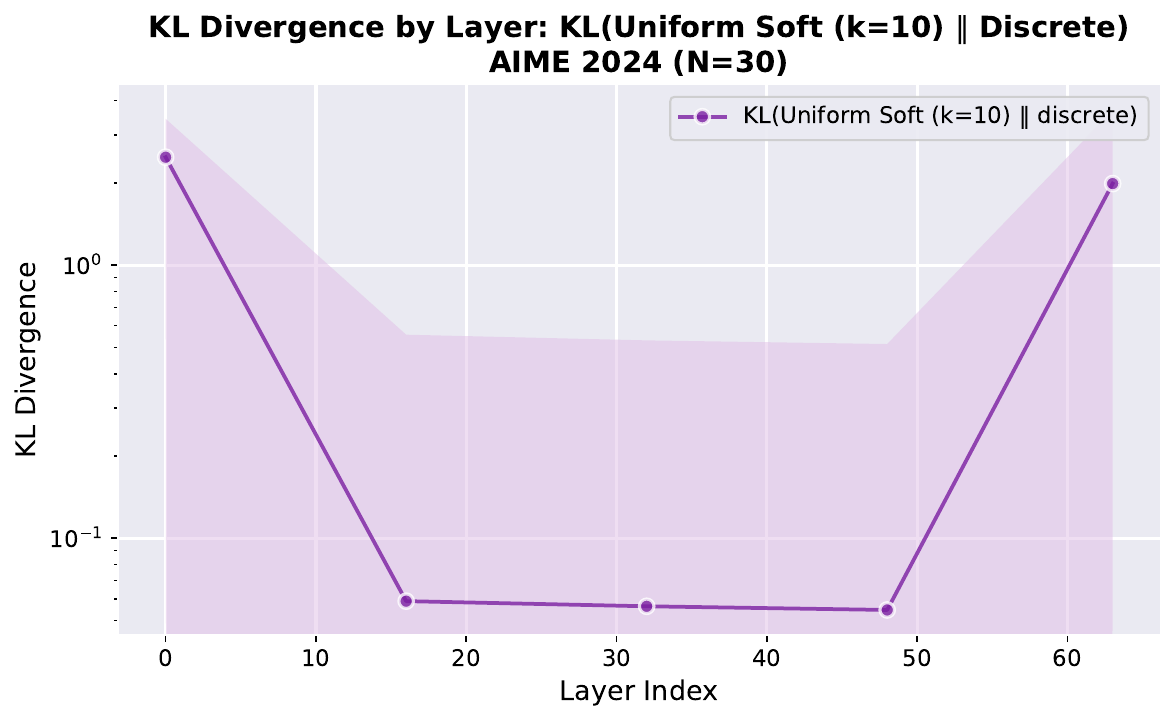}
        \caption{KL divergence — AIME 2024}
    \end{subfigure}
    \hfill
    \begin{subfigure}[b]{0.48\textwidth}
        \includegraphics[width=\textwidth]{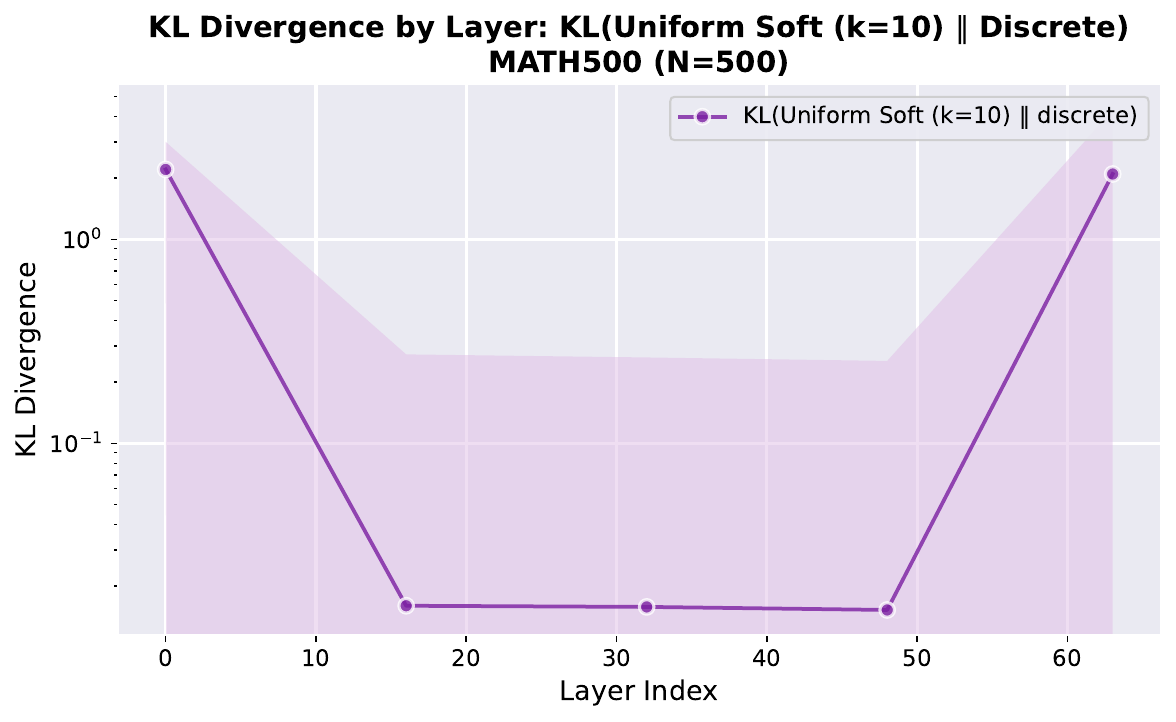}
        \caption{KL divergence — MATH500}
    \end{subfigure}
    \caption{\textbf{QwQ-32B, uniform weighting $k=10$.} Entropy profiles (top) and KL divergence (bottom) for AIME 2024 (left) and MATH500 (right).}
    \label{fig:uniform-32b-k10}
\end{figure}

\begin{figure}[h]
    \centering
    \begin{subfigure}[b]{0.48\textwidth}
        \includegraphics[width=\textwidth]{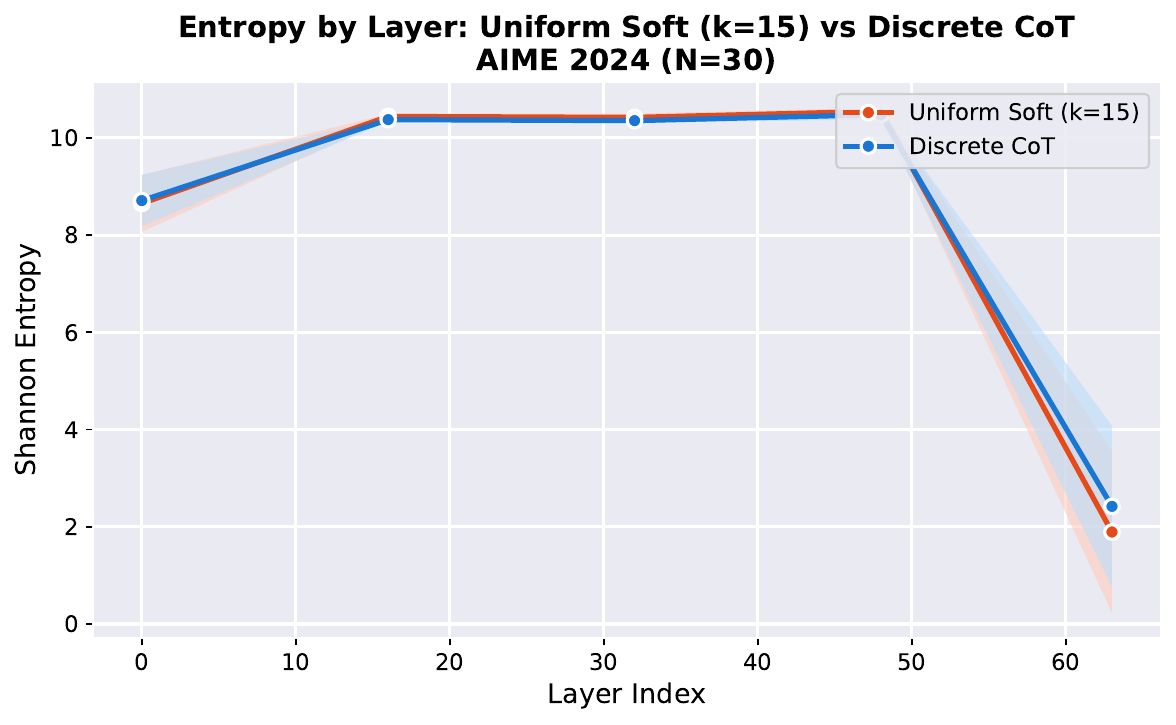}
        \caption{Entropy — AIME 2024}
    \end{subfigure}
    \hfill
    \begin{subfigure}[b]{0.48\textwidth}
        \includegraphics[width=\textwidth]{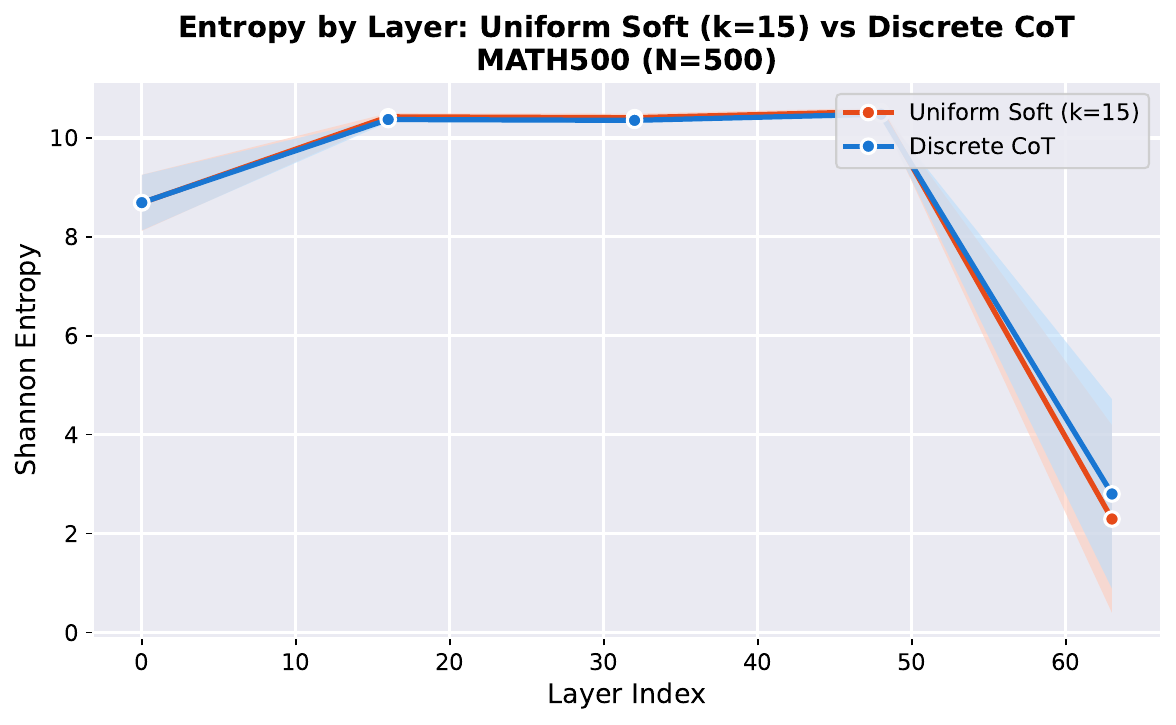}
        \caption{Entropy — MATH500}
    \end{subfigure}

    \vspace{0.5em}

    \begin{subfigure}[b]{0.48\textwidth}
        \includegraphics[width=\textwidth]{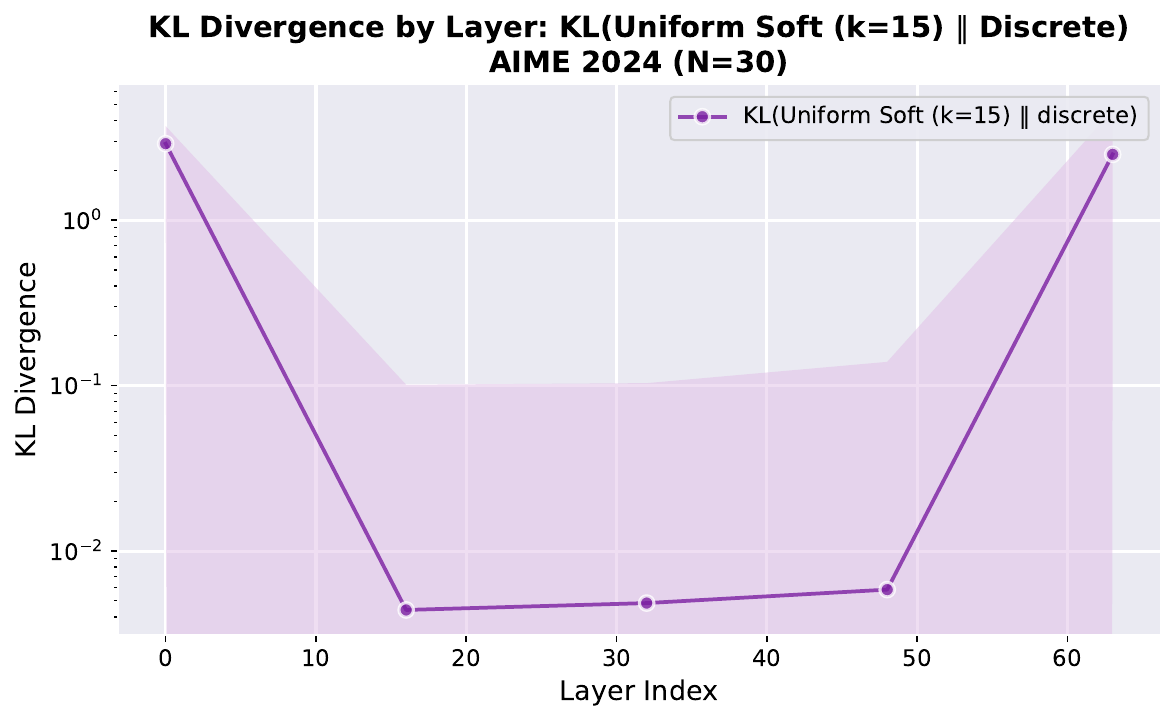}
        \caption{KL divergence — AIME 2024}
    \end{subfigure}
    \hfill
    \begin{subfigure}[b]{0.48\textwidth}
        \includegraphics[width=\textwidth]{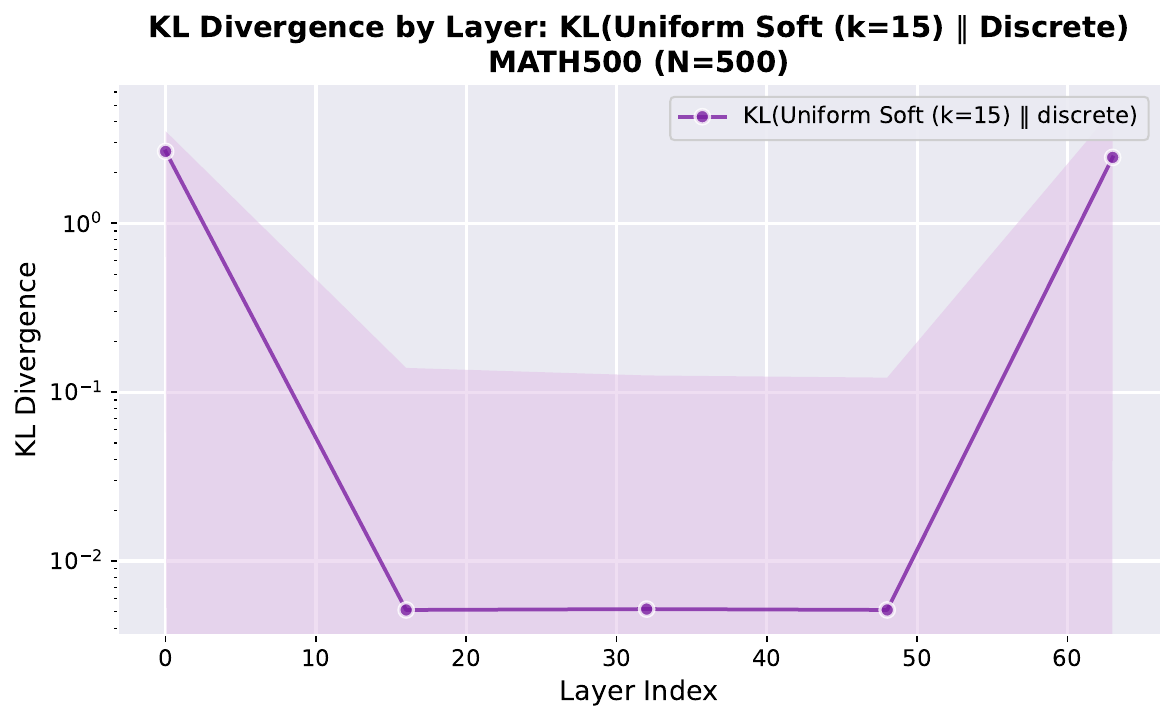}
        \caption{KL divergence — MATH500}
    \end{subfigure}
    \caption{\textbf{QwQ-32B, uniform weighting $k=15$.} Matching the top-$k$ of the standard softmax experiments, uniform weighting still collapses in middle layers. The entropy profiles are indistinguishable from discrete CoT.}
    \label{fig:uniform-32b-k15}
\end{figure}

\begin{figure}[h]
    \centering
    \begin{subfigure}[b]{0.48\textwidth}
        \includegraphics[width=\textwidth]{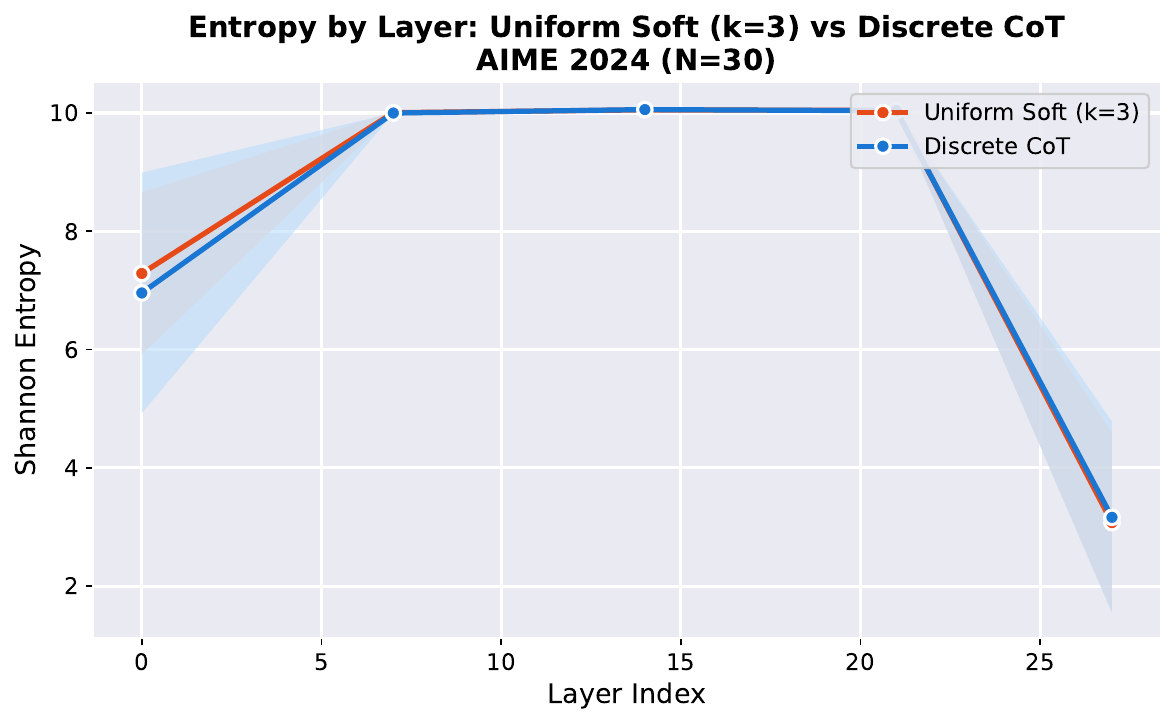}
        \caption{Entropy — AIME 2024}
    \end{subfigure}
    \hfill
    \begin{subfigure}[b]{0.48\textwidth}
        \includegraphics[width=\textwidth]{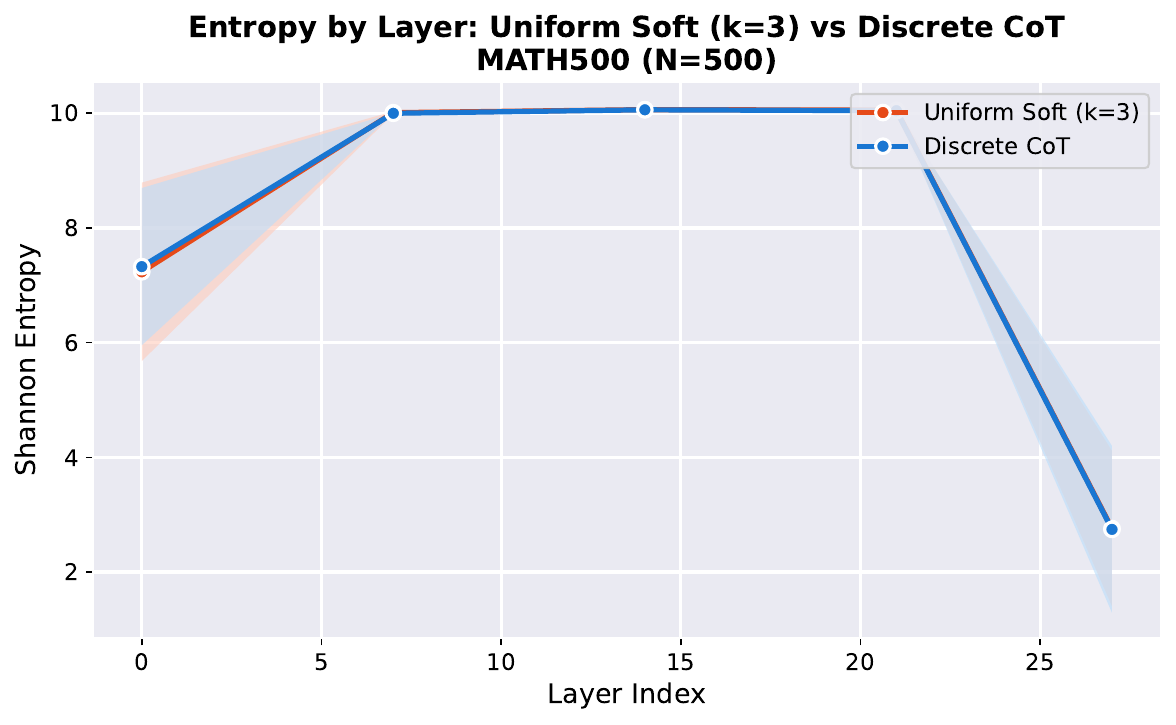}
        \caption{Entropy — MATH500}
    \end{subfigure}

    \vspace{0.5em}

    \begin{subfigure}[b]{0.48\textwidth}
        \includegraphics[width=\textwidth]{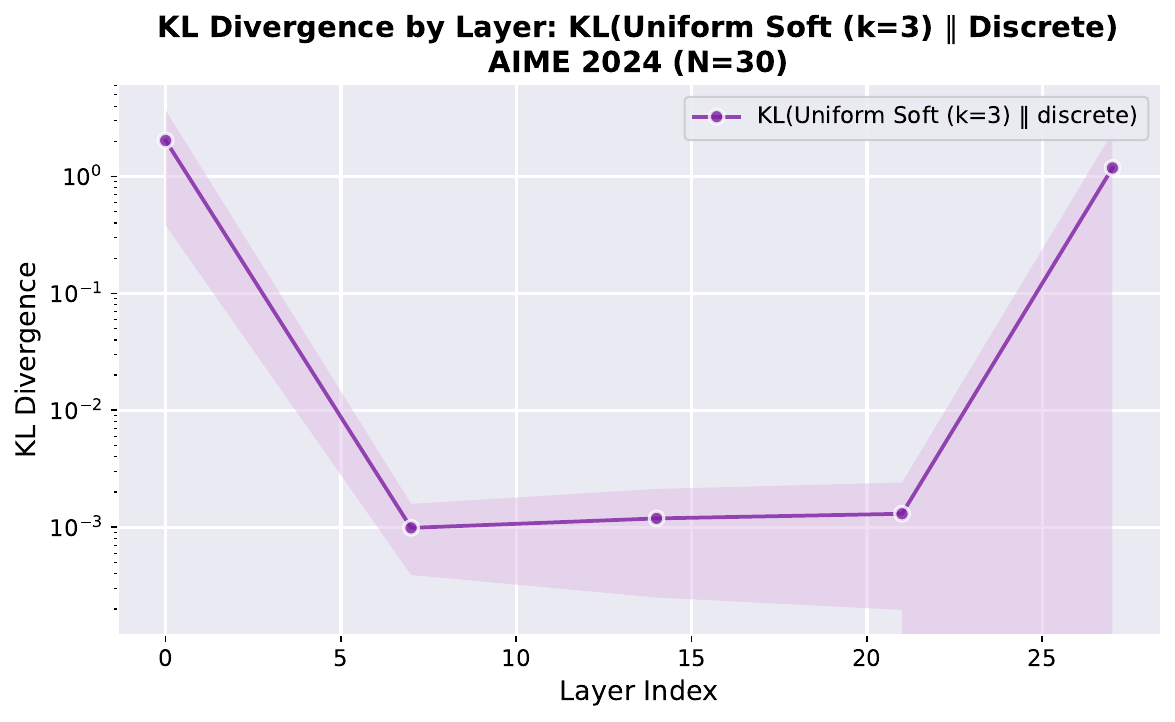}
        \caption{KL divergence — AIME 2024}
    \end{subfigure}
    \hfill
    \begin{subfigure}[b]{0.48\textwidth}
        \includegraphics[width=\textwidth]{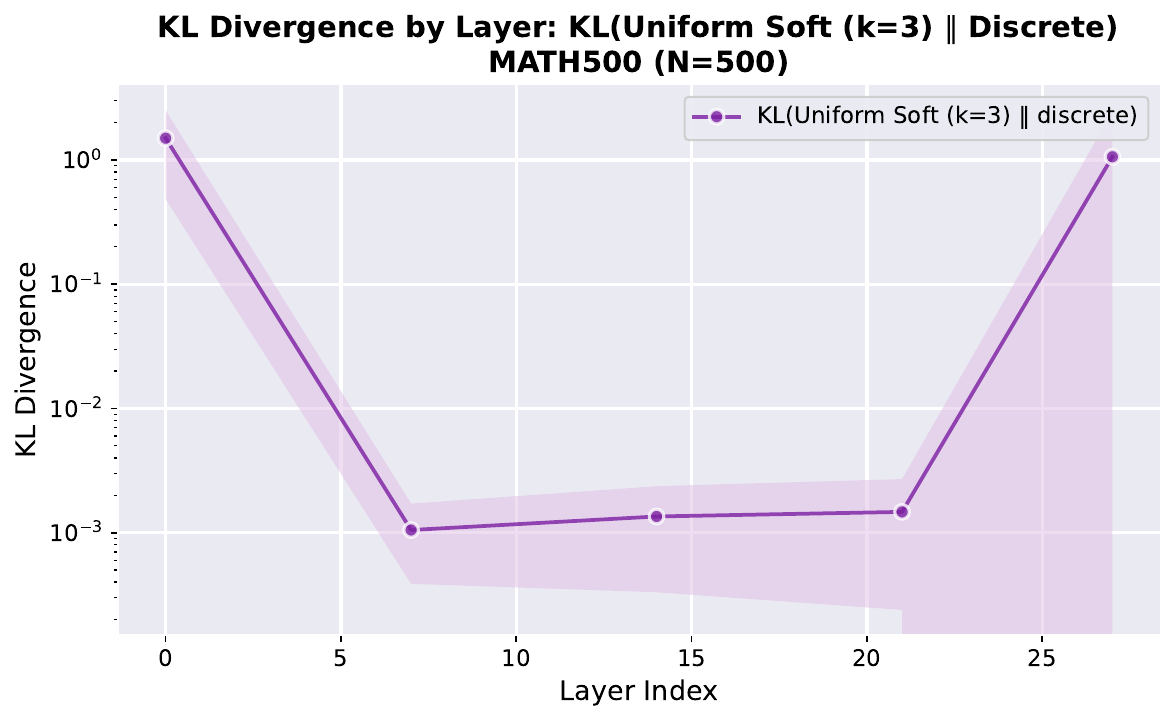}
        \caption{KL divergence — MATH500}
    \end{subfigure}
    \caption{\textbf{Qwen2-1.5B, uniform weighting $k=3$.} Entropy profiles (top) and KL divergence (bottom) for AIME 2024 (left) and MATH500 (right). Collapse is consistent with the 32B findings.}
    \label{fig:uniform-1.5b-k3}
\end{figure}

\begin{figure}[]
    \centering
    \begin{subfigure}[b]{0.48\textwidth}
        \includegraphics[width=\textwidth]{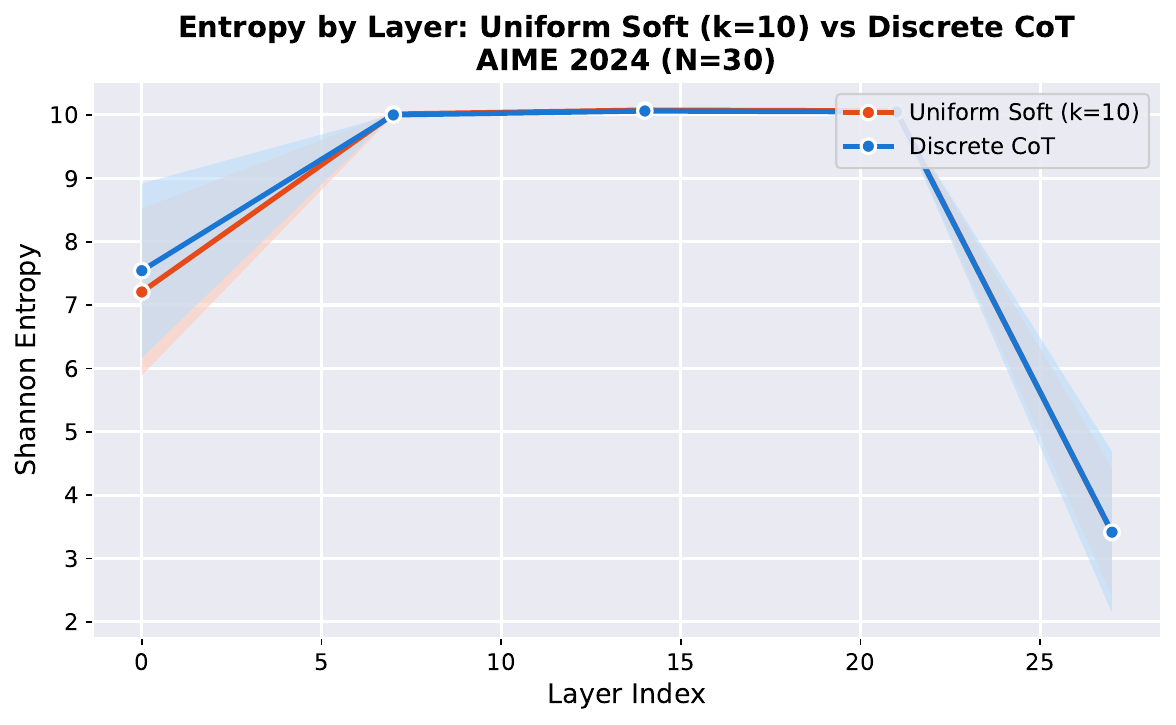}
        \caption{Entropy — AIME 2024}
    \end{subfigure}
    \hfill
    \begin{subfigure}[b]{0.48\textwidth}
        \includegraphics[width=\textwidth]{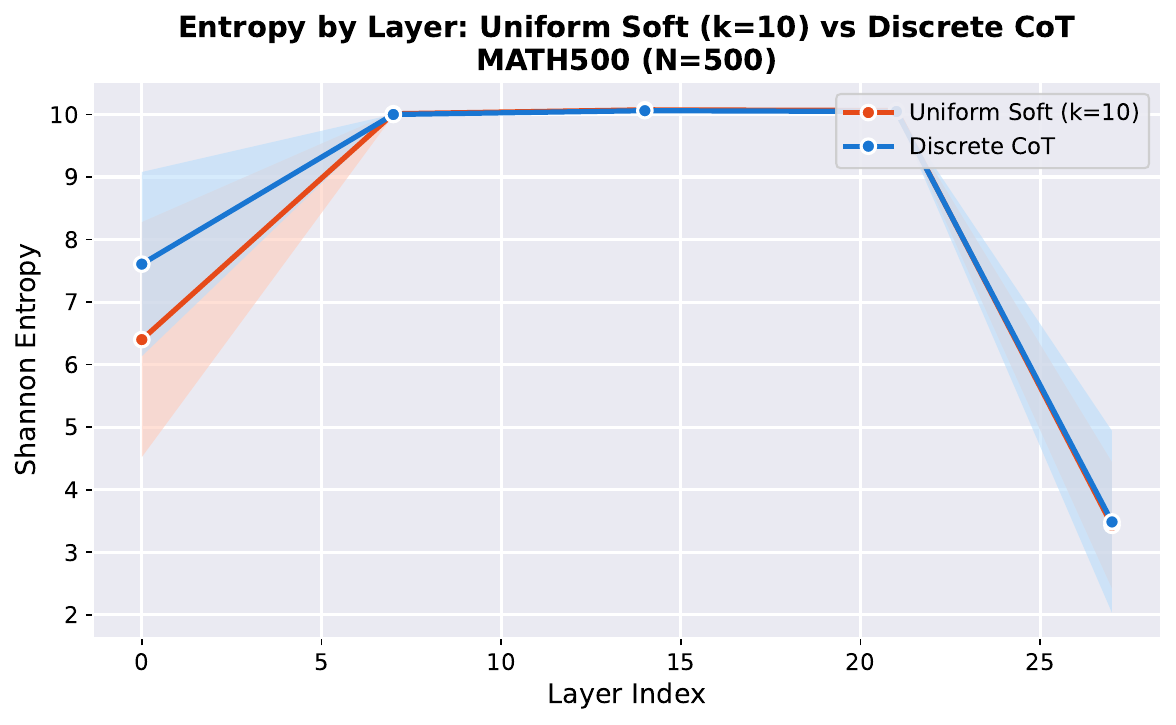}
        \caption{Entropy — MATH500}
    \end{subfigure}

    \vspace{0.5em}

    \begin{subfigure}[b]{0.48\textwidth}
        \includegraphics[width=\textwidth]{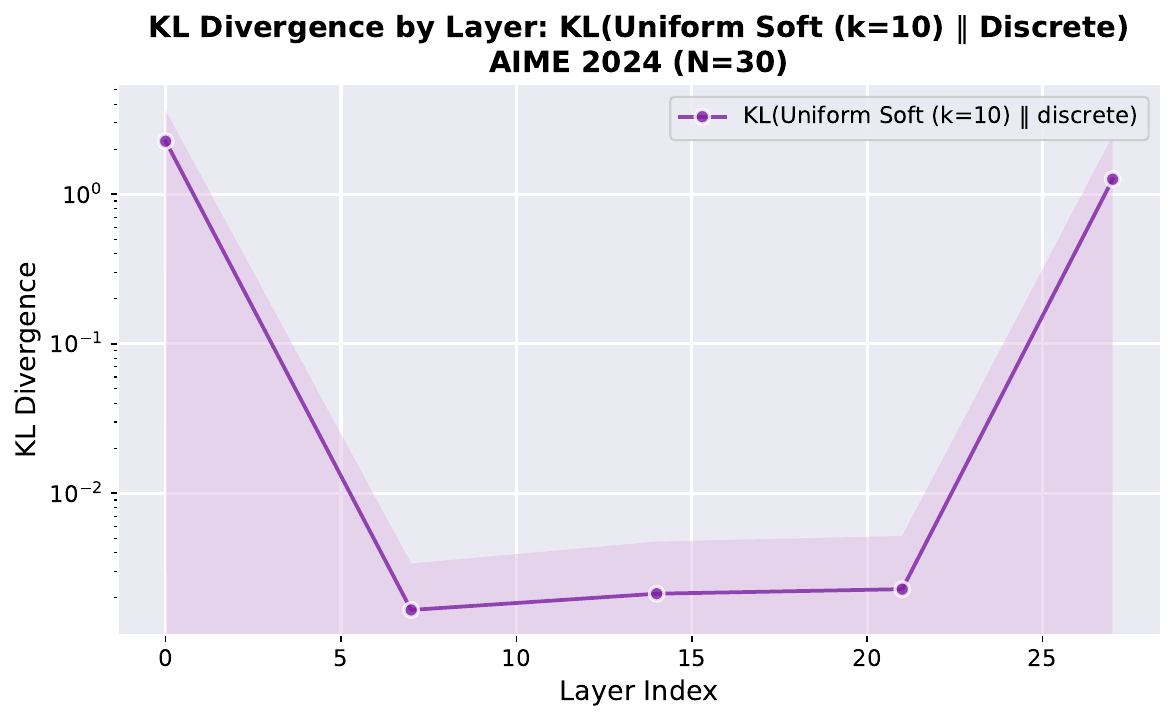}
        \caption{KL divergence — AIME 2024}
    \end{subfigure}
    \hfill
    \begin{subfigure}[b]{0.48\textwidth}
        \includegraphics[width=\textwidth]{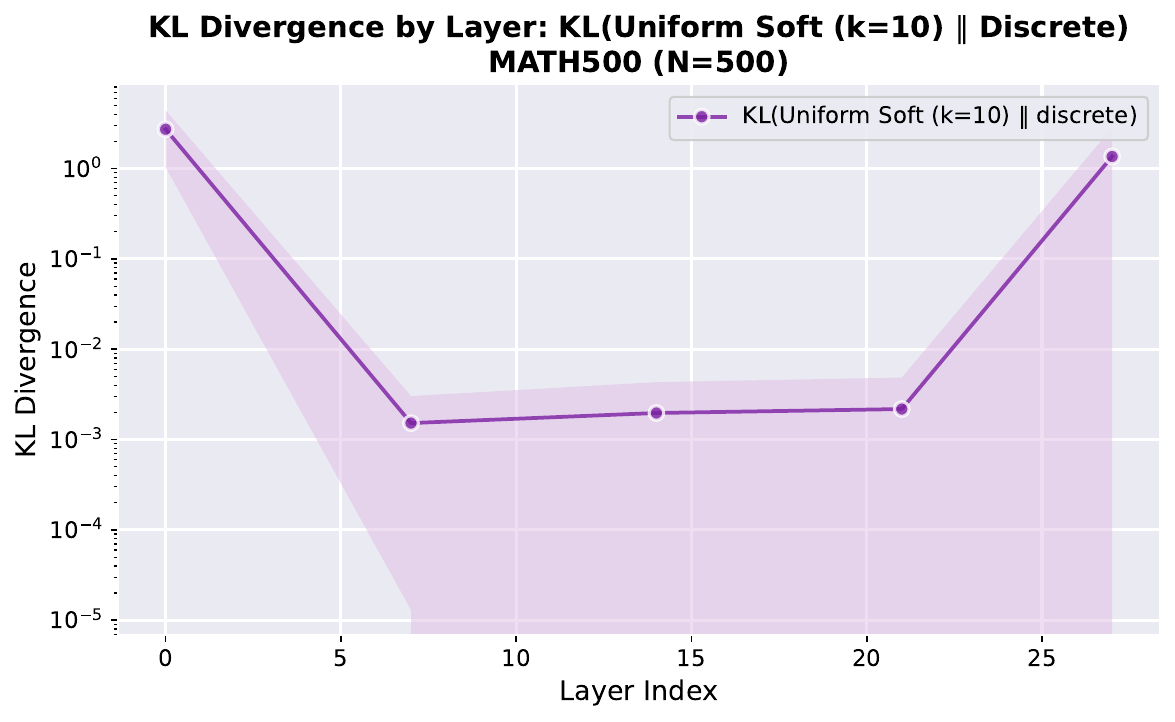}
        \caption{KL divergence — MATH500}
    \end{subfigure}
    \caption{\textbf{Qwen2-1.5B, uniform weighting $k=10$.} Entropy profiles (top) and KL divergence (bottom) for AIME 2024 (left) and MATH500 (right).}
    \label{fig:uniform-1.5b-k10}
\end{figure}

\begin{figure}[h]
    \centering
    \begin{subfigure}[b]{0.48\textwidth}
        \includegraphics[width=\textwidth]{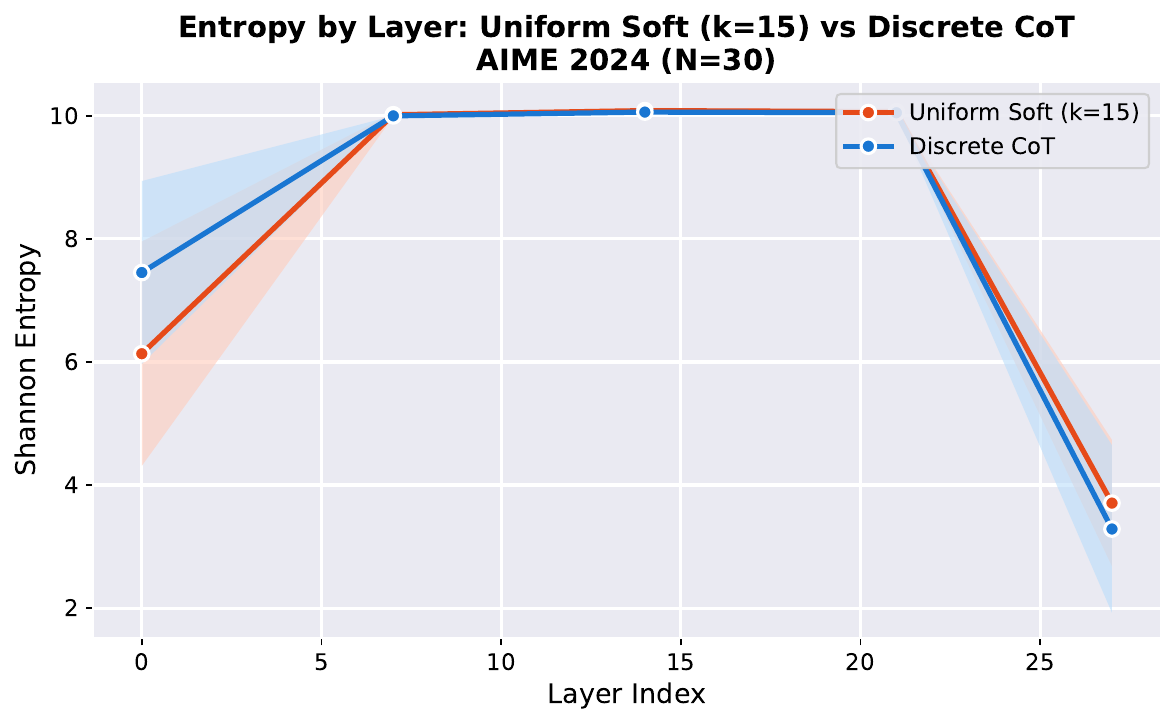}
        \caption{Entropy — AIME 2024}
    \end{subfigure}
    \hfill
    \begin{subfigure}[b]{0.48\textwidth}
        \includegraphics[width=\textwidth]{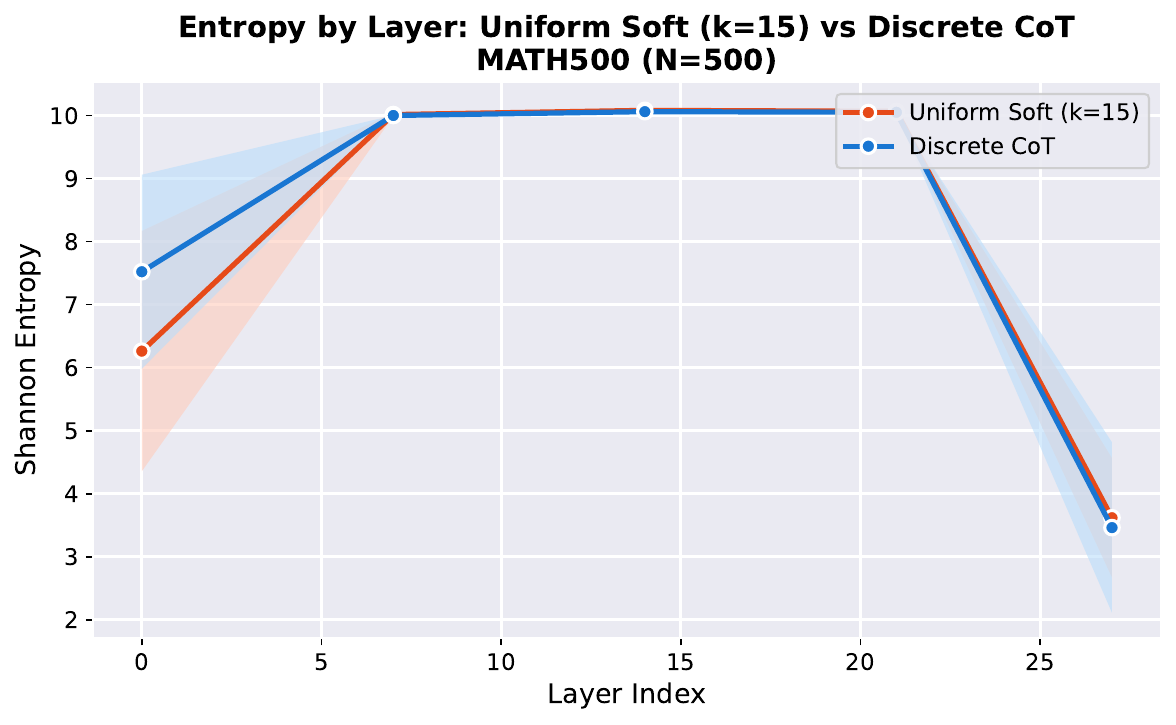}
        \caption{Entropy — MATH500}
    \end{subfigure}

    \vspace{0.5em}

    \begin{subfigure}[b]{0.48\textwidth}
        \includegraphics[width=\textwidth]{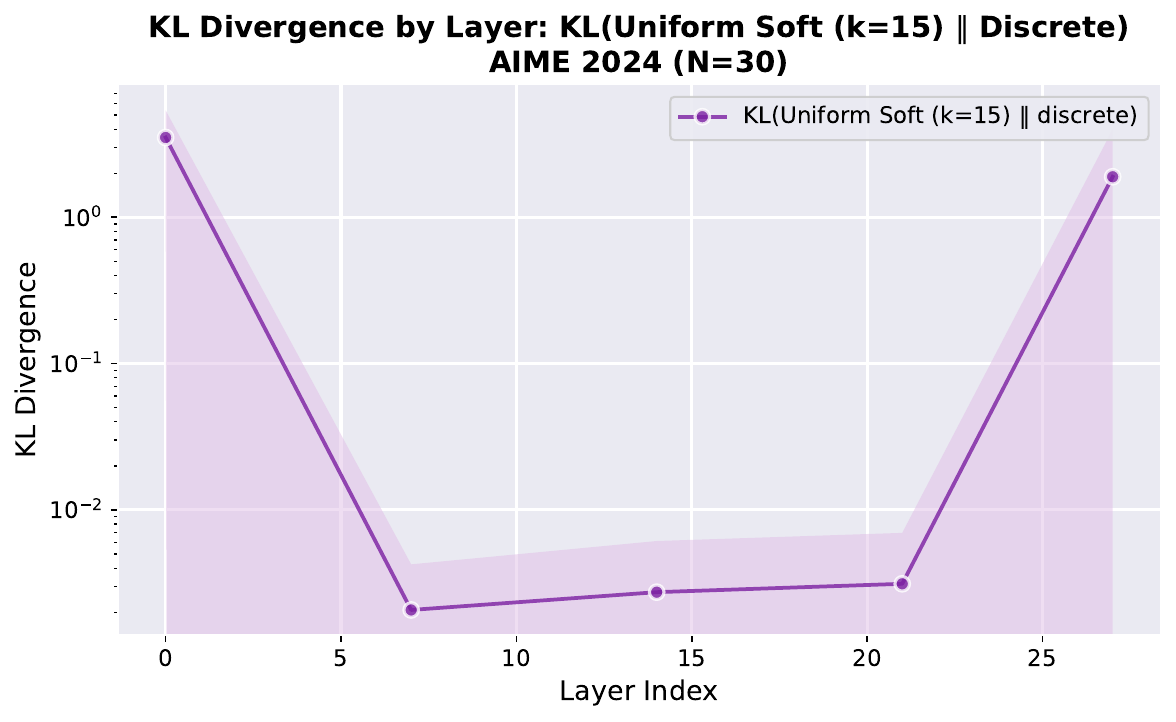}
        \caption{KL divergence — AIME 2024}
    \end{subfigure}
    \hfill
    \begin{subfigure}[b]{0.48\textwidth}
        \includegraphics[width=\textwidth]{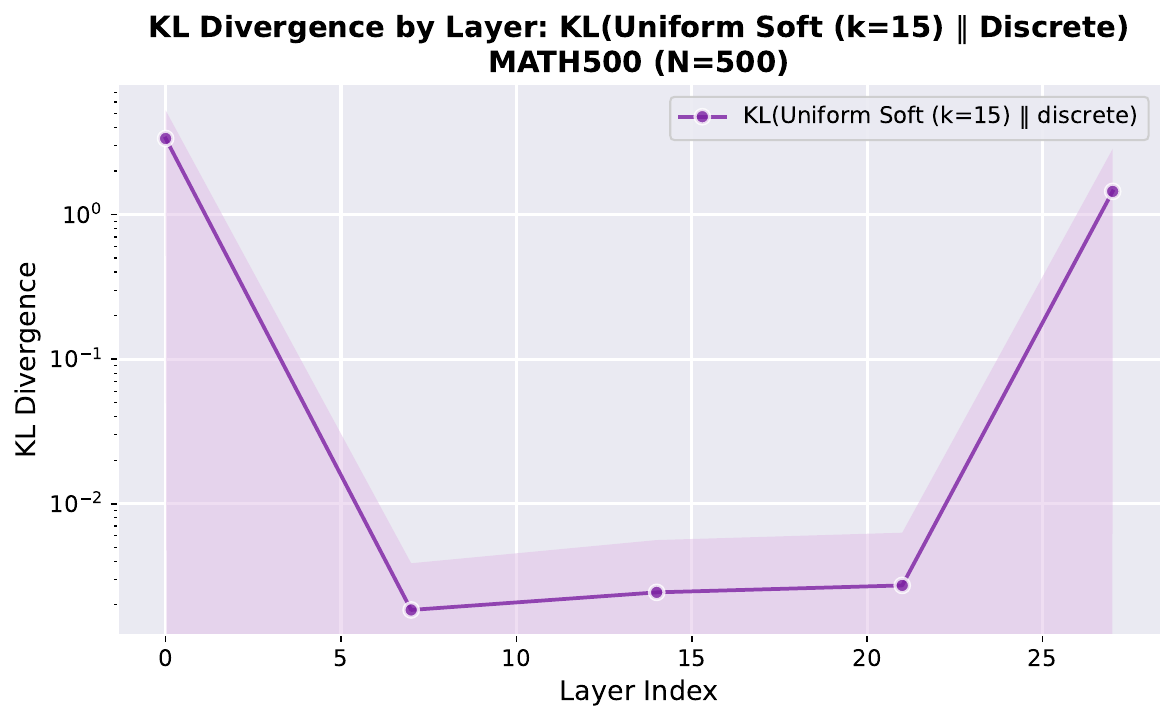}
        \caption{KL divergence — MATH500}
    \end{subfigure}
    \caption{\textbf{Qwen2-1.5B, uniform weighting $k=15$.} Entropy profiles (top) and KL divergence (bottom) for AIME 2024 (left) and MATH500 (right). Results are consistent across both model scales: uniform superposition collapses regardless of $k$.}
    \label{fig:uniform-1.5b-k15}
\end{figure}

For each combination of model (QwQ-32B, Qwen2-1.5B) and $k \in \{3, 10, 15\}$, we show entropy profiles and KL divergence side by side for AIME 2024 (N=30) and MATH500 (N=500) in \Cref{fig:uniform-32b-k3,fig:uniform-32b-k10,fig:uniform-32b-k15} (32B) and \Cref{fig:uniform-1.5b-k3,fig:uniform-1.5b-k10,fig:uniform-1.5b-k15} (1.5B). 

The finding is consistent across all conditions: uniform soft tokens collapse to near-argmax representations in middle layers, confirming that the collapse is not an artifact of the weighting scheme.

\subsubsection{QwQ-32B}

\subsubsection{Qwen2-1.5B}

\subsection{Benchmark Results}
\label{app:benchmark-results}
In this section we present a reproduction of accuracy results on MATH500 and AIME2024 for QwQ-32B. We find similar but not identical numbers to~\citet{zhang2025soft}. We hypothesize this is due to numerical precision issues.

\begin{table}[htbp]
    \centering
    \caption{MATH500 results with QwQ-32B (N=500). Discrete decoding uses standard greedy/sampling; Softmax and Uniform refer to the weighting scheme used to construct Soft Thinking embeddings.}
    \label{tab:math500}
    \begin{tabular}{llcccr}
    \toprule
    Run & Decoding & max\_topk & Cold Stop & Accuracy (\%) & Avg Tokens \\
    \midrule
    1 & Discrete & -- & 0.0 (none) & 97.08 & 4,326 \\
    2 & Discrete & -- & 0.1 & 97.08 & 4,326 \\
    3 & Discrete & -- & 0.2 & \textbf{97.19} & 4,307 \\
    4 & Softmax & 10 & 0.0 (none) & 96.47 & 4,222 \\
    5 & Softmax & 10 & 0.1 & 96.84 & 4,056 \\
    6 & Softmax & 15 & 0.01 & 96.84 & 4,044 \\
    7 & Softmax & 15 & 0.1 & 96.80 & 4,003 \\
    8 & Uniform & 3 & 0.1 & 93.97 & 5,388 \\
    \bottomrule
    \end{tabular}
\end{table}

\begin{table}[htbp]
    \centering
    \caption{AIME2024 results with QwQ-32B (N=30).}
    \label{tab:aime2024}
    \begin{tabular}{llcccr}
    \toprule
    Run & Decoding & max\_topk & Cold Stop & Accuracy (\%) & Avg Tokens \\
    \midrule
    1 & Discrete & -- & 0.0 (none) & \textbf{77.29} & 13,445 \\
    2 & Discrete & -- & 0.1 & \textbf{77.29} & 13,445 \\
    3 & Discrete & -- & 0.2 & \textbf{77.29} & 13,445 \\
    4 & Softmax & 15 & 0.01 & 75.83 & 12,445 \\
    5 & Softmax & 15 & 0.1 & 76.25 & 11,818 \\
    \bottomrule
    \end{tabular}
\end{table}

\section{Coconut: Experimental Details}
\label{app:coconut-details}

\subsection{ProsQA Task}
\label{app:prosqa-task}

ProsQA~\citep{hao2024training} is a synthetic graph-traversal QA task designed to evaluate multi-hop reasoning.
Each example consists of a randomly generated directed graph over named entities, a starting node, and a target node reachable via a sequence of directed edges.
The question provides the graph structure (as a list of edges) and asks for the entity reachable from a given starting node after a specified number of hops.
The ground-truth chain-of-thought consists of the sequence of intermediate entities visited during the traversal.

We use the ProsQA dataset provided with the original Coconut codebase, which contains 17,886 training examples and 300 validation examples.
Graph depths range from 3 to 6 hops, with the distribution concentrated at 4 and 5 hops.

\subsection{Training Setup}
\label{app:coconut-training}

We train GPT-2 (124M parameters, 12 layers, hidden dimension 768) on ProsQA using the Coconut training procedure of \citet{hao2024training}.
\Cref{tab:coconut-hyperparams} summarizes the key hyperparameters.

\begin{table}[htbp]
    \centering
    \caption{Coconut training hyperparameters on ProsQA.}
    \label{tab:coconut-hyperparams}
    \begin{tabular}{lcc}
    \toprule
    \textbf{Parameter} & \textbf{CoT baseline} & \textbf{Coconut} \\
    \midrule
    Base model & GPT-2 (124M) & GPT-2 (124M) \\
    Learning rate & $10^{-4}$ & $10^{-4}$ \\
    Optimizer & AdamW & AdamW \\
    Weight decay & 0.01 & 0.01 \\
    Batch size (per device) & 16 & 16 \\
    Gradient accumulation steps & 2 & 2 \\
    Total epochs & 50 & 50 \\
    Epochs per stage & -- & 5 \\
    Latent tokens per step ($c_\mathrm{thought}$) & -- & 1 \\
    Max latent stage & -- & 6 \\
    Precision & FP32 & FP32 \\
    Random seed & 0 & 0 \\
    \bottomrule
    \end{tabular}
\end{table}

Training follows a staged curriculum.
At stage $k$ (epochs $5k$ through $5(k+1)-1$), the first $k$ chain-of-thought steps are replaced by $k$ continuous latent tokens; the remaining steps are kept as discrete text.
The model is trained to predict only the remaining CoT steps and the final answer (labels for question and latent positions are masked with $-100$).
By stage 6 (epochs 30--34), all reasoning steps have been replaced by latent tokens, and the model must produce the answer using only continuous latent computation.
The optimizer is reset at the start of each epoch (\texttt{reset\_optimizer=True}).

Three special tokens are added to the vocabulary: \texttt{<|start-latent|>}, \texttt{<|latent|>}, and \texttt{<|end-latent|>}, all initialized from the embedding of the \texttt{<<} token.
During the forward pass, each \texttt{<|latent|>} token's embedding is replaced by the last hidden state from the preceding forward pass, implementing the continuous thought recurrence.
A custom collator left-pads batches to align latent token positions across examples, enabling KV cache reuse in the multi-pass forward.

We use \texttt{torchrun} with 4 GPUs; FSDP wraps the model but does not shard GPT-2's layers (effectively acting as DDP at this model scale).
The best CoT checkpoint is at epoch 49 (85.3\% validation accuracy); the best Coconut checkpoint is at epoch 50 (99\% validation accuracy).

\paragraph{Reproducibility note.}
The fine-tuned Coconut experiments reported in \Cref{sec:coconut} use a single training seed (seed\,=\,0), consistent with the setup of the original Coconut paper \citep{hao2024training}.
The core finding, that counterfactually removing the latent tokens changes the model's predictions by at most 1.0\% across models, replicates under independently trained SmolLM2-135M, SmolLM2-360M and SmolLM2-1.7B checkpoints (see \Cref{tab:no_latent_accuracy}), providing evidence that the result is not seed-specific.

\subsection{Entity Probing: Sample Sizes}
\label{app:coconut-probing-note}

\begin{figure}[htbp]
    \centering
    \includegraphics[width=\linewidth]{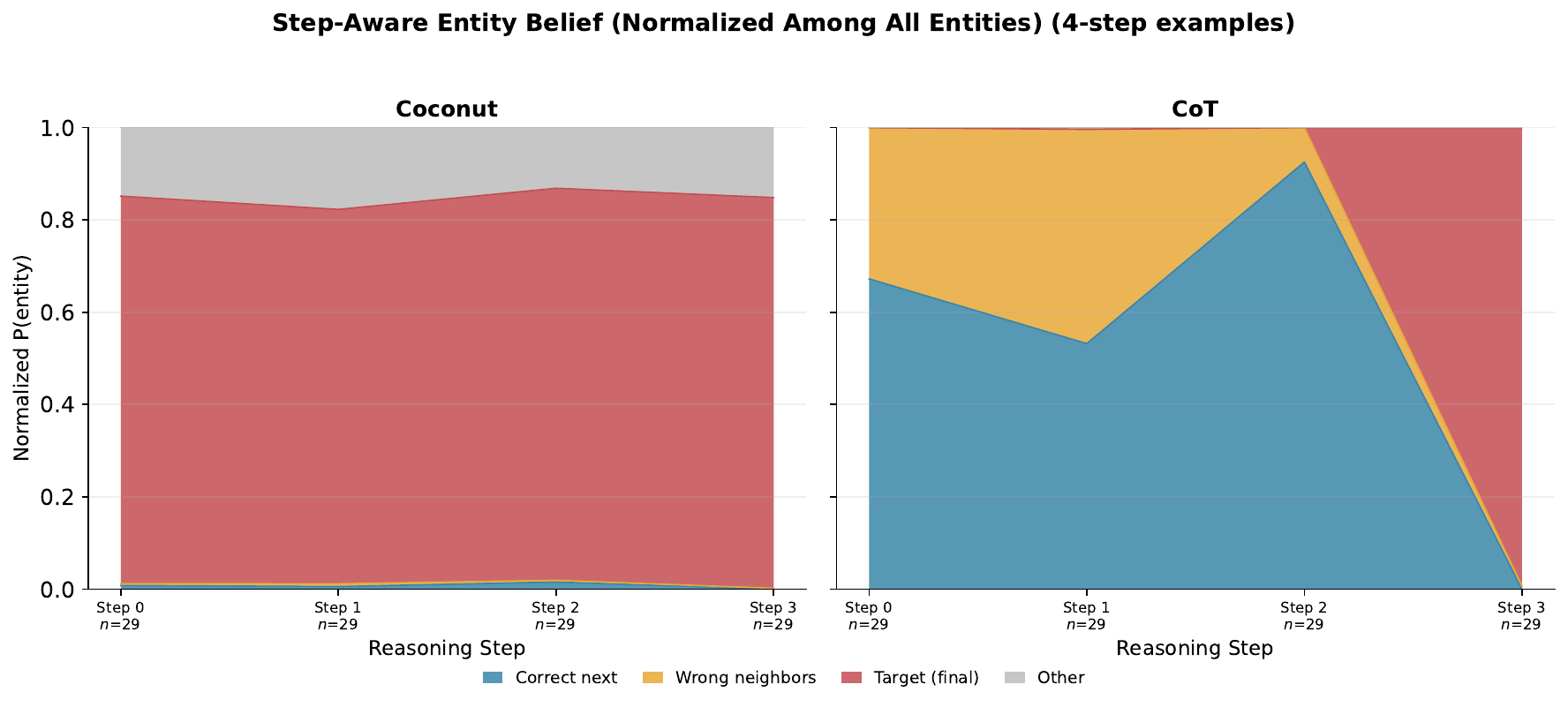}
    \caption{Normalized entity probability mass at each reasoning step for 4-step ProsQA examples (fine-tuned Coconut). Each stacked area shows the fraction of probability assigned to four entity categories: correct next hop (blue), wrong graph neighbors (orange), target/final answer (red), and other entities (gray). In the CoT model, probability mass shifts progressively toward the correct next entity at each step, consistent with step-by-step chain traversal. In the Coconut model, the target entity dominates from the first reasoning step onward, indicating that the model commits to the final answer without tracking intermediate hops.}
    \label{fig:stepwise-entity-norm-4step}
\end{figure}

The entity probing results shown in \Cref{fig:belief-evolution-finetuned,fig:stepwise-entity-norm-4step} use different sample sizes for the two conditions.
The Coconut model is probed on all $N=300$ ProsQA validation examples.
The CoT model is probed on a $N=50$ subset (the first 50 examples), due to the higher cost of teacher-forced probing over full CoT sequences.
Both subsets are sampled from the same validation split in the same order, so the 50 CoT examples are a subset of the 300 Coconut examples.
The qualitative pattern — target dominance from step 0 in Coconut vs.\ progressive correct-next tracking in CoT — is visually consistent and robust across both subsets, and holds on all step-count subgroups (3-step through 5-step) analyzed separately in \Cref{app:coconut-entity}.

\section{Coconut: Additional Results}
\label{app:coconut-results}

\subsection{Entity Distributions}
\label{app:coconut-entity}
In this section, we present entity distribution probing results for different step counts or scenarios which are not presented in the main paper.

\begin{figure}[htbp]
    \centering
    \includegraphics[width=\linewidth]{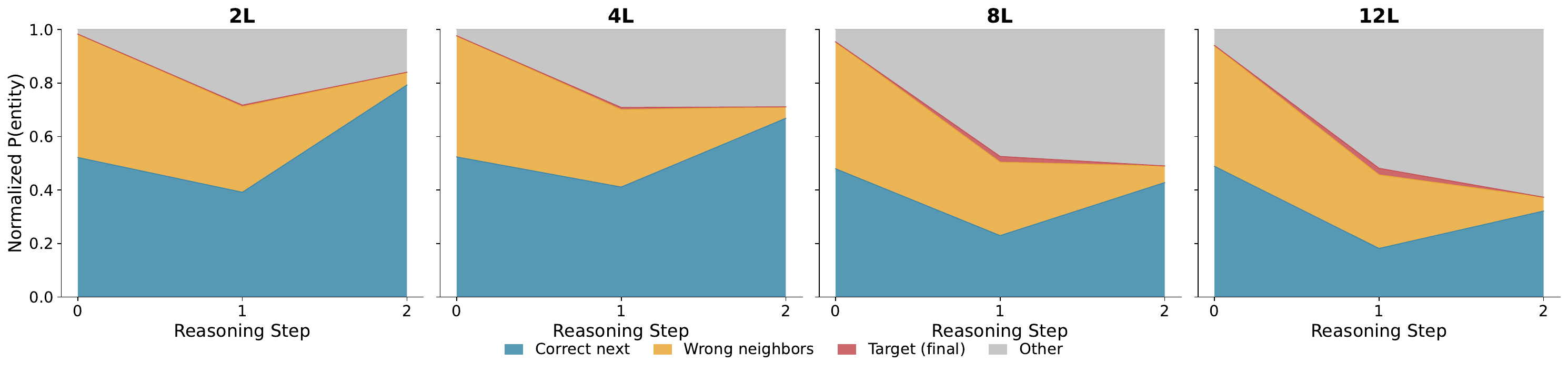}
    \caption{Step-aware entity belief across model depths for 3-step ProsQA examples (from-scratch Coconut). Each panel corresponds to a model trained from random initialization at a different depth (2L, 4L, 8L, 12L). Shallow models (2L, 4L) exhibit belief evolution across latent steps, with probability mass shifting toward the correct next entity at intermediate positions. Deeper models (8L, 12L) show no such progression; the target entity appears early without intermediate tracking.}
    \label{fig:depth-spectrum-3step}
\end{figure}

\begin{figure}[htbp]
    \centering
    \begin{subfigure}[b]{0.48\textwidth}
        \includegraphics[width=\textwidth]{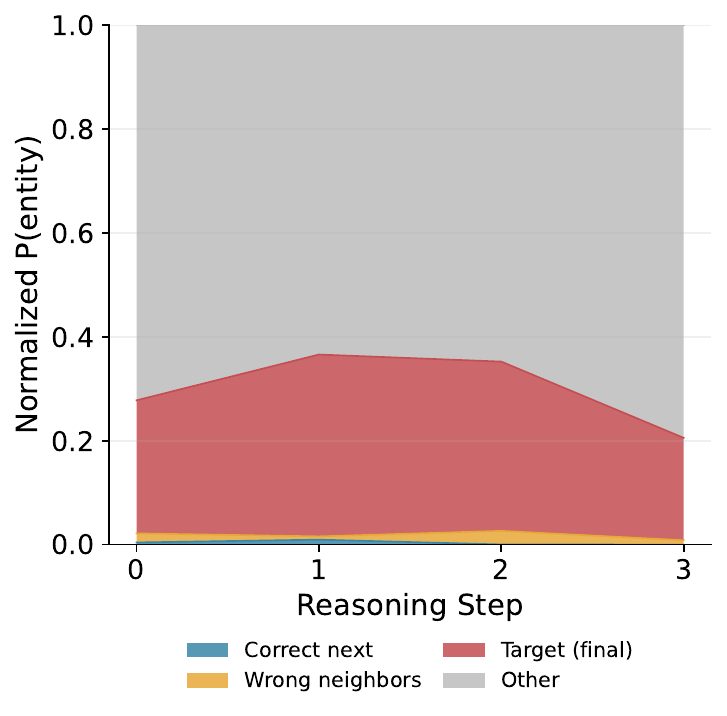}
        \caption{Coconut (latent positions)}
    \end{subfigure}
    \hfill
    \begin{subfigure}[b]{0.48\textwidth}
        \includegraphics[width=\textwidth]{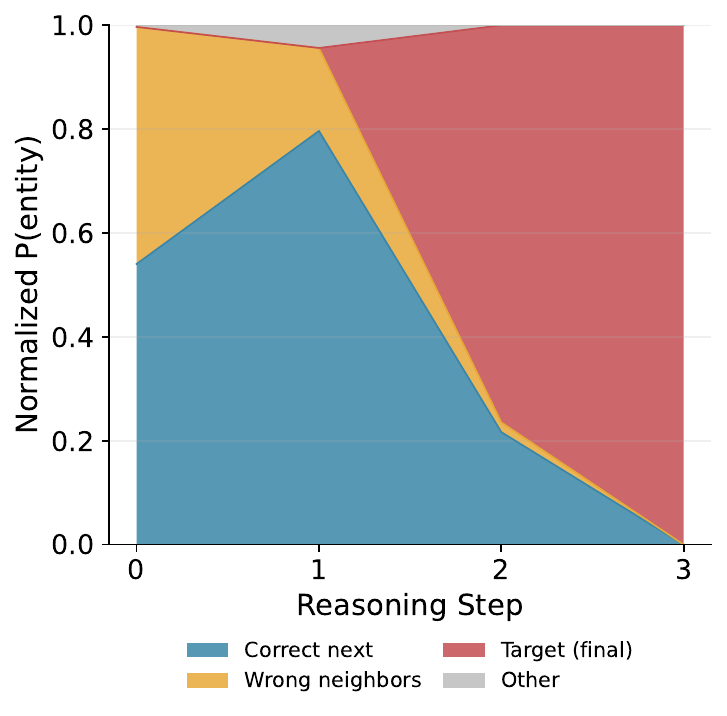}
        \caption{CoT (reasoning positions)}
    \end{subfigure}
    \caption{Step-aware entity belief for the 3.4\% of ProsQA test examples (17/500) that the Coconut model answers correctly with latent tokens but incorrectly without them. In the Coconut panel, the ``other'' category dominates (${\sim}70\%$), with the target receiving only ${\sim}25{-}35\%$ of the probability mass; the model does not confidently commit to the final answer from the question embedding alone on these examples, unlike the general population (\Cref{fig:stepwise-entity-norm-4step}). In contrast, the CoT model still performs step-by-step chain traversal on the same examples. This suggests that the latent tokens may be load-bearing for a small subset of examples where the question embedding does not suffice.}
    \label{fig:stepwise-hard-examples}
\end{figure}

\begin{figure}[htbp]
    \centering
    \begin{subfigure}[b]{0.48\textwidth}
        \includegraphics[width=\textwidth]{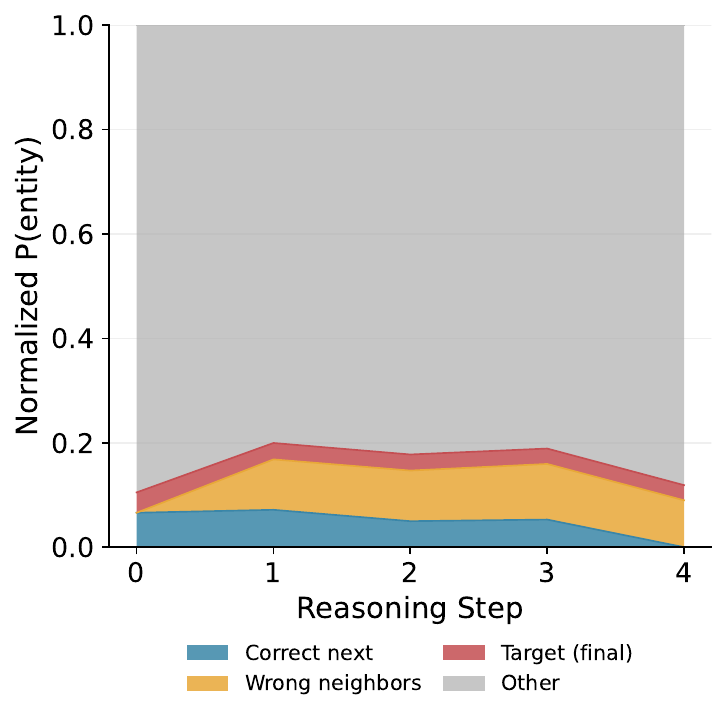}
        \caption{Coconut (latent positions)}
    \end{subfigure}
    \hfill
    \begin{subfigure}[b]{0.48\textwidth}
        \includegraphics[width=\textwidth]{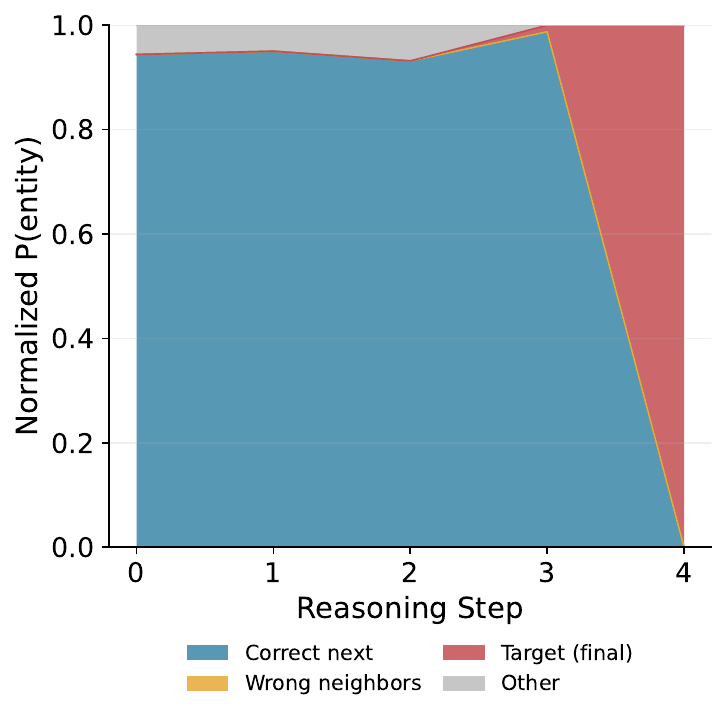}
        \caption{CoT (reasoning positions)}
    \end{subfigure}
    \caption{Normalized entity probability mass at each reasoning step on ProntoQA (fine-tuned Coconut). In the Coconut model, the ``other'' category dominates throughout (${\sim}80\%$), with correct next, wrong neighbors, and target each receiving minimal probability mass. In contrast, the CoT model tracks the correct next entity at each intermediate step before transitioning to the target at the final step.}
    \label{fig:prontoqa-stepwise-entity}
\end{figure}

\begin{figure}[htbp]
    \centering
    \includegraphics[width=0.5\linewidth]{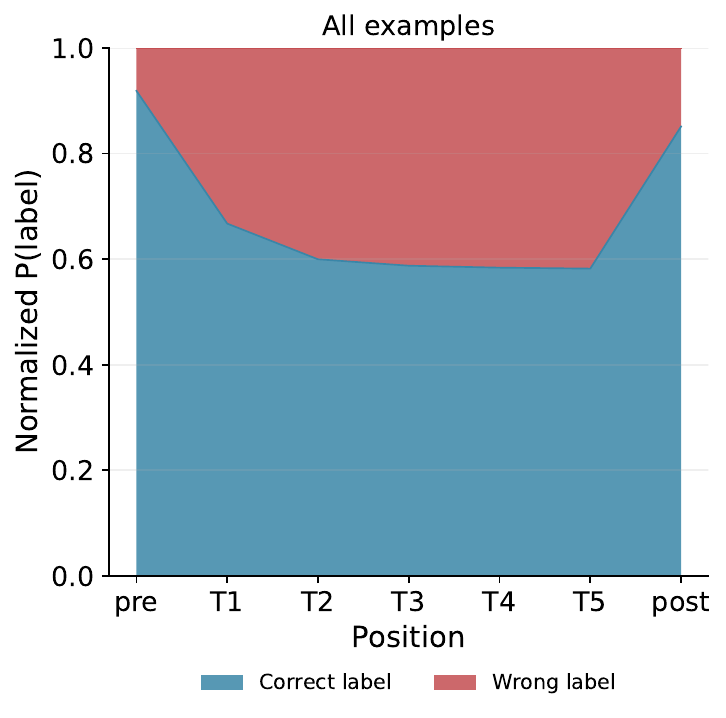}
    \caption{Normalized probability of the correct label (``True'' or ``False'') across latent positions for the Coconut model on ProntoQA. At the pre-thinking position the model assigns ${\sim}93\%$ of the probability mass to the correct label; this drops to ${\sim}60\%$ during latent reasoning (T1--T5) before recovering at the post-thinking position. The dip during latent positions is consistent with the model redistributing probability mass without performing productive intermediate computation. Probing for these labels is more consistent with the ProntoQA task as the model must return a binary ``True or False'' response, as opposed to returning the entity name as is the case in ProsQA.}
    \label{fig:true-false-belief}
\end{figure}

\subsection{Coconut Other Results}
\label{app:coconut-logitlens}
In this section, we present additional experiments not presented in the paper. Mainly, we look at the entropy throughout layers for Coconut and CoT models on the ProsQA task, plot Coconut gradient norms throughout training and show entity belief plots for the ProntoQA~\citep{saparov2022language} task.

\begin{figure}[htbp]
    \centering
    \includegraphics[width=\linewidth]{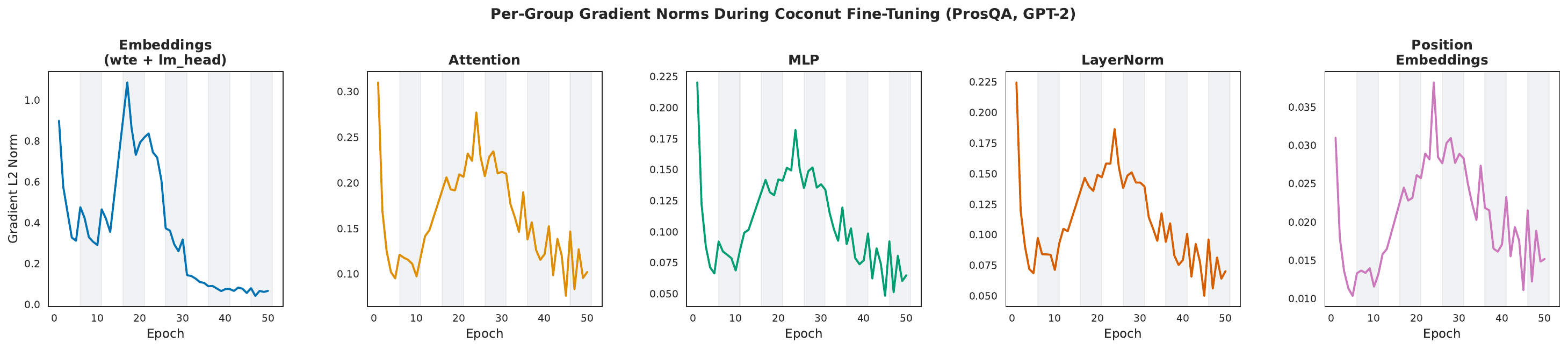}
    \caption{Mean $L_2$ gradient norms per parameter group during Coconut fine-tuning on ProsQA (GPT-2). Each panel shows one parameter group: word token embeddings (wte + lm\_head), attention, MLP, LayerNorm, and positional embeddings (wpe). Alternating shaded bands denote the five-epoch training stages of the Coconut curriculum. Note that gradient magnitudes remain non-trivial across all groups and stages, indicating that the model parameters are being actively updated throughout training; however, as shown in \Cref{sec:coconut}, this training does not yield latent representations that participate in multi-step reasoning.}
    \label{fig:grad-norms-by-group}
\end{figure}

\begin{figure}
    \centering
    \includegraphics[width=\linewidth]{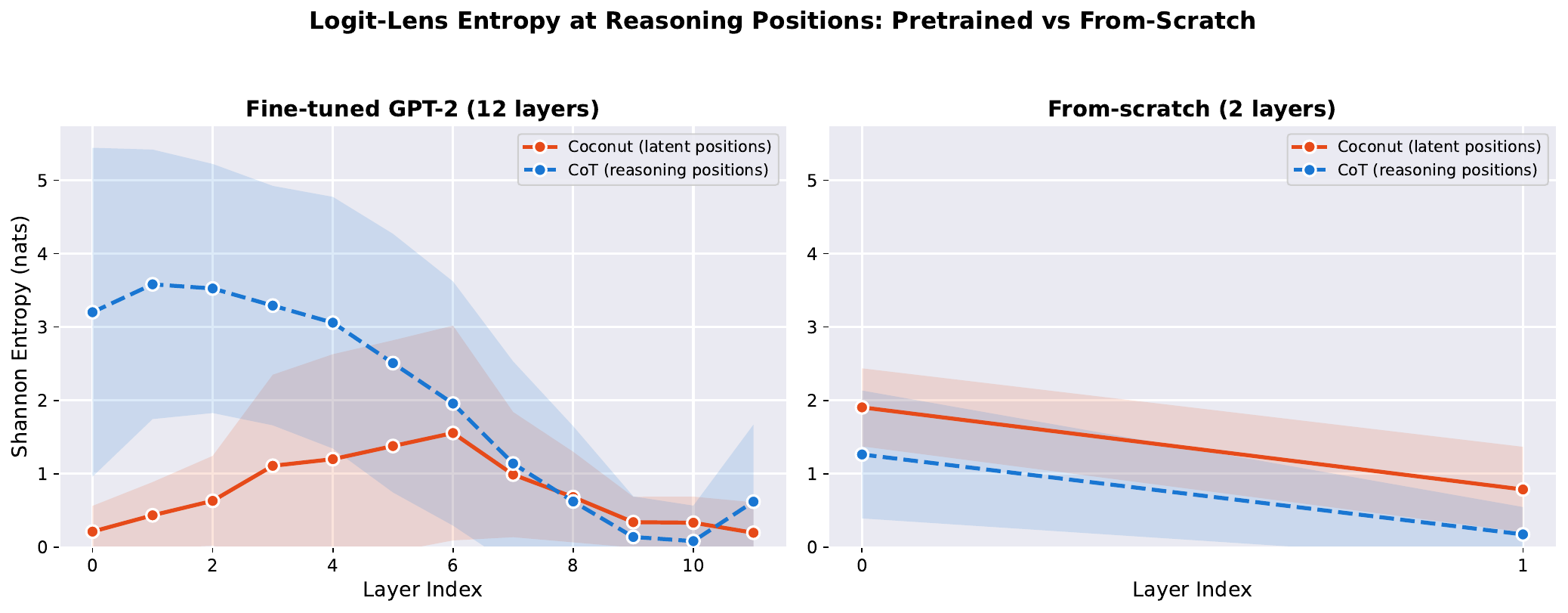}
    \caption{\logitlens entropy at reasoning positions for fine-tuned GPT-2 (left) and a from-scratch 2-layer model (right). In the fine-tuned model, Coconut latent positions maintain near-zero entropy across all layers, consistent with early commitment to a single token. In the from-scratch model, this pattern reverses: Coconut latent positions retain higher entropy than CoT, suggesting richer latent representations. Shaded regions denote $\pm 1$ standard deviation.}
    \label{fig:placeholder}
\end{figure}

\subsection{Coconut experiments on different models}
We replicate the fine-tuned Coconut no-latent finding using independently trained SmolLM2-135M, SmolLM2-360M and SmolLM2-1.7B models on ProsQA.
Each model was trained with a fresh random initialization and the same Coconut curriculum as the GPT-2 baseline; training seeds differ from the GPT-2 runs.
To probe the effect of the latent tokens causally, we exploit the binary structure of ProsQA (queries of the form \textit{``Is Rex a blicket or a gorple?''}): we remove the latent thoughts and score the log probabilities the model assigns to both candidate completions, thereby measuring counterfactually how it would have answered without any latent computation.
\Cref{tab:no_latent_accuracy} shows that this intervention changes the predictions by at most 1.0\% across all four model families, confirming that the latent tokens are not load-bearing and that the behavior is not specific to GPT-2 or a single training seed.

\begin{table}[h]
\centering
\caption{Counterfactual no-latent evaluation on ProsQA across independently trained models. \emph{Coconut (no latents)} removes all latent tokens and scores both candidate completions; Drop $=$ Coconut (no latents) $-$ Coconut.}
\label{tab:no_latent_accuracy}
\begin{tabular}{lcccc}
\toprule
\textbf{Model} & \textbf{CoT} & \textbf{Coconut} & \textbf{Coconut (no latents)} & \textbf{Drop} \\
\midrule
GPT-2 (124M)       & 85.3 & 99.0  & 99.0  & $-0.0$ \\
SmolLM2-135M       & 72.7 & 93.3  & 92.3  & $-1.0$ \\
SmolLM2-360M       & 85.0 & 98.7  & 98.0  & $-0.7$ \\
SmolLM2-1.7B       & 98.3 & 100.0 & 100.0 & $-0.0$ \\
\bottomrule
\end{tabular}
\end{table}

\subsection{Ablation over embedding width}
Table showing the importance of the width in models:
\begin{table}[t]
\centering
\caption{Test accuracy (\%) with and without latent tokens across embedding dimensions,
for 2-layer and 4-layer from-scratch Coconut models on ProsQA.
The gap (w/ $-$ w/o) measures reliance on latent reasoning;
a large gap indicates the model genuinely uses latent tokens.
Graph width $\approx 28$ nodes; the theoretical lower bound is $\Omega(28)$.}
\label{tab:embd_ablation}
\begin{tabular}{r ccc ccc}
\toprule
& \multicolumn{3}{c}{\textbf{2-Layer}} & \multicolumn{3}{c}{\textbf{4-Layer}} \\
\cmidrule(lr){2-4} \cmidrule(lr){5-7}
$d_\text{embd}$ & w/ latents & w/o latents & gap & w/ latents & w/o latents & gap \\
\midrule
8   & 40.8 & 40.6 & $+$0.2 & 35.3 & 35.1 & $+$0.2 \\
16  & 57.5 & 54.9 & $+$2.6 & 55.6 & 55.4 & $+$0.2 \\
32  & 58.9 & 57.8 & $+$1.1 & 53.9 & 53.2 & $+$0.7 \\
64  & 64.9 & 62.3 & $+$2.6 & 63.2 & 62.5 & $+$0.7 \\
128 & 67.1 & 60.6 & $+$6.5 & 61.3 & 60.4 & $+$0.9 \\
\midrule
256 & 75.4 & 27.2 & $\mathbf{+48.2}$ & 66.6 & 60.6 & $+$6.0 \\
384 & 91.6 & 57.8 & $\mathbf{+33.8}$ & 90.5 & 32.7 & $\mathbf{+57.8}$ \\
768 & 96.7 & 32.0 & $\mathbf{+64.7}$ & 96.2 & 16.0 & $\mathbf{+80.2}$ \\
\bottomrule
\end{tabular}
\end{table}

\subsection{From-Scratch Coconut: Multi-Seed Results}
\label{app:fromscratch-seeds}

To assess reproducibility of the from-scratch Coconut depth scaling results (\Cref{fig:depth-spectrum-3step,fig:latent-collapse-comparison}), we train three independently seeded runs (seeds 0, 1, 2) for each depth.
\Cref{tab:fromscratch-seeds} reports accuracy with and without latent tokens, and the resulting gap, as mean $\pm$ standard deviation across the three runs.

\begin{table}[h]
\centering
\caption{From-scratch Coconut on ProsQA ($d_\text{embd}=768$): accuracy (\%) with and without latent tokens, across 3 independent seeds. Gap = w/ $-$ w/o. Shallow models (2L, 4L) show a large, consistent gap; deeper models show a smaller but more variable gap, consistent with the depth-dependent shortcut pattern discussed in the main text.}
\label{tab:fromscratch-seeds}
\setlength{\tabcolsep}{5pt}
\begin{tabular}{r ccc}
\toprule
\textbf{Depth} & \textbf{w/ latents} & \textbf{w/o latents} & \textbf{gap} \\
\midrule
2L  & $96.0 \pm 0.9$  & $30.9 \pm 8.4$  & $65.1 \pm 9.0$  \\
4L  & $96.9 \pm 1.2$  & $33.2 \pm 25.0$ & $63.7 \pm 23.7$ \\
8L  & $83.4 \pm 2.5$  & $67.9 \pm 5.5$  & $15.4 \pm 3.1$  \\
12L & $71.7 \pm 8.8$  & $54.4 \pm 23.4$ & $17.3 \pm 26.8$ \\
\bottomrule
\end{tabular}
\end{table}

The 2L gap is large and tight across seeds ($65.1 \pm 9.0$\,pp).
The 4L gap is similarly large on average but more variable ($63.7 \pm 23.7$\,pp), driven by one seed where the no-latent condition reaches 61.8\% — suggesting that at this depth the model can sometimes learn shortcuts.
At 8L and 12L the gap shrinks substantially, consistent with the main text argument that deeper models can extract the answer without latent tokens.
These results confirm that the shallow-model reliance on latent tokens reported in the paper is not an artifact of a single training run.

\end{document}